\documentclass{article}
\usepackage{microtype}
\usepackage{graphicx}
\usepackage{subfigure}
\usepackage{booktabs}
\usepackage{amssymb}
\usepackage{longtable}
\usepackage{supertabular}
\usepackage{multirow}
\usepackage{multicol}
\usepackage{float}
\usepackage{listings}
\usepackage{hyperref}

\let\appendixAddContentsLine\addcontentsline
\usepackage[accepted]{mlsys2025}
\newcommand{\methodname}{StrataFlux}
\newcommand{\papertitlefirstline}{Memory-Hierarchy-Native}
\newcommand{\papertitlesecondline}{Low-Bit LLM Inference}
\newcommand{\papertitlesuffix}{\papertitlefirstline{} \papertitlesecondline{}}
\newcommand{\papertitledisplay}{\papertitlefirstline\texorpdfstring{\\}{ }\papertitlesecondline}
\newcommand{\papertitle}{\methodname{}: \papertitlesuffix}
\newcommand{\paperrunningtitle}{\papertitle}
\newcommand{\figureversiondir}{figures/arxiv}
\newif\ifprovisionalname
\provisionalnametrue
\newcommand{\provisionalnamenote}{\methodname{} is a provisional system name and may change in future revisions.}

\hypersetup{hypertexnames=false,pdfsubject={},pdftitle={\papertitle},pdfkeywords={LLM inference, quantization, memory hierarchy}}
\input{shared/appendix_setup}
\newlength{\templatetextfloatsep}
\newlength{\templatefloatsep}
\newlength{\templatedbltextfloatsep}
\makeatletter
\let\versionlabel\label
\newcommand{\versionlabelprefix}{main-}
\renewcommand{\label}[1]{\versionlabel{#1}\versionlabel{\versionlabelprefix#1}}
\makeatother
\begin{document}
\setlength{\textfloatsep}{10pt plus 2pt minus 2pt}
\setlength{\floatsep}{8pt plus 2pt minus 2pt}
\setlength{\dbltextfloatsep}{10pt plus 2pt minus 2pt}
\twocolumn[
\ifprovisionalname
\mlsystitle{\methodname{}\texorpdfstring{\textsuperscript{*}}{}: \papertitledisplay}
\else
\mlsystitle{\methodname{}: \papertitledisplay}
\fi
\begin{mlsysauthorlist}
\mlsysauthor{Jake Hyun}{cornell}
\mlsysauthor{Ziyu Zhao}{cmu}
\mlsysauthor{Jae-sun Seo}{cornell}
\mlsysauthor{Mohamed S. Abdelfattah}{cornell}
\end{mlsysauthorlist}
\mlsysaffiliation{cornell}{Cornell University}
\mlsysaffiliation{cmu}{Carnegie Mellon University}
\begin{center}
\textsuperscript{1}Cornell University\qquad
\textsuperscript{2}Carnegie Mellon University
\end{center}

\vskip 0.15in
\begin{abstract}
Quantization reduces the memory and bandwidth demands of large language model (LLM) inference but can degrade accuracy; offloading higher-precision data can instead stall generation. We introduce \methodname{}, a framework that challenges this tradeoff by recovering accuracy lost to low-bit weight and key--value (KV) cache quantization while retaining quantization's memory and speed advantages. \methodname{} co-designs refinement representation, selection, and delivery to make host RAM and solid-state drives useful sources of additional precision. Its \emph{selective precision streaming} reduces transfer volume, overlaps delivery with computation, and reuses fetched precision across decoding steps so generation need not wait for every transfer. We evaluate 2--4-bit weights and 2--3-bit KV-cache on dedicated GPUs and unified-memory platforms. For 2-bit Llama-3.1-8B-Instruct weights, \methodname{} reduces WikiText-2 perplexity from 9.46 to 8.40 while maintaining 95.4\% of the quantized base's throughput on RTX 5090. With 2-bit weights and KV, \methodname{} raises Qwen3-8B's AIME 2024/2025 accuracy from 11.0\% to 47.1\% at unchanged GPU memory use. At 128K context, Llama-3.1-8B-Instruct with 3-bit weights and 2-bit KV achieves comparable perplexity and speed to vLLM's 4-bit AWQ weights and 8-bit KV on RTX 5090, using approximately half the GPU memory. These results show that GPU-resident precision need not set the accuracy ceiling, nor offloading-tier latency dictate decoding speed.

\end{abstract}
]
\ifprovisionalname
\begingroup
\renewcommand{\thefootnote}{*}
\footnotetext{\provisionalnamenote}
\endgroup
\fi
\section{Introduction}
\label{sec:introduction}

LLM inference involves three coupled objectives: accuracy, latency, and memory use. Model weights encode learned parameters, while the key--value (KV) cache retains keys and values from earlier tokens. Their representations determine both the precision available for computation and the volume of data stored and moved during generation.

Quantization and offloading address limited fast-memory capacity differently. Quantization reduces storage and traffic with fewer bits; it can accelerate bandwidth-bound inference but may degrade accuracy~\citep{frantar2023optq,liu2024kivi}. Offloading preserves stored precision by placing data in a larger, slower backing tier, but transfer latency can delay generation~\citep{na2025flexinfer}. These approaches suggest an apparent choice between accuracy loss and transfer latency.

We argue that this is a false choice imposed by separating representation from placement. Considered alone, quantization discards precision that another tier could retain, while offloading brings in that tier only after choosing the representation. Can we use fewer bits in fast memory without sacrificing accuracy or slowing inference?

\begin{figure}[t]
\centering
\includegraphics[width=\linewidth]{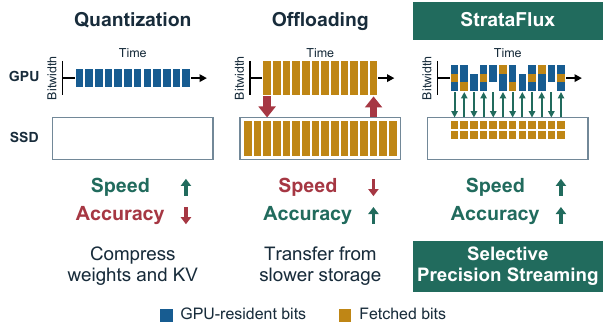}
\caption{\methodname{} against quantization and offloading. Column height indicates bitwidth; horizontal flow indicates time. Selective precision streaming recovers accuracy while retaining low-bit inference's memory and speed advantages.}
\label{fig:idea-overview}
\end{figure}

We address this question with \methodname{}, which employs \emph{selective precision streaming} to distribute weight and KV precision across tiers (Figure~\ref{fig:idea-overview}). Compact, quantized weights and KV stay in GPU memory as the \emph{resident base}, providing high-bandwidth access.
Larger, cheaper tiers---host memory or solid-state drives (SSDs)---retain additional precision, but their lower bandwidth and higher latency call for selective fetching and flexible scheduling.

Recovering accuracy by fetching all additional precision would erode quantization bandwidth savings. Selectivity is viable because protecting a small subset of weights can improve low-bit accuracy~\citep{lin2024awq,kim2024squeezellm}, while attention can concentrate on a small, query-dependent subset of tokens~\citep{tang2024quest,ribar2024sparq}. \methodname{} exploits this uneven importance to fetch additional precision selectively for the current computation.

Retaining quantization's speed advantage also requires flexible scheduling: selective transfers can still stall generation when their results are needed immediately. For weights, overlapping a fetch only with the weight-matrix operation it refines limits the time available to hide delivery. \methodname{} widens this window by selecting channels earlier or adding their refinement later, allowing transfers and refinement computation to overlap intervening operations. For both weights and KV, it also reuses fetched precision across decoding steps, avoiding repeated transfers and letting data arriving too late for one step benefit later selections.

\methodname{} makes LLM inference native to the memory hierarchy, recovering accuracy while retaining the memory and speed advantages of quantization. We contribute:

\begin{list}{\arabic{enumi}.}{\usecounter{enumi}
\setlength{\leftmargin}{1.6em}
\setlength{\topsep}{3pt}
\setlength{\partopsep}{0pt}
\setlength{\itemsep}{2pt}
\setlength{\parsep}{0pt}}
\item \textbf{Selective precision streaming.} We augment resident low-bit weights and KV with refinement bits from host RAM or SSD, concentrating transfers on channels and tokens that matter most to each computation.
\item \textbf{Asynchronous precision delivery.} By decoupling when precision is fetched from when it is used, \methodname{} hides the latency of fetches without stalling generation.
\item \textbf{Hierarchy-native inference.} By co-designing numerical representation with memory placement, \methodname{} removes the choice between quantization and offloading, enabling fast, accurate, and memory-efficient inference on both dedicated GPU and unified memory platforms.
\end{list}

For 2-bit Llama-3.1-8B-Instruct weights, \methodname{} reduces WikiText-2 perplexity from 9.46 to 8.40 while maintaining 95.4\% of the quantized base's throughput on RTX 5090. With 2-bit weights and KV, \methodname{} raises Qwen3-8B's AIME 2024/2025 accuracy from 11.0\% to 47.1\% at unchanged GPU memory use. At 128K context, Llama-3.1-8B-Instruct with 3-bit weights and 2-bit KV achieves comparable perplexity and speed to vLLM's 4-bit AWQ weights and 8-bit KV on RTX 5090, using approximately half the GPU memory.

\section{Background and Motivation}
\label{sec:background}

\subsection{Post-Training Quantization}
\label{sec:hierarchy-motivation}

Interactive on-device LLM inference repeatedly reads model weights and the KV cache to generate new tokens. Their size constrains which models and contexts fit in fast memory, while the traffic per token can make memory bandwidth the bottleneck for generation speed~\citep{pope2023scaling,liu2024kivi}. Quantization reduces these demands by representing values with low-bit codes, such as codebook indices~\citep{han2016deepcompression} or scaled and offset integers~\citep{jacob2018quantization}. Post-training quantization (PTQ) compresses weights without retraining the model~\citep{frantar2023optq}, while KV-cache quantization compresses the keys and values generated during inference~\citep{liu2024kivi}.

These savings introduce errors that affect accuracy differently across values. Large-magnitude \emph{outliers} can be difficult to represent alongside typical values~\citep{dettmers2022llmint8}, while \emph{saliency} describes how strongly representation errors affect computation or loss. PTQ methods address these differences: SmoothQuant rescales activation and weight channels~\citep{xiao2023smoothquant}, and AWQ protects salient weight channels through scaling~\citep{lin2024awq}. SqueezeLLM instead combines sensitivity-weighted codebooks with higher-precision storage for selected sensitive weights and outliers~\citep{kim2024squeezellm}.

\subsection{Dynamic Precision Refinement}
\label{sec:selection-motivation}

These methods account for weight sensitivity when constructing the quantized representation, but that representation remains fixed during decoding. Input activations change with each token, shifting which weight channels contribute most to the projection error. Let $W$ be the original floating-point weights, $\widehat W$ their dequantized approximation, and $R=W-\widehat W$ the weight residual. For input $x$, each residual column contributes $x_iR_{:,i}$ to the projection error. Large activation magnitudes amplify these contributions, making $|x_i|$ a proxy for which channels to refine:
\begin{equation}
 (W-\widehat W)x=Rx=\sum_i x_iR_{:,i}.
\label{eq:projection-error}
\end{equation}

DecDEC~\citep{park2025decdec} finds that calibration-selected channels often differ from those with the largest activation magnitudes during decoding. At each step, it selects the $k$ largest $|x_i|$, fetches their 4-bit residual columns from host memory, and adds their products with signed $x_i$ to the quantized projection's output. During single-sequence decoding, the memory-bound projection leaves GPU cores available for concurrent refinement without slowing the projection. Fetching awaits the projection's input, and refinement must finish before its output is consumed to avoid stalls, confining overlap to that projection's execution window.

We now turn to the KV cache, where the tokens receiving the most attention can likewise change between decoding steps. Attention compares a \emph{query}, a vector derived from the current token's representation, with cached keys to weigh the cached values. Top-$k$ attention and Quest exploit this dependence to restrict attention to selected tokens~\citep{gupta2021topk,tang2024quest}. SpeCache instead retains a full low-bit cache and substitutes selected 16-bit entries fetched from host memory~\citep{jie2025specache}. It uses a speculative token to prefetch selected entries for the next decoding step while current-step computation continues. However, fetching complete 16-bit replacements overlooks the opportunity to jointly design refinement encoding and attention computation to reduce transfer and execution costs.

\subsection{Bandwidth and Latency Constraints}
\label{sec:precision-delivery}
\label{sec:latency-motivation}

\begin{figure}[!htb]
\centering
\includegraphics[width=0.805\columnwidth]{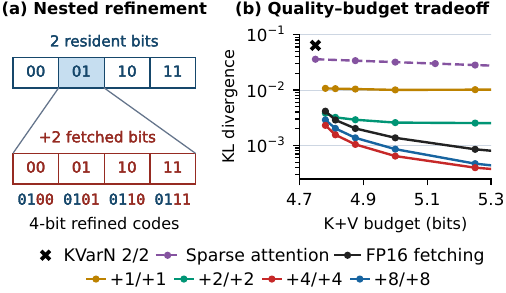}
\caption{(a) Nested refinement subdivides GPU-resident quantization bins with fetched bits. (b) KL divergence from BF16 Llama-3.1-8B-Instruct on PG-19 at matched K+V budgets, including resident, fetched, and metadata bits per K/V pair. K/V bitwidth labels give resident bits per key/value element, respectively. All schemes use KVarN 2/2 and select tokens by attention weights from resident low-bit keys (Equation~\ref{eq:kv-token-score}). Sparse attention uses selected FP16 entries alone; FP16 fetching and nested refinement retain unselected low-bit entries. FP16 fetching replaces selected entries; nested $+b/+b$ adds $b$ bits per key/value element.}
\label{fig:dense-sparse-attention}
\end{figure}

Sparse attention reduces KV traffic by omitting unselected tokens from attention. SpeCache preserves their contributions through a low-bit cache and fetches complete 16-bit floating-point (FP16) entries for selected tokens. We instead keep every key and value in low bits on the GPU and, for selected tokens, fetch only the additional bits that refine those resident values, adapting the \emph{nested quantization} principle used for weights in Any-Precision LLM~\citep{park2024anyprecision}: the fetched bits pick a finer value inside the interval that the resident low-bit value already represents. Figure~\ref{fig:dense-sparse-attention} compares these three representation choices at matched K+V budgets. Nested refinement with 4 or 8 additional bits per selected key/value element yields lower Kullback--Leibler (KL) divergence than both sparse attention and complete FP16 replacement.

\begin{figure}[!htb]
\centering
\includegraphics[width=0.805\columnwidth]{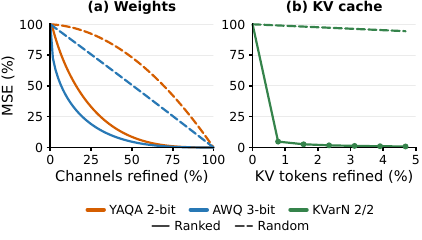}
\caption{Ranked versus random selection in Llama-3.1-8B-Instruct. (a) With WikiText-2 inputs, weight channels are ranked by input activation magnitude and refined with corresponding columns of $R$. (b) For PG-19 inputs, KV tokens are ranked by attention weights from resident low-bit keys and restored to FP16. Output mean squared error (MSE) against FP16 is normalized to the error without refinement, then averaged over tokens and Transformer blocks, across projections in (a) and KV heads in (b).}
\label{fig:error-concentration}
\end{figure}

Encoding determines how much data each refinement requires; selection determines which errors it addresses. Figure~\ref{fig:error-concentration} isolates selection by adding selected channels' residual contributions or restoring selected KV entries to FP16. We rank weight channels by input-activation magnitude and KV tokens by attention weights from resident low-bit keys. Refining only a small top-ranked fraction removes a disproportionate share of output mean squared error. Activation-magnitude-based selection remains effective even on YAQA, whose rounding compensates for quantization errors across input and output channels~\citep{tseng2026yaqa}.

Compact representations and selective refinement reduce traffic, but do not eliminate access latency. DecDEC's overlap window shrinks as lower-bit quantization and higher GPU bandwidth accelerate the underlying projection~\mbox{\citep{park2025decdec}}. On DGX Spark, isolated 2~KB random SSD reads have a median latency of 200~$\mu$s (709~$\mu$s at the 99th percentile)---far longer than the 21~$\mu$s average for Llama-3.1-8B-Instruct's YAQA 2-bit output projection, making long stalls unavoidable with projection-local overlap.

Two prior mechanisms suggest a solution. Each Transformer block’s attention and MLP outputs are added to a running activation vector, the residual stream. Deja Vu finds that this vector's direction changes slowly across neighboring OPT blocks and uses earlier activations to predict which attention heads and MLP neurons to compute later~\mbox{\citep{liu2023dejavu}}. KTransformers' Expert Deferral adds the outputs of CPU-computed experts in mixture-of-experts (MoE) models to the residual stream one block later, so that their computation overlaps with the GPU’s work~\citep{chen2025ktransformers}.

\begin{figure*}[!t]
\centering
\includegraphics[width=\textwidth]{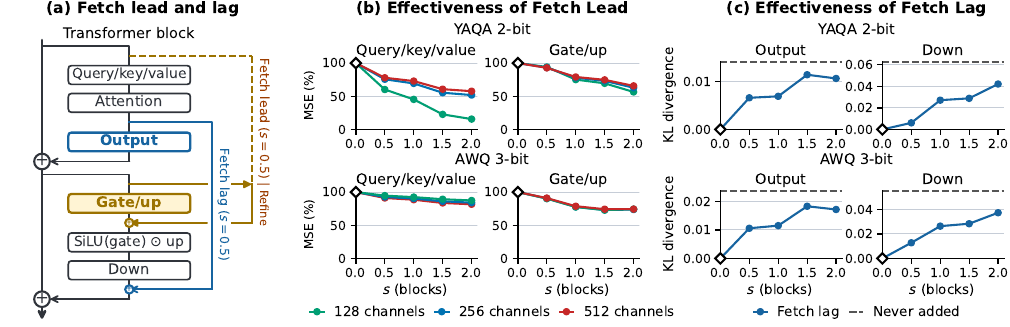}
\caption{Fetch lead/lag in Llama-3.1-8B-Instruct on WikiText-2. Shifts $s$ are measured in Transformer blocks; $s=0$ means no shift. (a) Gate/up fetch lead and attention-output fetch lag at $s=0.5$. The dashed gold path shows early selection; the solid gold input arrow feeds the later refinement step. (b) Projection-output MSE reduction retained by fetch lead, relative to selection at the projection's input (100\%), using errors summed over tokens and blocks. (c) Fetch-lag refinement is applied to all input channels to isolate the effect of delayed addition. Only attention-output or down projections are quantized; KL to FP16 is averaged over predicted tokens across all sequences.}
\label{fig:residual-stream-similarity}
\end{figure*}

These mechanisms extend to weight refinement (Section~\ref{sec:selection-motivation}).  Query/key/value and gate/up projections read from the residual stream, allowing earlier activations to predict channels and initiate fetching before those projections execute (\emph{fetch lead}). Refinement is executed once each projection's input becomes available. Attention-output and down projections write to the stream, allowing refinement to be deferred to a later stream update (\emph{fetch lag}). Together, these cases expose overlap opportunities for every attention and MLP projection, extending refinement beyond the execution window of a single projection (Figure~\ref{fig:residual-stream-similarity}(a)).

We evaluate the shifts separately: fetch lead tests how well earlier activations predict which channels to refine, while fetch lag tests how delaying refinement addition affects model accuracy. Earlier selection can preserve most of the projection-output mean squared error (MSE) reduction at small shifts (Figure~\ref{fig:residual-stream-similarity}(b)); delayed additions continue to reduce KL divergence relative to omitting refinement (Figure~\ref{fig:residual-stream-similarity}(c)). The benefit of refinement thus persists beyond the original projection's execution window. In Section~\ref{sec:end-to-end}, \methodname{} combines fetch lead and lag across Transformer blocks to overlap refinement transfers and refinement computation with intervening model computation.

\begin{figure}[!t]
\centering
\includegraphics[width=\linewidth]{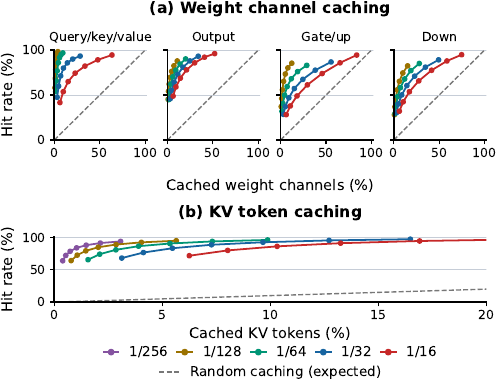}
\caption{Reuse across decoding steps in Llama-3.1-8B-Instruct on PG-19. Selections rank (a) weight channels by input activation magnitude and (b) KV tokens by attention weight (Equation~\ref{eq:kv-token-score}); legend fractions give the share selected per step. Caches contain the union of selections from the preceding 1, 2, 4, 8, 16, 32, or 64 steps. Axes show the cached share of all entries against hit rate, averaged over decoding steps and Transformer blocks. Dashed lines show expected hit rates under equally likely random caching.}
\label{fig:cross-decode-reuse}
\end{figure}

\begin{figure*}[t]
\centering
\includegraphics[width=\textwidth]{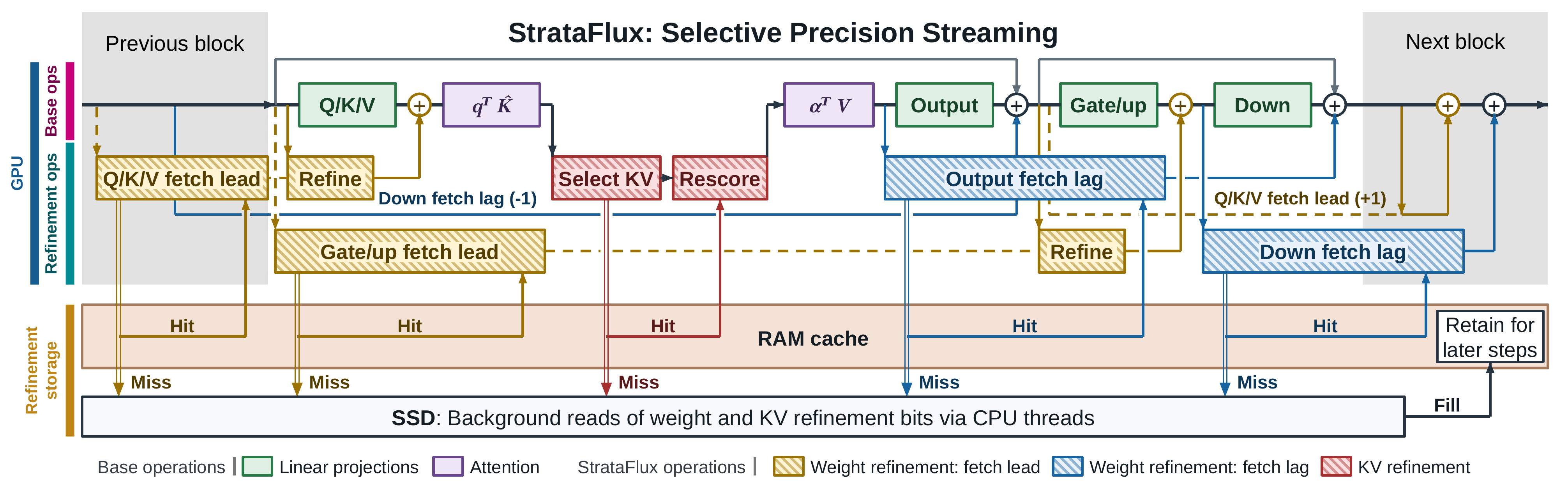}
\caption{\methodname{}\textquotesingle{}s selective precision streaming in action. Fetch lead/lag overlaps weight refinement with model computation; KV refinement runs within attention. Labeled lines replace refinement boxes for previous ($-1$) and next ($+1$) Transformer blocks. In this unified-memory configuration, the GPU reads the RAM cache directly; misses retain resident low-bit precision while background SSD reads supply later decoding steps. Dedicated GPUs fetch directly from host RAM, without SSD. Some Transformer operations omitted.}
\label{fig:system-overview}
\end{figure*}

An independent opportunity arises from repeated selections across consecutive decoding steps. Figure~\ref{fig:cross-decode-reuse} shows that caching only a fraction of all weight channels or KV tokens can yield very high hit rates. Caching refinement bits for recurring selections avoids repeated transfers. Data arriving too late for one step can also serve subsequent selections. We exploit these two benefits to reduce transfers and waits for weight and KV refinement (Section~\ref{sec:reuse}).

\section{\methodname{}: Selective Precision Streaming}
\label{sec:method}

\methodname{}'s selective precision streaming expands the precision available to inference beyond what fits in fast memory or can be fetched within a single projection. Figure~\ref{fig:system-overview} outlines its weight and KV computation paths and their shared refinement-cache policy. We tailor refinement delivery to both dedicated-GPU and unified-memory platforms.

\subsection{Distributing Precision Across the Memory Hierarchy}
\label{sec:overview}

On dedicated-GPU platforms, GPU memory holds quantized weights and a compact KV cache covering all tokens, while the GPU fetches refinement bits directly from pinned host RAM. On unified-memory platforms, the CPU and GPU share RAM holding these compact representations, while SSD stores refinement bits that we load into shared RAM for GPU access (Section~\ref{sec:reuse}). These relatively inexpensive backing tiers let \methodname{} use capacity that would otherwise remain underutilized during LLM inference.

\subsection{Designing Precision for Selective Refinement}
\label{sec:partial-refinement}

\subsubsection{Weight Refinement}
\label{sec:weight-streaming}

\methodname{} retains the resident weight quantizer's format and projection kernels. We store quantized weight-refinement columns as refinement bits; $C$ is the corresponding matrix of dequantized values. For selected input-channel indices $S(x)$, we add the selected columns' contribution to the base quantized projection:
\begin{equation}
 y'=\widehat W x+\sum_{i\in S(x)}x_iC_{:,i}.
\label{eq:weight-correction}
\end{equation}
\textbf{Channel selection.} Following DecDEC~\citep{park2025decdec}, we partition each projection's input channels into chunks of 1,024 and select $k$ channels per chunk, targeting the largest $|x_i|$. These indices form $S(x)$; each selected column's refinement bits are fetched contiguously. Separate per-chunk budgets $k_{\mathrm{qkv}}$, $k_{\mathrm{o}}$, $k_{\mathrm{gu}}$, and $k_{\mathrm{d}}$ govern four projection sets: query/key/value, attention-output, gate/up, and down, respectively. We encode refinement values as integers with separate scale factors, giving the refinement kernel a common format independent of the resident weight quantizer.

\textbf{Fitting refinement values.} DecDEC quantizes the weight residual $R$ (Section~\ref{sec:selection-motivation}). \methodname{} uses \emph{refinement fitting} to adjust refinement values so selected columns can compensate for omitted contributions. Let $z_k(x)$ zero all but the $k$ largest-magnitude activations per input chunk. We minimize projection error while penalizing deviations from $R$ with strength $\lambda$:
\begin{equation}
 C_k^*=\arg\min_C\;\mathrm{E}_x[\|Rx-Cz_k(x)\|_2^2]+\lambda\|C-R\|_F^2.
\label{eq:fitted-residual}
\end{equation}
The minimizer of Equation~\ref{eq:fitted-residual} solves Equation~\ref{eq:residual-fit-solve}, where $M_k=\mathrm{E}_x[z_k(x)z_k(x)^\top]$, $G_k=\mathrm{E}_x[xz_k(x)^\top]$, and $I$ is the identity matrix. We estimate the expectations on a small RedPajama calibration set~\citep{weber2024redpajama} and use the mean diagonal of $M_k$ as the default $\lambda$:
\begin{equation}
 C_k^*(M_k+\lambda I)=RG_k+\lambda R.
\label{eq:residual-fit-solve}
\end{equation}
We quantize the fitted matrix to 4 bits before inference, following DecDEC's 4-bit encoding of weight residuals.

\textbf{Neuron-wise alternative.} For input $x$, the quantized MLP combines gate and up outputs into neuron activations $h$, then computes its unrefined output $y_{\mathrm q}$:
\begin{equation}
\begin{array}{r@{}l@{\qquad}r@{}l}
g&=\widehat W_{\mathrm{gate}}x, & u&=\widehat W_{\mathrm{up}}x,\\
h&=\phi(g)\odot u, & y_{\mathrm q}&=\widehat W_{\mathrm{down}}h.
\end{array}
\label{eq:quantized-mlp}
\end{equation}
Here $\phi$ is the gate's activation function, $\odot$ is elementwise multiplication, and $h_j$ is neuron $j$'s activation after gating. Channel-wise refinement targets large input activations separately: $x$ for gate/up and $h$ for down. Neuron-wise refinement instead selects the $k_{\mathrm n}$ neurons with largest $|h_j|$ in each chunk of 1,024 and refines their gate/up rows and down columns together. Each selected neuron's \emph{MLP neuron bundle} encodes gate/up refinement rows $c_j^{\mathrm{g}},c_j^{\mathrm{u}}$, a down refinement column $c_j^{\mathrm{d}}$, and a 4-bit copy $d_j$ of the BF16 down column. For selected neuron indices $N(h)$, we recompute $h_j$ from Equation~\ref{eq:quantized-mlp} as $h'_j$. As in Equation~\ref{eq:weight-correction}, the resulting output refinement is added to $y_{\mathrm q}$:
\begin{equation}
\begin{array}{r@{}l}
 h'_j&\displaystyle=\phi(g_j+c_j^{\mathrm{g}}x)(u_j+c_j^{\mathrm{u}}x),\\
 y'&\displaystyle=y_{\mathrm q}+\sum_{j\in N(h)}\bigl[
 d_j(h'_j-h_j)+c_j^{\mathrm{d}}h_j\bigr].
\end{array}
\label{eq:neuron-correction}
\end{equation}
Neither channel-wise nor neuron-wise MLP refinement consistently offers the better speed--accuracy tradeoff across models, quantizers, platforms, and speed budgets. We therefore retain both options for \methodname{}.

\subsubsection{KV Refinement}
\label{sec:kv-streaming}

For KV refinement, we tailor nested reconstruction to the resident cache format. Our main KV implementation, \methodname{}-KV, adopts KVarN's~\citep{muller2026kvarn} quantization format for its balance of accuracy and memory footprint. We co-design its low-bit attention kernels and selective refinement. For adaptations of \methodname{} to other KV-cache formats, see Appendix~\ref{app-sec:appendix-kv-bases}.

\textbf{Nested encoding.} As illustrated in Figure~\ref{fig:dense-sparse-attention}(a), we subdivide each KVarN code's quantization interval into equal sub-intervals and store their indices as refinement bits in the backing tier. The resident code and fetched index identify a finer reconstruction, independently for keys and values.

\textbf{Token selection.} We compute attention weights by applying softmax to scores between the current query and resident low-bit keys. A KV head stores keys and values shared by one or more query heads. Let $\alpha_{h,i}$ denote the resulting attention weight assigned to token $i$ by query head $h$, and let $H(j)$ be the query heads sharing KV head $j$. For token $i$ and KV head $j$, we define the selection score
\begin{equation}
 s_j(i)=\max_{h\in H(j)}\alpha_{h,i}.
\label{eq:kv-token-score}
\end{equation}
This score captures the strongest attention to the token among query heads sharing its keys and values. For each KV head, we select a configurable number of the highest-scoring tokens for refinement.

\textbf{Attention refinement.} Combining fetched bits with resident codes reconstructs selected tokens' higher-precision keys and values (Figure~\ref{fig:system-overview}). For one query head, resident scores $\ell_i$ and values $v_i$ give output $o$ and normalizer $Z=\sum_i e^{\ell_i}$. Refining tokens $S$ gives scores $\ell_i'$, values $v_i'$, and output
\begin{equation}
o'=\frac{Zo+\sum_{i\in S}\bigl(e^{\ell_i'}v_i'-e^{\ell_i}v_i\bigr)}
{Z+\sum_{i\in S}\bigl(e^{\ell_i'}-e^{\ell_i}\bigr)}.
\label{eq:kv-refined-attention}
\end{equation}

In practice, we group consecutive tokens within each KV head and pack each group's nested refinement bits contiguously for larger reads. We score token groups using Equation~\ref{eq:kv-token-score}, with each query head's attention weights summed over the group. Because token selection depends on the initial attention scores and the weighted sum depends on refined scores, KV refinement adds a region that cannot overlap the surrounding attention computation. We limit this overhead through contiguous fetches and fusion of value refinement into the weighted sum.

\textbf{Producing KV refinement bits.} During prefill, we compute attention on uncompressed keys and values, then encode resident low-bit codes and offload refinement bits. During decoding, we encode completed blocks of new KV entries in the same format.

\subsection{Extending the Weight Delivery Interval}
\label{sec:end-to-end}

The runtime uses fetch lead/lag to decouple refinement fetching from its application, hiding latency through overlap with base operations (Section~\ref{sec:latency-motivation}). Appendix~\ref{app-sec:concurrent-weight-refinement} explains the CUDA stream and event scheduling.

\textbf{Fetch lead.} For each query/key/value or gate/up projection, we apply its normalization to a residual-stream activation $s$ Transformer blocks earlier, then select channels and initiate fetching. Once the projection's input is available, the refinement kernel multiplies fetched columns by corresponding activations and adds the summed products to the quantized projection's output (Figure~\ref{fig:system-overview}).

\textbf{Fetch lag.} We add attention-output and down refinements to the residual stream $s$ Transformer blocks after their original addition point, overlapping their fetching and computation with intervening projections (Figure~\ref{fig:system-overview}). Neuron-wise MLP refinement likewise supports fetch lag.

\begin{figure}[!htb]
\centering
\includegraphics[width=\linewidth]{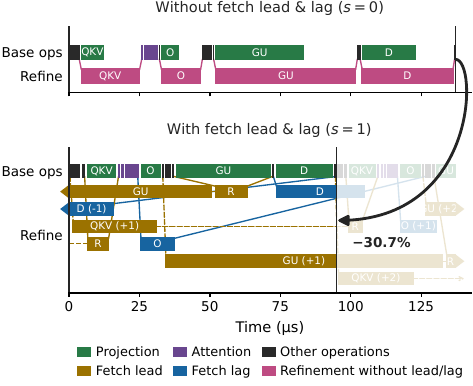}
\caption{Nsight Systems kernel traces comparing refinement without fetch lead/lag ($s=0$) and with fetch lead/lag ($s=1$) on Llama-3.1-8B-Instruct (AWQ 3-bit, RTX~5090, $k=32$). QKV/GU/O/D denote query/key/value, gate/up, output, and down projections; R denotes refinement after fetch lead. Refinement for the previous and next blocks are marked with an offset (-1 or +1). Dashed lines show fetch-lead's early selection; solid lines show activation reads and writes for refinement.}
\label{fig:lead-lag-timeline}
\end{figure}

The Nsight Systems kernel traces of our runtime system in Figure~\ref{fig:lead-lag-timeline} confirm the cross-block overlap illustrated in Figure~\ref{fig:system-overview}. With the projection-local schedule used by DecDEC ($s=0$), the main stream waits for refinement after each projection, delaying subsequent operations in the Transformer block's computation graph. Fetch lead/lag at $s=1$ successfully overlaps refinement with computation across neighboring Transformer blocks, reducing these waits and shortening block execution at matched channel count $k$.

\begin{figure}[!htb]
\centering
\includegraphics[width=0.805\columnwidth]{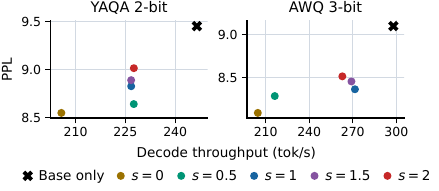}
\caption{Perplexity--throughput tradeoffs on Llama-3.1-8B-Instruct with $k=32$ and a common fetch lead/lag distance $s$ across all four projection sets. PPL uses WikiText-2; throughput uses PG-19 on RTX~5090 with refinement bits in host RAM.}
\label{fig:lead-lag-distance}
\end{figure}

However, the shift needed to hide refinement depends on the quantization method. At $k=32$, Figure~\ref{fig:lead-lag-distance} shows that slower YAQA 2-bit obtains most of its throughput gain at $s=0.5$, whereas faster AWQ 3-bit needs $s=1$ because its shorter projections leave less time to hide refinement. Furthermore, these sweeps apply the same $s$ to all four projection sets within each method, leaving their individual speed--accuracy tradeoffs unexplored. In Section~\ref{sec:tuning}, we introduce a tuner that directly explores this search space to select $s$ and $k$ per projection set under a configurable speed budget.

\subsection{Reusing Precision Across Decoding Steps}
\label{sec:reuse}

For dedicated-GPU platforms, host RAM's lower access latency than SSD allows refinement bits to be fetched directly without an intermediate cache. For unified-memory systems, however, SSD's higher access latency motivates caching fetched refinement bits in GPU-accessible RAM.

\textbf{Caching fetched precision.} Weight and KV selections recur across consecutive decoding steps (Figure~\ref{fig:cross-decode-reuse}). To avoid repeated SSD reads and their latency, we retain fetched refinement bits in a configurably sized RAM cache that the GPU reads directly (Figure~\ref{fig:system-overview}).

\textbf{Miss policy.} Selections can still miss the cache. For weights and KV, the current step can decide to skip missing refinement and use the resident low-bit representation. A CPU loader fetches the bits into the cache for future selections, keeping the transfer useful without making the requesting step wait. Alternatively, the runtime can wait and apply refinement in the current step, trading speed for accuracy.

\textbf{Cache organization.} Weight and KV refinement have separate cache budgets. The weight cache stores refinement bits for all four projection sets and uses least-recently-used replacement. KV token groups share capacity across all KV heads and Transformer blocks.

\textbf{Populating the KV refinement cache.} During prefill, we write KV refinement bits to SSD and cache copies for the most recently encoded token groups in RAM. During decoding, newly produced refinement bits enter the cache directly, serving subsequent selections without SSD readback.

\subsection{Tuning Weight Refinement Within a Speed Budget}
\label{sec:tuning}

\textbf{Search space.} The weight tuner selects a configuration subject to a user-specified slowdown limit relative to the quantized base. It chooses channel budgets $k$, fetch lead/lag distances $s$ for each projection set, and whether the MLP uses channel-wise refinement or neuron-wise refinement with budget $k_{\mathrm n}$. Following DecDEC~\citep{park2025decdec}, it also tunes the number of GPU thread blocks (groups of cooperating threads) executing refinement. More thread blocks can accelerate refinement but also slow the concurrent quantized projections by competing for GPU resources.

\textbf{Candidate selection.} For each combination of shift distances and channel-wise or neuron-wise MLP refinement, the tuner searches for the largest common channel budget $k$ that meets the slowdown limit in end-to-end decoding. It selects the retained candidate with the lowest KL divergence from the BF16 reference on held-out text, then increases individual channel budgets or the neuron budget while the measured slowdown remains within the limit. We tune each model, weight quantizer, and platform separately. We then fit refinements at the chosen selection budgets, as detailed in Section~\ref{sec:weight-streaming}.

\section{Evaluation}
\label{sec:evaluation}

We implement \methodname{} as an end-to-end system in llama.cpp~\citep{llamacpp}, with CUDA and Metal backends. We evaluate \methodname{} on Llama-3.1-8B-Instruct~\citep{metallama31}, Llama-3.3-70B-Instruct~\citep{metallama33}, Qwen3-8B, Qwen3-32B, Qwen3-30B-A3B~\citep{qwen3models}, Qwen3.5-9B~\citep{qwen35model}, and DeepSeek-V2-Lite~\citep{deepseekv2lite}. Platforms span dedicated GPUs (RTX 5090, RTX 5070 Ti, and RTX PRO 6000) and unified-memory platforms (DGX Spark and MacBook Air M3). Appendix~\ref{app-sec:appendix-mac} reports the Mac evaluation. We measure perplexity and KL divergence from BF16 on 2K-token WikiText-2 windows~\citep{merity2017pointer} and choose PG-19~\citep{rae2020compressive} for long-context perplexity. We measure accuracy on RULER~\citep{hsieh2024ruler}, MATH500~\citep{lightman2024verify}, and AIME 2024/2025~\citep{maaaime}; AIME averages eight samples per problem.  Timings use llama.cpp except for methods with their own runtimes. Appendix~\ref{app-sec:appendix-provenance} details protocols and implementations.

In tables and figures, \methodname{}-W and \methodname{}-KV isolate weight and KV refinement, respectively; \methodname{} without a suffix denotes the full system with both enabled.

Unless ablated, DGX Spark uses separate 128 MiB RAM caches for weight and KV refinement. At the small slowdown budgets used in our main results, we find that skipping cache misses gives a better latency--accuracy tradeoff on unified-memory platforms. There, we skip misses and set $s=0$: no SSD wait remains for fetch lead/lag to hide. Appendix~\ref{app-sec:appendix-weights} evaluates fetch lead/lag with waiting on cache misses at larger budgets.

\subsection{Weight Refinement}
\label{sec:eval-weights}
We compare \methodname{}-W with DecDEC~\citep{park2025decdec} on YAQA~\citep{tseng2026yaqa} with QTIP's format and kernels~\citep{tseng2024qtip}, LNQ with GuidedQuant (GQ)~\citep{kim2025guidedquant}, AWQ~\citep{lin2024awq}, and SqueezeLLM (SqLLM)~\citep{kim2024squeezellm} bases.
\begin{table}[!ht]
\centering
\caption{Weight-only refinement: WikiText-2 perplexity and KL divergence with 2--3-bit weights. Platform blocks report \methodname{}-W, alongside DecDEC on RTX 5090, tuned to a 5\% slowdown budget. DecDEC does not support SSD fetching on DGX Spark. Bold marks the better score within each RTX 5090 pair.}
\label{tab:main_weight_results-1}
\label{tab:main_weight_results-spark}
\fontsize{7}{8.4}\selectfont
\setlength{\tabcolsep}{1pt}
\setlength{\aboverulesep}{0.25ex}
\setlength{\belowrulesep}{0.35ex}
\renewcommand{\arraystretch}{1.00}\setlength{\aboverulesep}{.2ex}\setlength{\belowrulesep}{.25ex}
\begin{tabular*}{\columnwidth}{@{}l@{\hspace{2pt}}l@{\extracolsep{\fill}}lcccc@{}}
\toprule
 & & \textbf{Refinement} & \multicolumn{2}{c}{\textbf{L3.1-8B-I}} & \multicolumn{2}{c}{\textbf{Qwen3-8B}} \\
\cmidrule(lr){4-5}\cmidrule(lr){6-7}
 & \textbf{Weights} & \textbf{method} & \textbf{PPL} & \textbf{KL} & \textbf{PPL} & \textbf{KL} \\
\midrule[\heavyrulewidth]
 & BF16 & \multicolumn{1}{c}{--} & 7.22 & 0.000 & 9.73 & 0.000 \\
\midrule[\heavyrulewidth]
\multirow[c]{5}{*}[-\dimexpr0\aboverulesep+0\belowrulesep+0\cmidrulewidth\relax]{\textbf{Base}} & YAQA 2-bit & \multicolumn{1}{c}{\multirow[c]{5}{*}{--}} & 9.46 & 0.257 & 10.95 & 0.214 \\
 & YAQA 3-bit &  & 7.61 & 0.047 & 9.98 & 0.052 \\
 & LNQ+GQ 3-bit &  & 8.52 & 0.157 & 10.92 & 0.184 \\
 & AWQ 3-bit &  & 9.10 & 0.229 & 11.88 & 0.284 \\
 & SqLLM 3-bit &  & 9.36 & 0.263 & 11.71 & 0.255 \\
\midrule[\heavyrulewidth]
\multirow[c]{5}{*}[-\dimexpr0\aboverulesep+0\belowrulesep+0\cmidrulewidth\relax]{\textbf{DGX Spark}} & YAQA 2-bit & \methodname{}-W & 8.94 & 0.200 & 10.77 & 0.175 \\
 & YAQA 3-bit & \methodname{}-W & 7.55 & 0.041 & 9.89 & 0.043 \\
 & LNQ+GQ 3-bit & \methodname{}-W & 8.29 & 0.135 & 10.70 & 0.159 \\
 & AWQ 3-bit & \methodname{}-W & 8.64 & 0.179 & 11.30 & 0.222 \\
 & SqLLM 3-bit & \methodname{}-W & 8.56 & 0.174 & 11.03 & 0.201 \\
\midrule[\heavyrulewidth]
\multirow[c]{10}{*}[-\dimexpr2\aboverulesep+2\belowrulesep+2\cmidrulewidth\relax]{\textbf{RTX 5090}} & \multirow[c]{2}{*}{YAQA 2-bit} & DecDEC & 9.03 & 0.211 & 10.69 & 0.184 \\
 &  & \methodname{}-W & \textbf{8.40} & \textbf{0.139} & \textbf{10.51} & \textbf{0.132} \\
\cmidrule{2-7}
 & \multirow[c]{2}{*}{YAQA 3-bit} & DecDEC & 7.55 & 0.040 & \textbf{9.89} & 0.044 \\
 &  & \methodname{}-W & \textbf{7.49} & \textbf{0.033} & \textbf{9.89} & \textbf{0.033} \\
\cmidrule{2-7}
 & \multirow[c]{2}{*}{LNQ+GQ 3-bit} & DecDEC & 8.37 & 0.142 & 10.66 & 0.170 \\
 &  & \methodname{}-W & \textbf{8.18} & \textbf{0.121} & \textbf{10.47} & \textbf{0.139} \\
\cmidrule{2-7}
 & \multirow[c]{2}{*}{AWQ 3-bit} & DecDEC & 8.80 & 0.195 & 11.38 & 0.231 \\
 &  & \methodname{}-W & \textbf{8.38} & \textbf{0.145} & \textbf{10.71} & \textbf{0.167} \\
\cmidrule{2-7}
 & \multirow[c]{2}{*}{SqLLM 3-bit} & DecDEC & 9.02 & 0.225 & 11.36 & 0.215 \\
 &  & \methodname{}-W & \textbf{8.42} & \textbf{0.156} & \textbf{10.81} & \textbf{0.169} \\
\bottomrule
\end{tabular*}
\end{table}

\begin{table}[t]
\centering
\caption{Weight-only refinement: Qwen3-8B MATH500/AIME accuracy. Platform columns use \methodname{}-W\textquotesingle{}s respective 5\% slowdown configurations. 4-bit results appear in Appendix~\ref{app-sec:appendix-weights}.}
\label{tab:weight-downstream}
\fontsize{7}{8.4}\selectfont
\setlength{\tabcolsep}{1pt}
\renewcommand{\arraystretch}{1.00}\setlength{\aboverulesep}{.2ex}\setlength{\belowrulesep}{.25ex}
\setlength{\aboverulesep}{0.25ex}
\setlength{\belowrulesep}{0.35ex}
\begin{tabular*}{\columnwidth}{@{}l@{\extracolsep{\fill}}cc@{\hspace{5pt}}cc@{\hspace{5pt}}cc@{}}
\toprule
 & \multicolumn{2}{c}{\textbf{Base}} & \multicolumn{2}{c}{\begin{tabular}[b]{@{}c@{}}\textbf{\methodname{}-W}\\\textbf{DGX Spark}\end{tabular}} & \multicolumn{2}{c}{\begin{tabular}[b]{@{}c@{}}\textbf{\methodname{}-W}\\\textbf{RTX 5090}\end{tabular}} \\
\cmidrule(lr){2-3}\cmidrule(lr){4-5}\cmidrule(lr){6-7}
\textbf{Weights} & \textbf{M500} & \textbf{AIME} & \textbf{M500} & \textbf{AIME} & \textbf{M500} & \textbf{AIME} \\
\midrule
BF16 & 97.2 & 71.0 & -- & -- & -- & -- \\
\midrule
YAQA 2-bit & 77.2 & 19.2 & 89.4 & 40.0 & 92.4 & 46.7 \\
YAQA 3-bit & 95.0 & 66.7 & 95.0 & 66.0 & 95.2 & 70.0 \\
LNQ+GQ 3-bit & 91.8 & 39.4 & 93.0 & 47.3 & 93.2 & 49.2 \\
AWQ 3-bit & 91.8 & 46.5 & 92.6 & 51.9 & 93.8 & 51.0 \\
SqLLM 3-bit & 93.0 & 47.3 & 94.2 & 48.3 & 93.0 & 50.8 \\
\bottomrule
\end{tabular*}
\end{table}

\begin{figure}[!htb]
\centering
\includegraphics[width=\columnwidth]{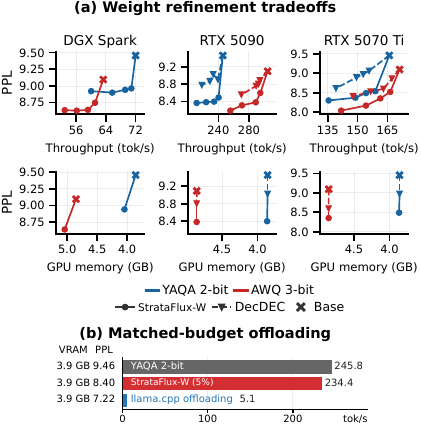}
\caption{Weight-refinement tradeoffs on Llama-3.1-8B-Instruct on WikiText-2. (a) Throughput plots connect configurations tuned for slowdown budgets of 2.5\% / 5\% / 10\% / 20\%; memory plots show the configurations tuned for the 5\% budget. DGX Spark memory includes the RAM cache. DecDEC does not support SSD fetching on Spark. (b) \methodname{}-W tuned for 5\% slowdown on YAQA 2-bit base, and llama.cpp BF16 partial offloading at a matched 3.9 GB GPU memory budget, RTX 5090.}
\label{fig:speed-accuracy}
\label{fig:weight-offloading}
\end{figure}

We use a 5\% slowdown budget for main weight comparisons because it is sufficient for large accuracy gains (Figure~\ref{fig:speed-accuracy}(a)). At matched budgets, \methodname{}-W outperforms DecDEC across quantizers (Table~\ref{tab:main_weight_results-1}). These gains extend to reasoning tasks (Table~\ref{tab:weight-downstream}) while retaining quantized speed and low GPU memory use (Figure~\ref{fig:speed-accuracy}). For Qwen3-8B with YAQA 2-bit weights, AIME accuracy rises from 19.2\% to 46.7\%. Appendix~\ref{app-sec:appendix-weights} gives full 2--4-bit, larger-model, and downstream results; Appendix~\ref{app-sec:weight-tuning-configurations} lists tuned configurations.

\subsection{KV Refinement}
\label{sec:eval-kv}
KV baselines include KVarN~\citep{muller2026kvarn}, KIVI~\citep{liu2024kivi}, KVQuant~\citep{hooper2024kvquant}, TurboQuant~\citep{zandieh2026turboquant}, and AQUA-KV~\citep{shutova2025aquakv}; offloading comparisons include ShadowKV~\citep{sun2025shadowkv}, ParisKV~\citep{qi2026pariskv}, TailorKV~\citep{yao2025tailorkv}, FreeKV~\citep{liu2026freekv}, SpeCache~\citep{jie2025specache}, and KVSwap~\citep{zhang2026kvswap}. \methodname{} is configured to select $1/128$ ($0.78\%$) of token groups for refinement, adding 4 bits per selected key and value element (+4/+4). Groups contain two consecutive tokens on dedicated GPU platforms and four on unified-memory platforms. For other refinement settings, see Appendix~\ref{app-sec:kv-settings-sweep}; for resident 2/1 KV, see Appendix~\ref{app-sec:kv-two-one}.

\begin{table}[!t]
\centering
\caption{KV-only refinement: PG-19 perplexity/KL and RULER accuracy at 128K, plus MATH500/AIME accuracy, with BF16 weights. Platform blocks report \methodname{}-KV.}
\label{tab:main-kv-comparison}
\fontsize{7}{8.4}\selectfont
\setlength{\tabcolsep}{0.7pt}
\setlength{\aboverulesep}{0.3ex}
\setlength{\belowrulesep}{0.4ex}
\renewcommand{\arraystretch}{1.0}\setlength{\aboverulesep}{.2ex}\setlength{\belowrulesep}{.25ex}
\begin{tabular*}{\columnwidth}{@{\extracolsep{\fill}}@{}llcccccccc@{}}
\toprule
\multicolumn{2}{l}{} & \multicolumn{3}{c}{\textbf{L3.1-8B-I}} & \multicolumn{5}{c}{\textbf{Qwen3-8B}} \\
\cmidrule(lr){3-5}\cmidrule(lr){6-10}
 & \textbf{KV} & \textbf{PPL} & \textbf{KL} & \textbf{R128} & \textbf{PPL} & \textbf{KL} & \textbf{R128} & \textbf{M500} & \textbf{AIME} \\
\midrule[\heavyrulewidth]
 & BF16 & 10.32 & 0.000 & 77.9 & 12.89 & 0.000 & 74.1 & 97.2 & 71.0 \\
\midrule[\heavyrulewidth]
\multirow[c]{3}{*}{\textbf{Base}} & 2/2 & 10.98 & 0.067 & 68.6 & 14.40 & 0.130 & 66.0 & 96.2 & 66.0 \\
 & 3/2 & 10.45 & 0.014 & 76.2 & 13.17 & 0.026 & 72.3 & 96.6 & 69.2 \\
 & 3/3 & 10.39 & 0.008 & 77.2 & 13.02 & 0.016 & 72.9 & 95.8 & 70.6 \\
\midrule[\heavyrulewidth]
\multirow[c]{3}{*}{\shortstack[l]{\textbf{\methodname{}-KV}\\\textbf{DGX Spark}}} & 2/2 & 10.47 & 0.011 & 75.2 & 13.32 & 0.030 & 71.4 & 96.0 & 68.3 \\
 & 3/2 & 10.34 & 0.002 & 77.6 & 12.98 & 0.005 & 74.0 & 95.0 & 70.2 \\
 & 3/3 & 10.33 & 0.001 & 77.5 & 12.93 & 0.004 & 74.2 & 96.2 & 69.4 \\
\midrule[\heavyrulewidth]
\multirow[c]{3}{*}{\shortstack[l]{\textbf{\methodname{}-KV}\\\textbf{RTX 5090}}} & 2/2 & 10.40 & 0.003 & 78.1 & 12.98 & 0.007 & 74.2 & 95.6 & 69.2 \\
 & 3/2 & 10.33 & 0.001 & 78.3 & 12.92 & 0.002 & 74.7 & 96.2 & 70.0 \\
 & 3/3 & 10.33 & 0.001 & 78.2 & 12.90 & 0.001 & 74.5 & 97.0 & 71.5 \\
\bottomrule
\end{tabular*}
\end{table}

\begin{figure}[!tp]
\centering
\includegraphics[width=0.955\columnwidth]{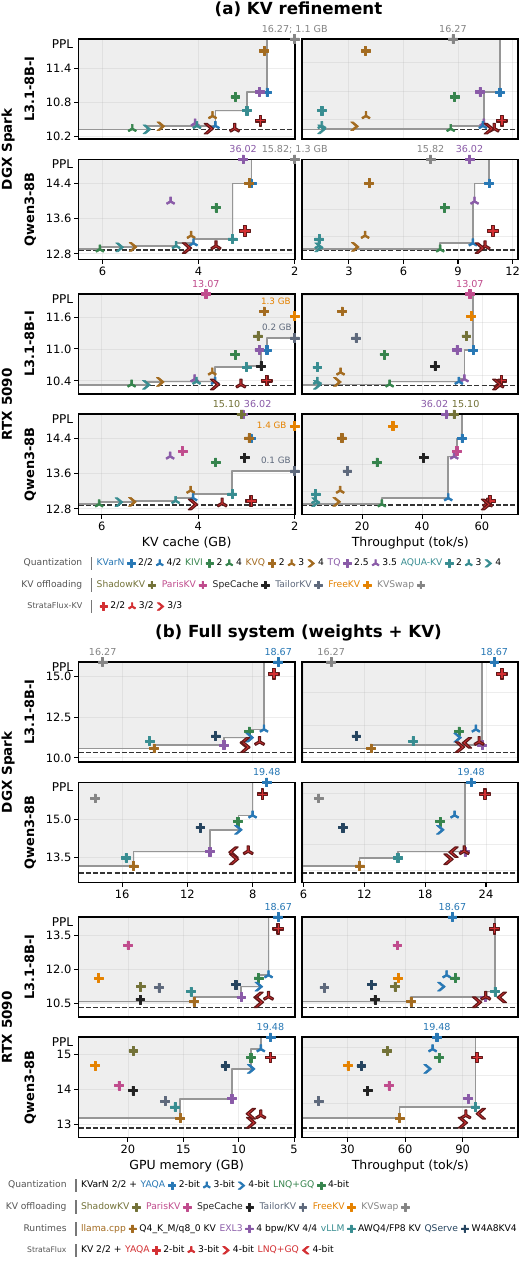}
\caption{PG-19 perplexity at 128K: (a) KV refinement with BF16 weights; (b) Weight and KV joint refinement. Memory counts KV in (a), weights+KV in (b), including DGX Spark's RAM cache. Out-of-range points are shown beside the corresponding borders with their values. Of the evaluated KV offloading systems, only KVSwap supports DGX Spark. The gray boundary marks the baseline frontier. Dashed lines mark BF16 PPL. W/A/KV: weight/activation/cache bits; bpw: bits/weight; KVQ: KVQuant; TQ: TurboQuant.} \input{figures/evaluation_frontiers_pending}
\label{fig:frontiers}
\end{figure}

\methodname{}-KV pushes beyond the memory--accuracy and throughput--accuracy frontiers of existing KV quantization and offloading methods, bringing near-BF16 accuracy to 2-bit keys and values without giving up low-bit memory savings or fast decoding (Table~\ref{tab:main-kv-comparison}; Figure~\ref{fig:frontiers}(a)). Accuracy recovery also extends to alternative attention architectures and MoE models (Appendix~\ref{app-sec:appendix-kv}, Table~\ref{app-tab:model_coverage-1}). On RTX 5090, Llama-3.1-8B-Instruct with 2/2 KV improves from 68.6\% to 78.1\% RULER accuracy at 128K.  For full task and context comparisons, see Appendix~\ref{app-sec:appendix-kv}.

\begin{figure}[!t]
\centering
\includegraphics[width=0.805\columnwidth]{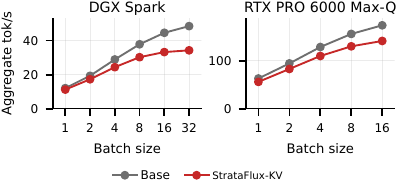}
\caption{\methodname{}'s KV refinement overhead versus batch size on Llama-3.1-8B-Instruct at 128K, using 2/2 KV.}
\label{fig:batched-throughput}
\end{figure}

At batch sizes 1 and 16, refinement reduces throughput by 11.2\% and 18.8\% on RTX PRO 6000 Max-Q, versus 7.0\% and 25.4\% on DGX Spark (Figure~\ref{fig:batched-throughput}).

\subsection{Full-System Results}
\label{sec:eval-joint}
We enable both refinement techniques and compare the full \methodname{} system with static weight/KV pairs, vLLM~\citep{kwon2023vllm}, EXL3~\citep{exllamav3}, QServe~\citep{lin2025qserve}, llama.cpp, and KV offloading systems.
\begin{table}[!htb]
\centering
\caption{Joint refinement: PG-19 perplexity/KL and RULER accuracy at 128K, plus MATH500/AIME accuracy. Platform blocks report \methodname{}.}
\label{tab:joint-downstream}
\label{tab:joint-generation}
\fontsize{6.8}{8.2}\selectfont
\setlength{\tabcolsep}{0.5pt}
\setlength{\aboverulesep}{0.3ex}
\setlength{\belowrulesep}{0.4ex}
\renewcommand{\arraystretch}{1.0}\setlength{\aboverulesep}{.2ex}\setlength{\belowrulesep}{.25ex}
\begin{tabular*}{\columnwidth}{@{\extracolsep{\fill}}@{}lllcccccccc@{}}
\toprule
\multicolumn{3}{l}{} & \multicolumn{3}{c}{\textbf{L3.1-8B-I}} & \multicolumn{5}{c}{\textbf{Qwen3-8B}} \\
\cmidrule(lr){4-6}\cmidrule(lr){7-11}
 & \textbf{Weights} & \textbf{KV} & \textbf{PPL} & \textbf{KL} & \textbf{R128} & \textbf{PPL} & \textbf{KL} & \textbf{R128} & \textbf{M500} & \textbf{AIME} \\
\midrule[\heavyrulewidth]
 & BF16 &  & 10.32 & 0.000 & 77.9 & 12.89 & 0.000 & 74.1 & 97.2 & 71.0 \\
\midrule[\heavyrulewidth]
\multirow[c]{4}{*}{\textbf{Base}} & YAQA 2-bit & 2/2 & 18.67 & 0.656 & 50.4 & 19.48 & 0.434 & 58.8 & 68.0 & 11.0 \\
 & YAQA 3-bit & 2/2 & 11.75 & 0.146 & 67.2 & 15.15 & 0.185 & 63.8 & 94.6 & 59.6 \\
 & YAQA 4-bit & 2/2 & 11.24 & 0.090 & 67.5 & 14.58 & 0.143 & 65.4 & 95.0 & 62.5 \\
 & LNQ+GQ 4-bit & 2/2 & 11.62 & 0.129 & 64.3 & 14.91 & 0.167 & 64.0 & 95.2 & 58.3 \\
\midrule[\heavyrulewidth]
\multirow[c]{4}{*}{\shortstack[l]{\textbf{\methodname{}}\\\textbf{DGX Spark}}} & YAQA 2-bit & 2/2 & 15.15 & 0.422 & 60.5 & 15.98 & 0.228 & 65.2 & 89.6 & 40.6 \\
 & YAQA 3-bit & 2/2 & 10.99 & 0.064 & 73.2 & 13.75 & 0.068 & 68.4 & 95.2 & 68.8 \\
 & YAQA 4-bit & 2/2 & 10.66 & 0.026 & 73.5 & 13.43 & 0.040 & 70.3 & 96.2 & 69.0 \\
 & LNQ+GQ 4-bit & 2/2 & 10.92 & 0.055 & 71.1 & 13.71 & 0.062 & 70.6 & 96.0 & 65.6 \\
\midrule[\heavyrulewidth]
\multirow[c]{4}{*}{\shortstack[l]{\textbf{\methodname{}}\\\textbf{RTX 5090}}} & YAQA 2-bit & 2/2 & 13.79 & 0.312 & 63.8 & 14.91 & 0.165 & 68.3 & 91.2 & 47.1 \\
 & YAQA 3-bit & 2/2 & 10.80 & 0.045 & 77.1 & 13.25 & 0.038 & 71.2 & 95.4 & 65.6 \\
 & YAQA 4-bit & 2/2 & 10.55 & 0.015 & 77.2 & 13.04 & 0.015 & 73.5 & 96.4 & 69.2 \\
 & LNQ+GQ 4-bit & 2/2 & 10.77 & 0.038 & 75.2 & 13.30 & 0.035 & 73.8 & 94.4 & 61.5 \\
\bottomrule
\end{tabular*}
\end{table}

Together, weight and KV refinement let \methodname{} recover long-context and reasoning accuracy with aggressively compressed weights and KV (Table~\ref{tab:joint-downstream}), pushing beyond the memory--accuracy and throughput--accuracy frontiers of static quantization pairs, KV offloading systems, and inference runtimes (Figure~\ref{fig:frontiers}(b)). On RTX 5090, refinement raises Qwen3-8B AIME accuracy from 11.0\% to 47.1\% with YAQA 2-bit weights and 2/2 KV. Appendix~\ref{app-sec:appendix-joint} gives the full configuration comparisons and task results.

\subsection{Ablations}
\label{sec:eval-ablations}
\begin{figure}[!htb]
\centering
\includegraphics[width=0.94\columnwidth]{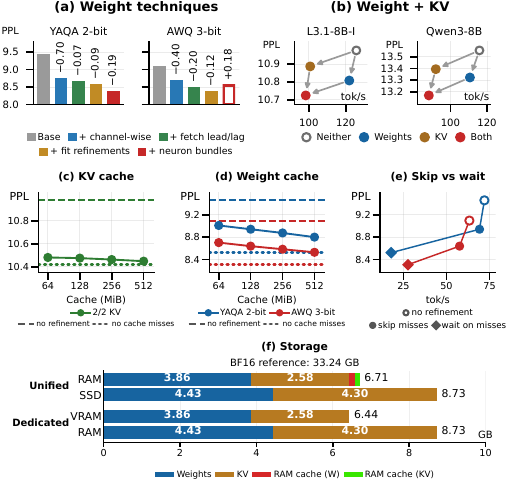}
\caption{Refinement ablations on Llama-3.1-8B-Instruct, using RTX 5090 in (a,b) and DGX Spark in (c--e). (a) Adding weight-refinement techniques, re-tuned to a 5\% slowdown budget. Labels show WikiText-2 PPL changes. The outlined AWQ bar forces neuron bundles. (b) PPL versus throughput for separate and combined weight/KV refinement (also on Qwen3-8B), using PG-19 at 128K, YAQA 3-bit weights, 3/3 KV, and the 5\% weight configuration. (c,d) PPL versus RAM-cache capacity with skipped misses: (c) 2/2 KV on PG-19; (d) weight refinement on WikiText-2 at 5\% configurations. (e) WikiText-2 PPL versus throughput for skipping and waiting on weight-cache misses with a 128 MiB cache. (f) Storage (GB) by tier: YAQA 2-bit weights and 2/2 KV at 128K.}
\label{fig:refinement-ablation}
\label{fig:component-ablation}
\label{fig:cache-ablation}
\end{figure}

Fetch lead/lag and fitting improve accuracy, while the better MLP scheme, channel-wise refinement or neuron bundles, depends on the quantizer (Figure~\ref{fig:refinement-ablation}(a)). Weight and KV refinement yield approximately additive accuracy gains and decoding-time overheads, allowing both to run together without negative interference (Figure~\ref{fig:component-ablation}(b)). RAM caches make SSD-backed refinement effective despite skipped misses. Larger caches improve accuracy, while skipping avoids the throughput penalty of waiting (Figure~\ref{fig:cache-ablation}(c--e)). Appendix~\ref{app-sec:appendix-kv} extends the fetch-budget and cache-capacity sweeps. With 4-bit weight and KV refinement values, backing-tier storage scales roughly with 4-bit weights and KV, plus metadata and values duplicated in neuron bundles (Figure~\ref{fig:refinement-ablation}(f)). For other KV bases, see Appendix~\ref{app-sec:appendix-kv-bases}.

\section{Related Work}
\label{sec:related-work}

Heterogeneous-memory inference systems accommodate weights and KV caches beyond GPU memory. FlexGen~\citep{sheng2023flexgen} schedules compressed tensors across GPU, CPU, and disk; InfiniGen~\citep{lee2024infinigen} and FreeKV~\citep{liu2026freekv} prefetch selected KV entries from host memory. Quest~\citep{tang2024quest} and SparQ~\citep{ribar2024sparq} reduce KV reads through query-dependent selection. SpeCache~\citep{jie2025specache} supplements a full low-bit cache with prefetched 16-bit entries. \methodname{} instead co-designs low-bit attention and selectively fetched refinement bits, using caching and skipped misses to recover accuracy without stalling decoding on backing-tier reads.

For weights, DecDEC~\citep{park2025decdec} selects refinement channels from current activations and overlaps host-memory fetching and refinement with the corresponding quantized projection. Other inference systems overlap work across Transformer blocks: Deja Vu~\citep{liu2023dejavu} uses earlier activations to predict which attention heads and MLP neurons to compute, while KTransformers~\citep{chen2025ktransformers} defers expert outputs to overlap CPU and GPU computation. \methodname{} extends weight-refinement overlap beyond DecDEC's projection window through early channel selection and deferred addition, and fits refinement values to compensate for unselected channels.
\section{Conclusion}
\label{sec:conclusion}

\methodname{} recovers accuracy from low-bit weights and KV caches while retaining quantization's memory and speed advantages. Its selective precision streaming makes host RAM and SSD useful sources of additional precision by coordinating what is fetched, when it arrives, and how it refines computation. Across dedicated GPUs and unified-memory platforms, \methodname{} demonstrates that precision distributed across the memory hierarchy can improve accuracy without letting backing-tier transfers dictate decoding speed.

\nocite{bai2024longbench,gao2023lmeval}
\bibliography{references}
\bibliographystyle{mlsys2025}
\section*{AI Usage Disclosure}
We used Anthropic's Claude Code and OpenAI's Codex to assist with implementation and debugging, evaluation and visualization scripts, figure and table preparation, and manuscript revision. These tools were used iteratively under author direction. The authors determined the research objectives, directed revisions, and selected the material included in the paper, and remain responsible for the implementation, experimental results, and final manuscript.

\clearpage
\setlength{\textfloatsep}{\templatetextfloatsep}
\setlength{\floatsep}{\templatefloatsep}
\setlength{\dbltextfloatsep}{\templatedbltextfloatsep}
\onecolumn
\raggedbottom
\raggedcolumns
\let\addcontentsline\appendixAddContentsLine
\renewcommand{\contentsname}{Appendix Contents}
\setcounter{tocdepth}{2}
\tableofcontents
\clearpage
\twocolumn
\appendix
\setcounter{figure}{0}
\setcounter{table}{0}
\renewcommand{\thefigure}{A\arabic{figure}}
\renewcommand{\thetable}{A\arabic{table}}
\renewcommand{\theHfigure}{appendix.\arabic{figure}}
\renewcommand{\theHtable}{appendix.\arabic{table}}
\renewcommand{\versionlabelprefix}{app-}
\section{Evaluation Setup and Implementation Details}
\label{sec:appendix-provenance}
\begingroup
\setlength{\parindent}{0pt}
\noindent\textbf{Perplexity and KL.} Weight evaluation uses 2K-token WikiText-2 windows. For long-context evaluation, we sample 17 PG-19 test books with sufficient length and evaluate 16 windows of 128K tokens from each book. We prefill each window except its final 256 tokens, then score those tokens sequentially with teacher forcing. Context-length sweeps shorten the windows to 8K--128K tokens while keeping their endpoints and scored tokens fixed. KL uses the same datasets, windows, and scored tokens as perplexity. We approximate divergence from BF16 by keeping its 1,024 most probable vocabulary tokens separate and grouping all remaining tokens into one category in both distributions.

\noindent\textbf{Evaluation software.} Weight evaluation uses the Language Model Evaluation Harness (lm-eval)~\citep{gao2023lmeval}. We evaluate LongBench~\citep{bai2024longbench} using the benchmark's evaluation scripts.

\noindent\textbf{Timing and memory.} We implement \methodname{} in llama.cpp, with CUDA 13.4.2 on NVIDIA GPUs and Metal on the Mac. For fair comparisons, we port baselines with author-provided kernels into the same framework. Systems with their own runtimes, including KV offloading systems, vLLM, and ExLlamaV3, are measured in their respective implementations. KV and joint decode timing uses a PG-19 prompt, prefill to $C-641$ tokens for context length $C$, 512 untimed steps, and 128 timed steps. For weight-only comparisons on NVIDIA GPUs, we measure decode throughput at empty context for WikiText-2 and at 128K context for PG-19. Memory counts stored representations, metadata, and RAM caches where used; captions specify whether weights, KV, or both are counted. Memory values at 128K context assume 131,072 tokens.

\noindent\textbf{KV refinement settings.} Main evaluations use resident key/value bitwidths of 2/2, 3/2, and 3/3, with +4/+4 refinement bits and a 1/128 token-group fetch fraction. Groups contain two tokens on dedicated GPUs and four on unified-memory platforms. Appendix~\ref{sec:kv-settings-sweep} varies resident and fetched bitwidths, token-group size, and fetch fraction. The first 128 tokens remain in FP16 unless ablated. A 128-token buffer for the current token block also remains in FP16; resident-memory totals include both allocations.

\appendixgroup{Baseline Kernels Integrated into llama.cpp}
\noindent\textbf{YAQA.} We integrate QTIP's released decompression and matrix--vector kernels for YAQA's weight format.

\noindent\textbf{SqLLM and LNQ+GQ.} SqLLM uses Any-Precision LLM's quantizer and kernels; LNQ+GQ uses GuidedQuant's weights with adapted Any-Precision kernels.

\noindent\textbf{AWQ 4-bit.} We integrate the authors' released matrix--vector and matrix--matrix kernels.

\noindent\textbf{KVQuant.} We add grouped-query attention support to KVQuant's released quantization and attention kernels.

\noindent\textbf{KIVI.} We retain KIVI's released matrix--vector kernels and supply packing and softmax integration in llama.cpp.

\noindent\textbf{KVarN.} KVarN baselines use the released Triton kernels.

\noindent\textbf{DecDEC.} We port DecDEC's correction kernel into llama.cpp and select its configurations with DecDEC's tuner.

\appendixgroup{Our Baseline Implementations and Adaptations}
\noindent\textbf{AWQ 3-bit.} The authors release a 4-bit matrix--vector kernel but no 3-bit kernel, so we implement a group-size-128 3-bit kernel in llama.cpp.

\noindent\textbf{TurboQuant.} With no author-released implementation available at integration time, we adapt the community vLLM backend (v0.30.0) to llama.cpp and match the paper's quantization settings.

\noindent\textbf{AQUA-KV.} The authors release PyTorch code without CUDA kernels; we adapt it to CUDA, reconstructing cached keys and values during decoding.

\noindent\textbf{SpeCache.} With no author-released code available, we implement the paper's low-bit KV cache and selective FP16 fetching in llama.cpp.

\appendixgroup{Systems Evaluated in Their Own Runtimes}
\noindent We measure ShadowKV, FreeKV, ParisKV, TailorKV, and KVSwap in their released runtimes. The vLLM, ExLlamaV3, and QServe comparisons also use their respective runtimes; we add Qwen3-8B support to QServe.

\appendixgroup{\methodname{} Implementation}
\noindent\textbf{Weights.} We adapt DecDEC's dynamic error-compensation kernel, replacing bucket-based selection with exact radix-sort selection and improving zero-copy request batching at low GPU thread-block counts. We add fetch lead/lag scheduling, RAM caches, and refinement delivery logic, and develop an independent tuner inspired by DecDEC's search procedure.

\noindent\textbf{KV.} We adopt KVarN's quantization algorithm and independently develop CUDA and Metal attention kernels with fused refinement operations.

\par\endgroup

\section{Extended Weight Evaluation}
\label{sec:appendix-weights}
We extend Sections~\ref{main-sec:eval-weights} and~\ref{main-sec:eval-ablations} across weight bases, model sizes, slowdown budgets, and cache configurations. We also test weight refinement with YAQA's recovery fine-tuning (Section~\ref{sec:yaqa-recovery-finetuning}).
\subsection{Accuracy and Downstream Tasks}
Tables~\ref{tab:weight_comparison_complete-1} and~\ref{tab:weight-comparison-dedicated} extend the 2--4-bit PPL/KL comparisons across platforms; Table~\ref{tab:weight-downstream-complete} adds downstream DecDEC comparisons.
\begin{appendixcolumntable}
\centering
\caption{Weight-only refinement: WikiText-2 perplexity/KL with DGX Spark configurations at 5\% slowdown budgets.}
\label{tab:weight_comparison_complete-1}
\fontsize{6.5}{7.6}\selectfont\setlength{\tabcolsep}{0.6pt}\renewcommand{\arraystretch}{1.0}\setlength{\aboverulesep}{.2ex}\setlength{\belowrulesep}{.25ex}
\begin{tabular*}{\linewidth}{@{\extracolsep{\fill}}@{}llcccc@{}}
\toprule
 & & \multicolumn{2}{c}{\textbf{L3.1-8B-I}} & \multicolumn{2}{c}{\textbf{Qwen3-8B}}\\
\cmidrule(lr){3-4}\cmidrule(lr){5-6}
\textbf{Weights} & \textbf{Method} & \textbf{PPL} & \textbf{KL} & \textbf{PPL} & \textbf{KL}\\\midrule
BF16 &  & 7.22 & 0.000 & 9.73 & 0.000\\
\midrule
\multirow[c]{2}{*}{\begin{tabular}[c]{@{}l@{}}YAQA\\2-bit\end{tabular}} & Base & 9.46 & 0.257 & 10.95 & 0.214\\
 & \methodname{}-W & 8.94 & 0.200 & 10.77 & 0.175\\
\midrule
\multirow[c]{2}{*}{\begin{tabular}[c]{@{}l@{}}YAQA\\3-bit\end{tabular}} & Base & 7.61 & 0.047 & 9.98 & 0.052\\
 & \methodname{}-W & 7.55 & 0.041 & 9.89 & 0.043\\
\midrule
\multirow[c]{2}{*}{\begin{tabular}[c]{@{}l@{}}YAQA\\4-bit\end{tabular}} & Base & 7.33 & 0.014 & 9.79 & 0.016\\
 & \methodname{}-W & 7.31 & 0.012 & 9.72 & 0.014\\
\midrule
\multirow[c]{2}{*}{\begin{tabular}[c]{@{}l@{}}LNQ+GQ\\3-bit\end{tabular}} & Base & 8.52 & 0.157 & 10.92 & 0.184\\
 & \methodname{}-W & 8.29 & 0.135 & 10.70 & 0.159\\
\midrule
\multirow[c]{2}{*}{\begin{tabular}[c]{@{}l@{}}LNQ+GQ\\4-bit\end{tabular}} & Base & 7.48 & 0.034 & 9.95 & 0.044\\
 & \methodname{}-W & 7.43 & 0.031 & 9.87 & 0.038\\
\midrule
\multirow[c]{2}{*}{\begin{tabular}[c]{@{}l@{}}AWQ\\3-bit\end{tabular}} & Base & 9.10 & 0.229 & 11.88 & 0.284\\
 & \methodname{}-W & 8.64 & 0.179 & 11.30 & 0.222\\
\midrule
\multirow[c]{2}{*}{\begin{tabular}[c]{@{}l@{}}AWQ\\4-bit\end{tabular}} & Base & 7.59 & 0.049 & 10.01 & 0.059\\
 & \methodname{}-W & 7.51 & 0.039 & 9.96 & 0.047\\
\midrule
\multirow[c]{2}{*}{\begin{tabular}[c]{@{}l@{}}SqLLM\\3-bit\end{tabular}} & Base & 9.36 & 0.263 & 11.71 & 0.255\\
 & \methodname{}-W & 8.56 & 0.174 & 11.03 & 0.201\\
\midrule
\multirow[c]{2}{*}{\begin{tabular}[c]{@{}l@{}}SqLLM\\4-bit\end{tabular}} & Base & 7.56 & 0.052 & 9.98 & 0.062\\
 & \methodname{}-W & 7.48 & 0.040 & 9.97 & 0.048\\
\bottomrule\end{tabular*}
\end{appendixcolumntable}

\begin{appendixcolumntable}
\centering
\caption{Weight-only refinement: WikiText-2 perplexity/KL with configurations tuned on RTX 5090 at 5\% slowdown budgets.}
\label{tab:weight-comparison-dedicated}
\fontsize{7.5}{9}\selectfont\setlength{\tabcolsep}{0.6pt}\renewcommand{\arraystretch}{1.0}\setlength{\aboverulesep}{.2ex}\setlength{\belowrulesep}{.25ex}
\begin{tabular*}{\linewidth}{@{\extracolsep{\fill}}@{}llcccc@{}}
\toprule
 & & \multicolumn{2}{c}{\textbf{L3.1-8B-I}} & \multicolumn{2}{c}{\textbf{Qwen3-8B}}\\
\cmidrule(lr){3-4}\cmidrule(lr){5-6}
\textbf{Weights} & \textbf{Method} & \textbf{PPL} & \textbf{KL} & \textbf{PPL} & \textbf{KL}\\\midrule
BF16 &  & 7.22 & 0.000 & 9.73 & 0.000\\
\midrule
\multirow[c]{3}{*}{\begin{tabular}[c]{@{}l@{}}YAQA\\2-bit\end{tabular}} & Base & 9.46 & 0.257 & 10.95 & 0.214\\
 & DecDEC & 9.03 & 0.211 & 10.69 & 0.184\\
 & \methodname{}-W & 8.40 & 0.139 & 10.51 & 0.132\\
\midrule
\multirow[c]{3}{*}{\begin{tabular}[c]{@{}l@{}}YAQA\\3-bit\end{tabular}} & Base & 7.61 & 0.047 & 9.98 & 0.052\\
 & DecDEC & 7.55 & 0.040 & 9.89 & 0.044\\
 & \methodname{}-W & 7.49 & 0.033 & 9.89 & 0.033\\
\midrule
\multirow[c]{3}{*}{\begin{tabular}[c]{@{}l@{}}YAQA\\4-bit\end{tabular}} & Base & 7.33 & 0.014 & 9.79 & 0.016\\
 & DecDEC & 7.30 & 0.011 & 9.75 & 0.013\\
 & \methodname{}-W & 7.29 & 0.010 & 9.75 & 0.010\\
\midrule
\multirow[c]{3}{*}{\begin{tabular}[c]{@{}l@{}}LNQ+GQ\\3-bit\end{tabular}} & Base & 8.52 & 0.157 & 10.92 & 0.184\\
 & DecDEC & 8.37 & 0.142 & 10.66 & 0.170\\
 & \methodname{}-W & 8.18 & 0.121 & 10.47 & 0.139\\
\midrule
\multirow[c]{3}{*}{\begin{tabular}[c]{@{}l@{}}LNQ+GQ\\4-bit\end{tabular}} & Base & 7.48 & 0.034 & 9.95 & 0.044\\
 & DecDEC & 7.42 & 0.031 & 9.89 & 0.038\\
 & \methodname{}-W & 7.41 & 0.027 & 9.88 & 0.034\\
\midrule
\multirow[c]{3}{*}{\begin{tabular}[c]{@{}l@{}}AWQ\\3-bit\end{tabular}} & Base & 9.10 & 0.229 & 11.88 & 0.284\\
 & DecDEC & 8.80 & 0.195 & 11.38 & 0.231\\
 & \methodname{}-W & 8.38 & 0.145 & 10.71 & 0.167\\
\midrule
\multirow[c]{3}{*}{\begin{tabular}[c]{@{}l@{}}AWQ\\4-bit\end{tabular}} & Base & 7.59 & 0.049 & 10.01 & 0.059\\
 & DecDEC & 7.52 & 0.042 & 9.92 & 0.049\\
 & \methodname{}-W & 7.44 & 0.031 & 9.88 & 0.038\\
\midrule
\multirow[c]{3}{*}{\begin{tabular}[c]{@{}l@{}}SqLLM\\3-bit\end{tabular}} & Base & 9.36 & 0.263 & 11.71 & 0.255\\
 & DecDEC & 9.02 & 0.225 & 11.36 & 0.215\\
 & \methodname{}-W & 8.42 & 0.156 & 10.81 & 0.169\\
\midrule
\multirow[c]{3}{*}{\begin{tabular}[c]{@{}l@{}}SqLLM\\4-bit\end{tabular}} & Base & 7.56 & 0.052 & 9.98 & 0.062\\
 & DecDEC & 7.51 & 0.044 & 9.86 & 0.049\\
 & \methodname{}-W & 7.44 & 0.035 & 9.85 & 0.038\\
\bottomrule\end{tabular*}
\end{appendixcolumntable}

\begin{appendixcolumntable}\centering
\begingroup\setlength{\tabcolsep}{4pt}\renewcommand{\arraystretch}{1.0}\setlength{\aboverulesep}{.2ex}\setlength{\belowrulesep}{.25ex}\fontsize{8}{9.5}\selectfont
\caption{Weight-only refinement on non-Instruct Llama-3.1-8B: WikiText-2 PPL, 2K windows, 5\% slowdown budget on RTX 5090.}\label{tab:non-instruct-weights}
\begin{tabular*}{\linewidth}{@{\extracolsep{\fill}}@{}p{0.35\linewidth}ccc@{}}
\toprule
\textbf{Weights} & \textbf{Base} & \textbf{DecDEC} & \textbf{\methodname{}-W}\\\midrule

BF16 & 6.24 & \multicolumn{1}{c}{--} & \multicolumn{1}{c}{--} \\
LNQ+GQ 3-bit & 7.49 & 7.37 & 7.21\\
LNQ+GQ 4-bit & 6.53 & 6.50 & 6.47\\
YAQA 2-bit & 8.12 & 7.82 & 7.38\\
YAQA 3-bit & 6.62 & 6.56 & 6.51\\
YAQA 4-bit & 6.35 & 6.32 & 6.31\\
\bottomrule\end{tabular*}\endgroup
\end{appendixcolumntable}

Table~\ref{tab:non-instruct-weights} reports the non-Instruct Llama comparison.
\begin{appendixcolumntable}
\centering
\newlength{\weightdownstreamwidth}
\caption{Weight-only refinement on Qwen3-8B: MATH500 and AIME 2024/2025 (avg@8) accuracy at 5\% slowdown budgets; \methodname{}-W columns use the configuration tuned on each platform.}
\label{tab:weight-downstream-complete}
\fontsize{7.2}{8.6}\selectfont\setlength{\tabcolsep}{1pt}\renewcommand{\arraystretch}{1.0}\setlength{\aboverulesep}{.2ex}\setlength{\belowrulesep}{.25ex}
\begin{tabular*}{\linewidth}{@{\extracolsep{\fill}}lcccccccc@{}}\toprule
 & \multicolumn{2}{c}{\textbf{Base}} & \multicolumn{2}{c}{\textbf{DecDEC}} & \multicolumn{2}{c}{\begin{tabular}[b]{@{}c@{}}\textbf{\methodname{}-W}\\\textbf{DGX Spark}\end{tabular}} & \multicolumn{2}{c}{\begin{tabular}[b]{@{}c@{}}\textbf{\methodname{}-W}\\\textbf{RTX 5090}\end{tabular}}\\
\cmidrule(lr){2-3}\cmidrule(lr){4-5}\cmidrule(lr){6-7}\cmidrule(lr){8-9}
\textbf{Weights} & \textbf{M500} & \textbf{AIME} & \textbf{M500} & \textbf{AIME} & \textbf{M500} & \textbf{AIME} & \textbf{M500} & \textbf{AIME}\\\midrule
BF16 & 97.2 & 71.0 & \multicolumn{1}{c}{--} & \multicolumn{1}{c}{--} & \multicolumn{1}{c}{--} & \multicolumn{1}{c}{--} & \multicolumn{1}{c}{--} & \multicolumn{1}{c}{--} \\
YAQA 2-bit & 77.2 & 19.2 & 86.2 & 30.8 & 89.4 & 40.0 & 92.4 & 46.7\\
YAQA 3-bit & 95.0 & 66.7 & 95.4 & 66.9 & 95.0 & 66.0 & 95.2 & 70.0\\
YAQA 4-bit & 95.8 & 71.5 & 95.4 & 67.7 & 95.6 & 70.8 & 97.0 & 69.8\\
LNQ+GQ 3-bit & 91.8 & 39.4 & 93.0 & 44.4 & 93.0 & 47.3 & 93.2 & 49.2\\
LNQ+GQ 4-bit & 96.2 & 64.4 & 96.4 & 66.7 & 95.8 & 66.0 & 94.4 & 67.5\\
AWQ 3-bit & 91.8 & 46.5 & 92.8 & 50.8 & 92.6 & 51.9 & 93.8 & 51.0\\
AWQ 4-bit & 95.8 & 64.8 & 95.2 & 66.9 & 96.4 & 68.5 & 95.8 & 65.0\\
SqLLM 3-bit & 93.0 & 47.3 & 92.8 & 49.8 & 94.2 & 48.3 & 93.0 & 50.8\\
SqLLM 4-bit & 95.2 & 64.2 & 96.0 & 67.1 & 95.6 & 66.5 & 95.2 & 70.2\\
\bottomrule\end{tabular*}
\end{appendixcolumntable}

\begin{minipage}{\linewidth}
\setlength{\parskip}{6pt}
\subsection{Long-Context Weight-Only Refinement}
Table~\ref{tab:long-context-weights} evaluates weight refinement at 128K context with FP16 KV.
\begin{appendixcolumntable}
\centering
\caption{Weight-only refinement: PG-19 PPL/KL and RTX 5090 tok/s at 128K with FP16 KV, on Llama-3.1-8B-Instruct and Qwen3-8B. \methodname{}-W reuses its empty-context 5\% configurations.}
\label{tab:long-context-weights}
\fontsize{7}{8.4}\selectfont\setlength{\tabcolsep}{.6pt}\renewcommand{\arraystretch}{1.0}\setlength{\aboverulesep}{.2ex}\setlength{\belowrulesep}{.25ex}
\begin{tabular*}{\linewidth}{@{\extracolsep{\fill}}@{}llcccccc@{}}\toprule
 & & \multicolumn{3}{c}{\textbf{L3.1-8B-I}} & \multicolumn{3}{c}{\textbf{Qwen3-8B}}\\
\cmidrule(lr){3-5}\cmidrule(lr){6-8}
\textbf{Weights} & \textbf{Method} & \textbf{PPL} & \textbf{KL} & \textbf{tok/s} & \textbf{PPL} & \textbf{KL} & \textbf{tok/s}\\\midrule
BF16 &  & 10.32 & 0.000 & 47.6 & 12.89 & 0.000 & OOM\\
\midrule

\multirow[c]{2}{*}{YAQA 2-bit} & Base & 15.89 & 0.473 & 64.2 & 16.55 & 0.254 & 58.2\\
 & \methodname{}-W & 13.73 & 0.311 & 57.3 & 14.83 & 0.162 & 50.2\\
\midrule
\multirow[c]{2}{*}{YAQA 3-bit} & Base & 10.87 & 0.057 & 62.5 & 13.38 & 0.045 & 57.0\\
 & \methodname{}-W & 10.71 & 0.042 & 56.0 & 13.16 & 0.034 & 47.2\\
\midrule
\multirow[c]{2}{*}{YAQA 4-bit} & Base & 10.51 & 0.016 & 61.4 & 13.02 & 0.014 & 55.4\\
 & \methodname{}-W & 10.47 & 0.013 & 60.0 & 12.95 & 0.010 & 48.6\\
\midrule
\multirow[c]{2}{*}{LNQ+GQ 3-bit} & Base & 13.65 & 0.303 & 68.1 & 18.15 & 0.328 & 61.2\\
 & \methodname{}-W & 12.96 & 0.243 & 67.3 & 17.21 & 0.283 & 60.6\\
\midrule
\multirow[c]{2}{*}{LNQ+GQ 4-bit} & Base & 10.77 & 0.043 & 65.7 & 13.28 & 0.037 & 59.1\\
 & \methodname{}-W & 10.69 & 0.037 & 65.2 & 13.22 & 0.032 & 58.1\\
\bottomrule\end{tabular*}

\end{appendixcolumntable}

\end{minipage}
\appendixfirstgroupreserve{9}
\subsection{Larger Models}
Table~\ref{tab:large-weight-refinement} compares weight refinement on Llama-3.3-70B-Instruct and Qwen3-32B.
\begin{appendixcolumntable}
\centering
\caption{Weight-only refinement: WikiText-2 perplexity and KL divergence on Llama-3.3-70B-Instruct and Qwen3-32B at matched 5\% slowdown budgets. Throughput uses empty context: Qwen3-32B on RTX 5090; $^{*}$ Llama-3.3-70B-Instruct on RTX PRO 6000.}
\label{tab:large-weight-refinement}
\label{tab:large_model_accuracy-1}
\fontsize{6.7}{8.1}\selectfont
\setlength{\tabcolsep}{.5pt}
\renewcommand{\arraystretch}{1.0}\setlength{\aboverulesep}{.2ex}\setlength{\belowrulesep}{.25ex}
\begin{tabular*}{\linewidth}{@{\extracolsep{\fill}}@{}llcccccc@{}}
\toprule
 & & \multicolumn{3}{c}{\textbf{L3.3-70B-I}} & \multicolumn{3}{c}{\textbf{Qwen3-32B}} \\
\cmidrule(lr){3-5}\cmidrule(lr){6-8}
\textbf{Weights} & \textbf{Method} & \textbf{PPL} & \textbf{KL} & \textbf{tok/s\rlap{$^{*}$}} & \textbf{PPL} & \textbf{KL} & \textbf{tok/s} \\
\midrule
BF16 & Base & 3.86 & 0.000 & OOM & 7.61 & 0.000 & OOM \\
\midrule
\multirow[c]{3}{*}{YAQA 2-bit} & Base & 6.19 & 0.449 & 43.5 & 8.30 & 0.211 & 76.2 \\
 & DecDEC & 5.71 & 0.373 & 42.8 & 8.04 & 0.164 & 71.8 \\
 & \methodname{}-W & 5.52 & 0.349 & 41.9 & 8.04 & 0.153 & 72.8 \\
\midrule
\multirow[c]{3}{*}{YAQA 3-bit} & Base & 4.50 & 0.141 & 36.9 & 7.70 & 0.064 & 67.7 \\
 & DecDEC & 4.36 & 0.111 & 35.6 & 7.68 & 0.049 & 62.9 \\
 & \methodname{}-W & 4.33 & 0.107 & 35.3 & 7.69 & 0.043 & 62.5 \\
\midrule
\multirow[c]{3}{*}{YAQA 4-bit} & Base & 4.06 & 0.046 & 32.7 & 7.62 & 0.023 & 62.9 \\
 & DecDEC & 4.00 & 0.035 & 31.3 & 7.61 & 0.016 & 58.7 \\
 & \methodname{}-W & 3.99 & 0.033 & 31.4 & 7.62 & 0.015 & 59.6 \\
\midrule
\multirow[c]{3}{*}{LNQ+GQ 3-bit} & Base & 5.39 & 0.320 & 46.6 & 8.15 & 0.173 & 98.0 \\
 & DecDEC & 5.15 & 0.282 & 43.5 & 8.10 & 0.153 & 92.9 \\
 & \methodname{}-W & 5.05 & 0.261 & 44.9 & 7.94 & 0.132 & 92.6 \\
\midrule
\multirow[c]{3}{*}{LNQ+GQ 4-bit} & Base & 4.26 & 0.096 & 36.5 & 7.72 & 0.053 & 78.2 \\
 & DecDEC & 4.20 & 0.082 & 34.0 & 7.65 & 0.044 & 74.4 \\
 & \methodname{}-W & 4.17 & 0.078 & 35.1 & 7.65 & 0.039 & 73.4 \\
\midrule
\multirow[c]{3}{*}{AWQ 3-bit} & Base & 5.75 & 0.371 & 35.3 & 8.56 & 0.241 & 83.8 \\
 & DecDEC & 5.09 & 0.260 & 34.0 & 8.30 & 0.187 & 81.2 \\
 & \methodname{}-W & 5.04 & 0.244 & 34.1 & 8.01 & 0.146 & 80.8 \\
\midrule
\multirow[c]{3}{*}{AWQ 4-bit} & Base & 4.40 & 0.110 & 37.8 & 7.84 & 0.070 & 77.6 \\
 & DecDEC & 4.16 & 0.069 & 36.0 & 7.76 & 0.052 & 73.8 \\
 & \methodname{}-W & 4.16 & 0.066 & 36.1 & 7.71 & 0.043 & 73.3 \\
\midrule
\multirow[c]{3}{*}{SqLLM 3-bit} & Base & 5.64 & 0.377 & 46.3 & 8.68 & 0.242 & 98.1 \\
 & DecDEC & 5.10 & 0.272 & 44.2 & 8.52 & 0.211 & 94.6 \\
 & \methodname{}-W & 5.07 & 0.253 & 44.5 & 8.09 & 0.148 & 94.3 \\
\midrule
\multirow[c]{3}{*}{SqLLM 4-bit} & Base & 4.28 & 0.105 & 36.1 & 7.82 & 0.071 & 78.1 \\
 & DecDEC & 4.17 & 0.079 & 34.1 & 7.71 & 0.049 & 74.5 \\
 & \methodname{}-W & 4.17 & 0.074 & 34.9 & 7.69 & 0.043 & 73.8 \\
\bottomrule
\end{tabular*}
\end{appendixcolumntable}

\subsection{Slowdown-Budget Sweeps}
Tables~\ref{tab:spark-weight-budget-sweep}, \ref{tab:weight_budgets-1}, and~\ref{tab:weight_budgets-3} compare weight-refinement accuracy and throughput across slowdown budgets on DGX Spark, RTX 5090, and RTX 5070 Ti.
\appendixfirstgroupreserve{8}
\begin{appendixflowtable}
\caption{Weight-only refinement on DGX Spark: 2--4-bit bases across slowdown budgets. Columns report WikiText-2 PPL/KL and Llama-3.1-8B-Instruct empty-context tok/s.}\label{tab:spark-weight-budget-sweep}
\fontsize{6.8}{8.2}\selectfont\setlength{\tabcolsep}{2pt}\renewcommand{\arraystretch}{1.0}\setlength{\aboverulesep}{.2ex}\setlength{\belowrulesep}{.25ex}
\tablehead{\toprule
\multicolumn{7}{@{}l}{\textbf{DGX Spark}}\\
 & & \multicolumn{3}{c}{\textbf{L3.1-8B-I}} & \multicolumn{2}{c}{\textbf{Qwen3-8B}}\\\cmidrule(lr){3-5}\cmidrule(lr){6-7}
\textbf{Weights} & \textbf{Budget / method} & \textbf{PPL} & \textbf{KL} & \textbf{tok/s} & \textbf{PPL} & \textbf{KL}\\\midrule
}
\tabletail{\bottomrule}\tablelasttail{\bottomrule}
\begin{supertabular*}{\linewidth}{@{\extracolsep{\fill}}@{}llccccc@{}}
BF16 & Base & 7.22 & 0.000 & 15.6 & 9.73 & 0.000\\
\midrule

\appendixgroupreserve{5}
\multirow[c]{5}{*}{YAQA 2-bit} & Base & 9.46 & 0.257 & 72.0 & 10.95 & 0.214\\
 & 2.5\% / \methodname{}-W & 8.96 & 0.202 & 70.9 & 10.81 & 0.178\\
 & 5\% / \methodname{}-W & 8.94 & 0.200 & 69.2 & 10.77 & 0.175\\
 & 10\% / \methodname{}-W & 8.90 & 0.194 & 65.7 & 10.67 & 0.174\\
 & 20\% / \methodname{}-W & 8.92 & 0.198 & 60.0 & 10.69 & 0.174\\
\midrule
\appendixgroupreserve{5}
\multirow[c]{5}{*}{YAQA 3-bit} & Base & 7.61 & 0.047 & 57.7 & 9.98 & 0.052\\
 & 2.5\% / \methodname{}-W & 7.56 & 0.042 & 57.4 & 9.91 & 0.044\\
 & 5\% / \methodname{}-W & 7.55 & 0.041 & 55.8 & 9.89 & 0.043\\
 & 10\% / \methodname{}-W & 7.56 & 0.041 & 53.1 & 9.89 & 0.045\\
 & 20\% / \methodname{}-W & 7.56 & 0.041 & 49.0 & 9.88 & 0.044\\
\midrule
\appendixgroupreserve{5}
\multirow[c]{5}{*}{LNQ+GQ 3-bit} & Base & 8.52 & 0.157 & 60.9 & 10.92 & 0.184\\
 & 2.5\% / \methodname{}-W & 8.34 & 0.139 & 61.4 & 10.73 & 0.160\\
 & 5\% / \methodname{}-W & 8.29 & 0.135 & 59.8 & 10.70 & 0.159\\
 & 10\% / \methodname{}-W & 8.31 & 0.135 & 57.1 & 10.64 & 0.149\\
 & 20\% / \methodname{}-W & 8.31 & 0.135 & 52.8 & 10.66 & 0.154\\
\midrule
\appendixgroupreserve{5}
\multirow[c]{5}{*}{AWQ 3-bit} & Base & 9.10 & 0.229 & 63.3 & 11.88 & 0.284\\
 & 2.5\% / \methodname{}-W & 8.75 & 0.190 & 61.1 & 11.33 & 0.234\\
 & 5\% / \methodname{}-W & 8.64 & 0.179 & 59.2 & 11.30 & 0.222\\
 & 10\% / \methodname{}-W & 8.63 & 0.175 & 56.2 & 11.16 & 0.214\\
 & 20\% / \methodname{}-W & 8.64 & 0.177 & 52.9 & 11.12 & 0.211\\
\midrule
\appendixgroupreserve{5}
\multirow[c]{5}{*}{SqLLM 3-bit} & Base & 9.36 & 0.263 & 60.8 & 11.71 & 0.255\\
 & 2.5\% / \methodname{}-W & 8.72 & 0.194 & 60.5 & 11.19 & 0.210\\
 & 5\% / \methodname{}-W & 8.56 & 0.174 & 59.3 & 11.03 & 0.201\\
 & 10\% / \methodname{}-W & 8.51 & 0.170 & 57.4 & 11.14 & 0.187\\
 & 20\% / \methodname{}-W & 8.51 & 0.171 & 53.1 & 11.11 & 0.186\\
\midrule
\appendixgroupreserve{5}
\multirow[c]{5}{*}{YAQA 4-bit} & Base & 7.33 & 0.014 & 48.1 & 9.79 & 0.016\\
 & 2.5\% / \methodname{}-W & 7.31 & 0.012 & 47.5 & 9.73 & 0.014\\
 & 5\% / \methodname{}-W & 7.31 & 0.012 & 46.6 & 9.72 & 0.014\\
 & 10\% / \methodname{}-W & 7.31 & 0.012 & 44.4 & 9.71 & 0.014\\
 & 20\% / \methodname{}-W & 7.31 & 0.012 & 40.9 & 9.72 & 0.014\\
\midrule
\appendixgroupreserve{5}
\multirow[c]{5}{*}{LNQ+GQ 4-bit} & Base & 7.48 & 0.034 & 49.4 & 9.95 & 0.044\\
 & 2.5\% / \methodname{}-W & 7.45 & 0.031 & 49.9 & 9.88 & 0.038\\
 & 5\% / \methodname{}-W & 7.43 & 0.031 & 49.0 & 9.87 & 0.038\\
 & 10\% / \methodname{}-W & 7.44 & 0.031 & 46.7 & 9.88 & 0.039\\
 & 20\% / \methodname{}-W & 7.44 & 0.031 & 42.3 & 9.86 & 0.038\\
\midrule
\appendixgroupreserve{5}
\multirow[c]{5}{*}{AWQ 4-bit} & Base & 7.59 & 0.049 & 49.5 & 10.01 & 0.059\\
 & 2.5\% / \methodname{}-W & 7.52 & 0.040 & 48.8 & 9.96 & 0.047\\
 & 5\% / \methodname{}-W & 7.51 & 0.039 & 46.1 & 9.96 & 0.047\\
 & 10\% / \methodname{}-W & 7.51 & 0.038 & 45.4 & 9.90 & 0.046\\
 & 20\% / \methodname{}-W & 7.51 & 0.038 & 42.4 & 9.90 & 0.045\\
\midrule
\appendixgroupreserve{5}
\multirow[c]{5}{*}{SqLLM 4-bit} & Base & 7.56 & 0.052 & 49.5 & 9.98 & 0.062\\
 & 2.5\% / \methodname{}-W & 7.50 & 0.043 & 48.1 & 9.89 & 0.050\\
 & 5\% / \methodname{}-W & 7.48 & 0.040 & 48.6 & 9.97 & 0.048\\
 & 10\% / \methodname{}-W & 7.47 & 0.039 & 46.6 & 9.87 & 0.046\\
 & 20\% / \methodname{}-W & 7.47 & 0.039 & 44.0 & 9.88 & 0.046\cr
\end{supertabular*}

\end{appendixflowtable}

\appendixfirstgroupreserve{11}
\begin{appendixflowtable}
\caption{Weight-only refinement on RTX 5090: 2--4-bit bases across slowdown budgets. Columns report WikiText-2 PPL/KL and empty-context decode tok/s.}
\label{tab:weight_budgets-1}\label{tab:weight_budgets-2}
\fontsize{6.8}{8.2}\selectfont\setlength{\tabcolsep}{2pt}\renewcommand{\arraystretch}{1.0}\setlength{\aboverulesep}{.2ex}\setlength{\belowrulesep}{.25ex}
\tablehead{\toprule
\multicolumn{8}{@{}l}{\textbf{RTX 5090}}\\
 & & \multicolumn{3}{c}{\textbf{L3.1-8B-I}} & \multicolumn{3}{c}{\textbf{Qwen3-8B}}\\\cmidrule(lr){3-5}\cmidrule(lr){6-8}
\textbf{Weights} & \textbf{Budget / method} & \textbf{PPL} & \textbf{KL} & \textbf{tok/s} & \textbf{PPL} & \textbf{KL} & \textbf{tok/s}\\\midrule
}
\tabletail{\bottomrule}\tablelasttail{\bottomrule}
\begin{supertabular*}{\linewidth}{@{\extracolsep{\fill}}@{}llcccccc@{}}
BF16 & Base & 7.22 & 0.000 & 104.7 & 9.73 & 0.000 & 103.0\\
\midrule

\appendixgroupreserve{9}
\multirow[c]{9}{*}{YAQA 2-bit} & Base & 9.46 & 0.257 & 245.8 & 10.95 & 0.214 & 232.3\\
 & 2.5\% / DecDEC & 8.90 & 0.197 & 239.4 & 10.57 & 0.172 & 223.1\\
 & 2.5\% / \methodname{}-W & 8.49 & 0.149 & 240.3 & 10.66 & 0.144 & 227.7\\
 & 5\% / DecDEC & 9.03 & 0.211 & 233.2 & 10.69 & 0.184 & 220.8\\
 & 5\% / \methodname{}-W & 8.40 & 0.139 & 234.4 & 10.51 & 0.132 & 224.4\\
 & 10\% / DecDEC & 8.87 & 0.193 & 228.9 & 10.53 & 0.170 & 217.0\\
 & 10\% / \methodname{}-W & 8.39 & 0.137 & 223.9 & 10.43 & 0.123 & 209.6\\
 & 20\% / DecDEC & 8.78 & 0.183 & 218.3 & 10.46 & 0.163 & 205.5\\
 & 20\% / \methodname{}-W & 8.37 & 0.134 & 211.6 & 10.37 & 0.120 & 192.9\\
\midrule
\appendixgroupreserve{9}
\multirow[c]{9}{*}{YAQA 3-bit} & Base & 7.61 & 0.047 & 222.5 & 9.98 & 0.052 & 213.1\\
 & 2.5\% / DecDEC & 7.57 & 0.041 & 215.1 & 9.93 & 0.046 & 206.1\\
 & 2.5\% / \methodname{}-W & 7.50 & 0.034 & 217.5 & 9.89 & 0.036 & 208.3\\
 & 5\% / DecDEC & 7.55 & 0.040 & 213.3 & 9.89 & 0.044 & 202.0\\
 & 5\% / \methodname{}-W & 7.49 & 0.033 & 212.7 & 9.89 & 0.033 & 202.8\\
 & 10\% / DecDEC & 7.54 & 0.038 & 209.3 & 9.88 & 0.042 & 198.8\\
 & 10\% / \methodname{}-W & 7.48 & 0.032 & 202.6 & 9.87 & 0.032 & 194.2\\
 & 20\% / DecDEC & 7.52 & 0.037 & 197.5 & 9.87 & 0.041 & 191.7\\
 & 20\% / \methodname{}-W & 7.47 & 0.030 & 184.8 & 9.86 & 0.031 & 171.5\\
\midrule
\appendixgroupreserve{9}
\multirow[c]{9}{*}{LNQ+GQ 3-bit} & Base & 8.52 & 0.157 & 314.3 & 10.92 & 0.184 & 290.9\\
 & 2.5\% / DecDEC & 8.40 & 0.146 & 312.4 & 10.78 & 0.187 & 290.5\\
 & 2.5\% / \methodname{}-W & 8.20 & 0.122 & 306.0 & 10.51 & 0.144 & 284.5\\
 & 5\% / DecDEC & 8.37 & 0.142 & 311.5 & 10.66 & 0.170 & 287.1\\
 & 5\% / \methodname{}-W & 8.18 & 0.121 & 299.7 & 10.47 & 0.139 & 281.1\\
 & 10\% / DecDEC & 8.32 & 0.138 & 302.8 & 10.55 & 0.159 & 283.0\\
 & 10\% / \methodname{}-W & 8.11 & 0.112 & 285.8 & 10.54 & 0.130 & 263.7\\
 & 20\% / DecDEC & 8.39 & 0.147 & 287.4 & 10.85 & 0.181 & 265.6\\
 & 20\% / \methodname{}-W & 8.07 & 0.105 & 261.6 & 10.42 & 0.117 & 240.4\\
\midrule
\appendixgroupreserve{9}
\multirow[c]{9}{*}{AWQ 3-bit} & Base & 9.10 & 0.229 & 304.5 & 11.88 & 0.284 & 286.0\\
 & 2.5\% / DecDEC & 8.88 & 0.204 & 294.4 & 11.63 & 0.265 & 281.5\\
 & 2.5\% / \methodname{}-W & 8.60 & 0.170 & 295.0 & 11.24 & 0.208 & 273.9\\
 & 5\% / DecDEC & 8.80 & 0.195 & 292.6 & 11.38 & 0.231 & 276.1\\
 & 5\% / \methodname{}-W & 8.38 & 0.145 & 289.0 & 10.71 & 0.167 & 271.5\\
 & 10\% / DecDEC & 8.76 & 0.189 & 289.5 & 11.24 & 0.220 & 273.2\\
 & 10\% / \methodname{}-W & 8.31 & 0.140 & 270.9 & 10.49 & 0.143 & 259.0\\
 & 20\% / DecDEC & 8.56 & 0.166 & 270.2 & 11.21 & 0.218 & 253.8\\
 & 20\% / \methodname{}-W & 8.19 & 0.120 & 255.7 & 10.50 & 0.144 & 238.3\\
\midrule
\appendixgroupreserve{9}
\multirow[c]{9}{*}{SqLLM 3-bit} & Base & 9.36 & 0.263 & 315.4 & 11.71 & 0.255 & 292.1\\
 & 2.5\% / DecDEC & 9.10 & 0.233 & 312.2 & 11.42 & 0.234 & 291.2\\
 & 2.5\% / \methodname{}-W & 8.47 & 0.162 & 307.3 & 10.85 & 0.174 & 267.1\\
 & 5\% / DecDEC & 9.02 & 0.225 & 311.3 & 11.36 & 0.215 & 287.3\\
 & 5\% / \methodname{}-W & 8.42 & 0.156 & 299.6 & 10.81 & 0.169 & 280.3\\
 & 10\% / DecDEC & 8.91 & 0.213 & 305.2 & 11.20 & 0.201 & 283.8\\
 & 10\% / \methodname{}-W & 8.33 & 0.143 & 288.6 & 10.62 & 0.145 & 264.0\\
 & 20\% / DecDEC & 8.86 & 0.209 & 289.1 & 11.41 & 0.216 & 266.2\\
 & 20\% / \methodname{}-W & 8.13 & 0.121 & 263.6 & 10.53 & 0.132 & 244.9\\
\midrule
\appendixgroupreserve{9}
\multirow[c]{9}{*}{YAQA 4-bit} & Base & 7.33 & 0.014 & 204.2 & 9.79 & 0.016 & 193.8\\
 & 2.5\% / DecDEC & 7.30 & 0.012 & 197.6 & 9.76 & 0.015 & 185.9\\
 & 2.5\% / \methodname{}-W & 7.29 & 0.010 & 199.1 & 9.77 & 0.011 & 189.0\\
 & 5\% / DecDEC & 7.30 & 0.011 & 195.9 & 9.75 & 0.013 & 183.3\\
 & 5\% / \methodname{}-W & 7.29 & 0.010 & 194.3 & 9.75 & 0.010 & 184.1\\
 & 10\% / DecDEC & 7.30 & 0.011 & 190.6 & 9.74 & 0.013 & 180.6\\
 & 10\% / \methodname{}-W & 7.29 & 0.010 & 185.5 & 9.75 & 0.010 & 176.0\\
 & 20\% / DecDEC & 7.29 & 0.010 & 180.5 & 9.74 & 0.012 & 172.5\\
 & 20\% / \methodname{}-W & 7.29 & 0.009 & 169.5 & 9.74 & 0.010 & 158.4\\
\midrule
\appendixgroupreserve{9}
\multirow[c]{9}{*}{LNQ+GQ 4-bit} & Base & 7.48 & 0.034 & 264.7 & 9.95 & 0.044 & 248.3\\
 & 2.5\% / DecDEC & 7.43 & 0.031 & 263.8 & 9.89 & 0.038 & 247.6\\
 & 2.5\% / \methodname{}-W & 7.42 & 0.028 & 258.9 & 9.88 & 0.036 & 242.5\\
 & 5\% / DecDEC & 7.42 & 0.031 & 263.1 & 9.89 & 0.038 & 244.9\\
 & 5\% / \methodname{}-W & 7.41 & 0.027 & 254.5 & 9.88 & 0.034 & 238.0\\
 & 10\% / DecDEC & 7.45 & 0.032 & 249.9 & 9.89 & 0.041 & 236.7\\
 & 10\% / \methodname{}-W & 7.40 & 0.027 & 241.9 & 9.89 & 0.033 & 225.9\\
 & 20\% / DecDEC & 7.45 & 0.032 & 238.8 & 9.96 & 0.043 & 224.2\\
 & 20\% / \methodname{}-W & 7.39 & 0.025 & 218.9 & 9.89 & 0.031 & 193.4\\
\midrule
\appendixgroupreserve{9}
\multirow[c]{9}{*}{AWQ 4-bit} & Base & 7.59 & 0.049 & 262.8 & 10.01 & 0.059 & 247.5\\
 & 2.5\% / DecDEC & 7.53 & 0.043 & 259.5 & 9.95 & 0.055 & 245.2\\
 & 2.5\% / \methodname{}-W & 7.45 & 0.033 & 257.4 & 9.88 & 0.040 & 241.4\\
 & 5\% / DecDEC & 7.52 & 0.042 & 257.7 & 9.92 & 0.049 & 242.2\\
 & 5\% / \methodname{}-W & 7.44 & 0.031 & 250.4 & 9.88 & 0.038 & 236.9\\
 & 10\% / DecDEC & 7.51 & 0.041 & 254.1 & 9.90 & 0.047 & 238.0\\
 & 10\% / \methodname{}-W & 7.43 & 0.029 & 237.0 & 9.81 & 0.034 & 223.0\\
 & 20\% / DecDEC & 7.47 & 0.036 & 236.2 & 9.89 & 0.045 & 226.1\\
 & 20\% / \methodname{}-W & 7.40 & 0.028 & 218.4 & 9.79 & 0.031 & 206.0\\
\midrule
\appendixgroupreserve{9}
\multirow[c]{9}{*}{SqLLM 4-bit} & Base & 7.56 & 0.052 & 265.6 & 9.98 & 0.062 & 249.0\\
 & 2.5\% / DecDEC & 7.51 & 0.044 & 262.9 & 9.87 & 0.050 & 246.4\\
 & 2.5\% / \methodname{}-W & 7.46 & 0.037 & 257.9 & 9.89 & 0.042 & 239.3\\
 & 5\% / DecDEC & 7.51 & 0.044 & 262.2 & 9.86 & 0.049 & 245.0\\
 & 5\% / \methodname{}-W & 7.44 & 0.035 & 254.6 & 9.85 & 0.038 & 235.7\\
 & 10\% / DecDEC & 7.52 & 0.044 & 254.6 & 9.87 & 0.050 & 234.8\\
 & 10\% / \methodname{}-W & 7.42 & 0.032 & 241.9 & 9.81 & 0.038 & 226.3\\
 & 20\% / DecDEC & 7.52 & 0.044 & 238.8 & 9.88 & 0.051 & 222.2\\
 & 20\% / \methodname{}-W & 7.41 & 0.029 & 216.4 & 9.81 & 0.033 & 205.8\cr
\end{supertabular*}

\end{appendixflowtable}

\appendixfirstgroupreserve{11}
\begin{appendixflowtable}
\caption{Weight-only refinement on RTX 5070 Ti: 2--4-bit bases across slowdown budgets. Columns report WikiText-2 PPL/KL and empty-context decode tok/s.}
\label{tab:weight_budgets-3}\label{tab:weight_budgets-4}
\fontsize{6.8}{8.2}\selectfont\setlength{\tabcolsep}{2pt}\renewcommand{\arraystretch}{1.0}\setlength{\aboverulesep}{.2ex}\setlength{\belowrulesep}{.25ex}
\tablehead{\toprule
\multicolumn{8}{@{}l}{\textbf{RTX 5070 Ti}}\\
 & & \multicolumn{3}{c}{\textbf{L3.1-8B-I}} & \multicolumn{3}{c}{\textbf{Qwen3-8B}}\\\cmidrule(lr){3-5}\cmidrule(lr){6-8}
\textbf{Weights} & \textbf{Budget / method} & \textbf{PPL} & \textbf{KL} & \textbf{tok/s} & \textbf{PPL} & \textbf{KL} & \textbf{tok/s}\\\midrule
}
\tabletail{\bottomrule}\tablelasttail{\bottomrule}
\begin{supertabular*}{\linewidth}{@{\extracolsep{\fill}}@{}llcccccc@{}}
\appendixgroupreserve{9}
\multirow[c]{9}{*}{YAQA 2-bit} & Base & 9.46 & 0.257 & 165.9 & 10.95 & 0.214 & 158.7\\
 & 2.5\% / DecDEC & 9.06 & 0.215 & 155.8 & 10.58 & 0.175 & 146.4\\
 & 2.5\% / \methodname{}-W & 8.54 & 0.153 & 158.9 & 10.54 & 0.136 & 149.8\\
 & 5\% / DecDEC & 8.97 & 0.205 & 152.8 & 10.56 & 0.173 & 144.6\\
 & 5\% / \methodname{}-W & 8.49 & 0.147 & 154.2 & 10.45 & 0.129 & 149.4\\
 & 10\% / DecDEC & 8.89 & 0.196 & 149.5 & 10.54 & 0.171 & 141.0\\
 & 10\% / \methodname{}-W & 8.37 & 0.135 & 148.9 & 10.36 & 0.120 & 143.5\\
 & 20\% / DecDEC & 8.61 & 0.164 & 139.1 & 10.37 & 0.151 & 134.1\\
 & 20\% / \methodname{}-W & 8.30 & 0.127 & 135.5 & 10.34 & 0.115 & 130.8\\
\midrule
\appendixgroupreserve{9}
\multirow[c]{9}{*}{YAQA 3-bit} & Base & 7.61 & 0.047 & 149.4 & 9.98 & 0.052 & 145.1\\
 & 2.5\% / DecDEC & 7.57 & 0.041 & 141.7 & 9.90 & 0.044 & 136.9\\
 & 2.5\% / \methodname{}-W & 7.51 & 0.034 & 144.5 & 9.90 & 0.035 & 138.6\\
 & 5\% / DecDEC & 7.56 & 0.040 & 138.4 & 9.89 & 0.043 & 133.8\\
 & 5\% / \methodname{}-W & 7.49 & 0.032 & 141.0 & 9.88 & 0.034 & 137.6\\
 & 10\% / DecDEC & 7.51 & 0.035 & 133.8 & 9.88 & 0.042 & 129.1\\
 & 10\% / \methodname{}-W & 7.47 & 0.030 & 135.5 & 9.87 & 0.031 & 130.0\\
 & 20\% / DecDEC & 7.48 & 0.032 & 125.0 & 9.83 & 0.037 & 120.5\\
 & 20\% / \methodname{}-W & 7.45 & 0.028 & 124.2 & 9.85 & 0.030 & 120.0\\
\midrule
\appendixgroupreserve{9}
\multirow[c]{9}{*}{LNQ+GQ 3-bit} & Base & 8.52 & 0.157 & 183.1 & 10.92 & 0.184 & 172.5\\
 & 2.5\% / DecDEC & 8.23 & 0.126 & 179.4 & 10.42 & 0.145 & 169.4\\
 & 2.5\% / \methodname{}-W & 8.22 & 0.123 & 176.3 & 10.48 & 0.123 & 167.0\\
 & 5\% / DecDEC & 8.33 & 0.141 & 176.1 & 10.69 & 0.167 & 165.9\\
 & 5\% / \methodname{}-W & 8.03 & 0.103 & 170.7 & 10.51 & 0.126 & 163.3\\
 & 10\% / DecDEC & 8.33 & 0.141 & 170.1 & 10.70 & 0.171 & 160.8\\
 & 10\% / \methodname{}-W & 8.06 & 0.103 & 165.2 & 10.42 & 0.117 & 157.1\\
 & 20\% / DecDEC & 8.27 & 0.133 & 161.0 & 10.54 & 0.152 & 152.4\\
 & 20\% / \methodname{}-W & 8.00 & 0.098 & 147.8 & 10.39 & 0.109 & 139.6\\
\midrule
\appendixgroupreserve{9}
\multirow[c]{9}{*}{AWQ 3-bit} & Base & 9.10 & 0.229 & 171.0 & 11.88 & 0.284 & 163.2\\
 & 2.5\% / DecDEC & 8.86 & 0.202 & 167.2 & 11.53 & 0.243 & 156.9\\
 & 2.5\% / \methodname{}-W & 8.52 & 0.163 & 166.4 & 11.14 & 0.204 & 156.5\\
 & 5\% / DecDEC & 8.60 & 0.172 & 162.9 & 11.07 & 0.207 & 154.5\\
 & 5\% / \methodname{}-W & 8.35 & 0.142 & 161.4 & 10.69 & 0.153 & 154.0\\
 & 10\% / DecDEC & 8.53 & 0.162 & 156.5 & 11.04 & 0.203 & 148.6\\
 & 10\% / \methodname{}-W & 8.17 & 0.117 & 154.2 & 10.40 & 0.128 & 147.4\\
 & 20\% / DecDEC & 8.40 & 0.146 & 147.7 & 10.71 & 0.170 & 140.4\\
 & 20\% / \methodname{}-W & 8.04 & 0.102 & 141.7 & 10.33 & 0.114 & 134.0\\
\midrule
\appendixgroupreserve{9}
\multirow[c]{9}{*}{SqLLM 3-bit} & Base & 9.36 & 0.263 & 183.1 & 11.71 & 0.255 & 172.6\\
 & 2.5\% / DecDEC & 9.09 & 0.234 & 179.2 & 10.90 & 0.177 & 169.4\\
 & 2.5\% / \methodname{}-W & 8.44 & 0.158 & 173.0 & 11.02 & 0.179 & 164.1\\
 & 5\% / DecDEC & 8.64 & 0.183 & 175.8 & 11.21 & 0.192 & 166.0\\
 & 5\% / \methodname{}-W & 8.17 & 0.129 & 166.7 & 10.60 & 0.138 & 163.6\\
 & 10\% / DecDEC & 8.71 & 0.191 & 170.3 & 11.27 & 0.199 & 160.2\\
 & 10\% / \methodname{}-W & 8.08 & 0.114 & 166.3 & 10.50 & 0.125 & 153.0\\
 & 20\% / DecDEC & 8.32 & 0.145 & 160.9 & 10.71 & 0.150 & 152.5\\
 & 20\% / \methodname{}-W & 7.97 & 0.102 & 149.0 & 10.38 & 0.114 & 142.9\\
\midrule
\appendixgroupreserve{9}
\multirow[c]{9}{*}{YAQA 4-bit} & Base & 7.33 & 0.014 & 132.2 & 9.79 & 0.016 & 128.0\\
 & 2.5\% / DecDEC & 7.30 & 0.012 & 126.8 & 9.75 & 0.014 & 120.8\\
 & 2.5\% / \methodname{}-W & 7.28 & 0.009 & 127.7 & 9.76 & 0.011 & 123.1\\
 & 5\% / DecDEC & 7.29 & 0.010 & 123.5 & 9.74 & 0.013 & 118.1\\
 & 5\% / \methodname{}-W & 7.29 & 0.009 & 124.2 & 9.75 & 0.010 & 120.6\\
 & 10\% / DecDEC & 7.28 & 0.009 & 118.2 & 9.73 & 0.011 & 113.6\\
 & 10\% / \methodname{}-W & 7.28 & 0.008 & 119.2 & 9.75 & 0.010 & 115.8\\
 & 20\% / DecDEC & 7.28 & 0.009 & 110.6 & 9.73 & 0.011 & 106.4\\
 & 20\% / \methodname{}-W & 7.28 & 0.008 & 109.2 & 9.75 & 0.010 & 106.2\\
\midrule
\appendixgroupreserve{9}
\multirow[c]{9}{*}{LNQ+GQ 4-bit} & Base & 7.48 & 0.034 & 148.3 & 9.95 & 0.044 & 141.3\\
 & 2.5\% / DecDEC & 7.43 & 0.030 & 144.3 & 9.89 & 0.041 & 139.0\\
 & 2.5\% / \methodname{}-W & 7.38 & 0.024 & 142.1 & 9.90 & 0.032 & 137.8\\
 & 5\% / DecDEC & 7.42 & 0.029 & 142.2 & 9.85 & 0.039 & 136.5\\
 & 5\% / \methodname{}-W & 7.38 & 0.024 & 139.0 & 9.88 & 0.031 & 133.8\\
 & 10\% / DecDEC & 7.41 & 0.028 & 137.5 & 9.83 & 0.037 & 132.4\\
 & 10\% / \methodname{}-W & 7.38 & 0.023 & 133.4 & 9.88 & 0.030 & 129.5\\
 & 20\% / DecDEC & 7.40 & 0.026 & 129.0 & 9.83 & 0.035 & 124.5\\
 & 20\% / \methodname{}-W & 7.37 & 0.022 & 121.7 & 9.87 & 0.029 & 118.7\\
\midrule
\appendixgroupreserve{9}
\multirow[c]{9}{*}{AWQ 4-bit} & Base & 7.59 & 0.049 & 152.4 & 10.01 & 0.059 & 144.8\\
 & 2.5\% / DecDEC & 7.47 & 0.036 & 148.4 & 9.88 & 0.043 & 140.8\\
 & 2.5\% / \methodname{}-W & 7.41 & 0.029 & 147.2 & 9.86 & 0.035 & 139.7\\
 & 5\% / DecDEC & 7.48 & 0.037 & 145.7 & 9.95 & 0.048 & 138.6\\
 & 5\% / \methodname{}-W & 7.40 & 0.026 & 143.1 & 9.80 & 0.032 & 136.2\\
 & 10\% / DecDEC & 7.44 & 0.031 & 140.8 & 9.83 & 0.037 & 133.6\\
 & 10\% / \methodname{}-W & 7.38 & 0.024 & 136.5 & 9.77 & 0.029 & 130.6\\
 & 20\% / DecDEC & 7.40 & 0.026 & 132.1 & 9.80 & 0.034 & 126.1\\
 & 20\% / \methodname{}-W & 7.36 & 0.021 & 126.8 & 9.84 & 0.026 & 119.5\\
\midrule
\appendixgroupreserve{9}
\multirow[c]{9}{*}{SqLLM 4-bit} & Base & 7.56 & 0.052 & 148.3 & 9.98 & 0.062 & 141.4\\
 & 2.5\% / DecDEC & 7.46 & 0.037 & 144.5 & 9.85 & 0.048 & 138.9\\
 & 2.5\% / \methodname{}-W & 7.38 & 0.028 & 142.5 & 9.83 & 0.036 & 137.3\\
 & 5\% / DecDEC & 7.44 & 0.035 & 142.2 & 9.79 & 0.041 & 136.4\\
 & 5\% / \methodname{}-W & 7.38 & 0.026 & 139.4 & 9.78 & 0.032 & 134.3\\
 & 10\% / DecDEC & 7.41 & 0.031 & 137.5 & 9.77 & 0.038 & 132.3\\
 & 10\% / \methodname{}-W & 7.37 & 0.025 & 134.0 & 9.79 & 0.030 & 129.8\\
 & 20\% / DecDEC & 7.39 & 0.029 & 129.0 & 9.75 & 0.034 & 124.4\\
 & 20\% / \methodname{}-W & 7.36 & 0.023 & 123.7 & 9.79 & 0.027 & 118.3\cr
\end{supertabular*}

\end{appendixflowtable}

\begin{figure}[!tb]
\centering
\includegraphics[width=\linewidth]{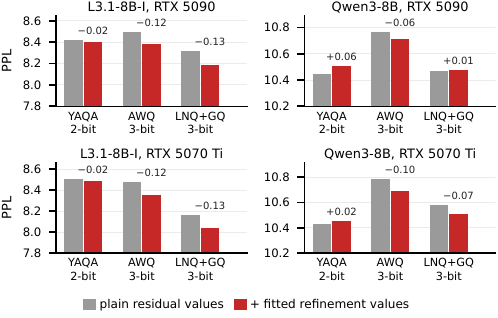}
\caption{Fitted versus unfitted weight refinement on WikiText-2, at the 5\% configurations tuned on RTX 5090 and RTX 5070 Ti. Each pair fixes selection counts, schedule, and MLP scheme: gray bars use plain residuals; red bars use fitted refinement. Labels give the PPL change from fitting.}
\label{fig:weight-fitting-ablation}
\end{figure}

\subsection{YAQA Recovery Fine-Tuning}
\label{sec:yaqa-recovery-finetuning}
\methodname{}-W is compatible with YAQA recovery fine-tuning. We use QTIP\textquotesingle{}s RedPajama fine-tuning recipe, but freeze embeddings, final normalization, and the output head during end-to-end fine-tuning; Qwen3-8B's query/key norms remain frozen during both block-wise and end-to-end fine-tuning. This preserves the BF16 refinement target; normalization changes elsewhere are absorbed by rescaling its projection weights. We find that this preserves the gains of fine-tuning while maintaining compatibility with weight refinement (Table~\ref{tab:weight_finetuning-1}).
\begin{appendixcolumntable}
\centering
\caption{Weight-only refinement after YAQA recovery fine-tuning: WikiText-2 PPL/KL using configurations tuned on RTX 5090. Original and fine-tuned bases are shown for each bitwidth; both DecDEC and \methodname{}-W refine the fine-tuned weights.}
\label{tab:weight_finetuning-1}
\fontsize{7}{8.7}\selectfont\setlength{\tabcolsep}{.7pt}\renewcommand{\arraystretch}{1.0}\setlength{\aboverulesep}{.2ex}\setlength{\belowrulesep}{.25ex}
\setlength{\aboverulesep}{.35ex}\setlength{\belowrulesep}{.35ex}
\begin{tabular*}{\linewidth}{@{\extracolsep{\fill}}@{}llccccc@{}}\toprule
 & & & \multicolumn{2}{c}{\textbf{L3.1-8B-I}} & \multicolumn{2}{c}{\textbf{Qwen3-8B}}\\
\cmidrule(lr){4-5}\cmidrule(lr){6-7}
\textbf{Weights} & \textbf{Method} & \textbf{Budget} & \textbf{PPL} & \textbf{KL} & \textbf{PPL} & \textbf{KL}\\\midrule
\multirow[c]{10}{*}[-4.8pt]{YAQA 2-bit} & Original base & & 9.46 & 0.257 & 10.95 & 0.214\\
 & Fine-tuned base & & 8.45 & 0.161 & 10.24 & 0.162\\
\cmidrule(lr){2-7}
 & DecDEC & \multirow[c]{2}{*}{2.5\%} & 8.32 & 0.142 & 10.22 & 0.146\\
 & \methodname{}-W &  & 8.24 & 0.133 & 10.50 & 0.137\\
\cmidrule(lr){2-7}
 & DecDEC & \multirow[c]{2}{*}{5\%} & 8.35 & 0.149 & 10.26 & 0.153\\
 & \methodname{}-W &  & 8.19 & 0.128 & 10.40 & 0.125\\
\cmidrule(lr){2-7}
 & DecDEC & \multirow[c]{2}{*}{10\%} & 8.32 & 0.141 & 10.21 & 0.146\\
 & \methodname{}-W &  & 8.18 & 0.126 & 10.37 & 0.120\\
\cmidrule(lr){2-7}
 & DecDEC & \multirow[c]{2}{*}{20\%} & 8.28 & 0.137 & 10.14 & 0.140\\
 & \methodname{}-W &  & 8.22 & 0.126 & 10.36 & 0.117\\
\midrule
\multirow[c]{10}{*}[-4.8pt]{YAQA 3-bit} & Original base & & 7.61 & 0.047 & 9.98 & 0.052\\
 & Fine-tuned base & & 7.50 & 0.040 & 9.73 & 0.046\\
\cmidrule(lr){2-7}
 & DecDEC & \multirow[c]{2}{*}{2.5\%} & 7.48 & 0.036 & 9.75 & 0.043\\
 & \methodname{}-W &  & 7.48 & 0.032 & 9.78 & 0.035\\
\cmidrule(lr){2-7}
 & DecDEC & \multirow[c]{2}{*}{5\%} & 7.48 & 0.035 & 9.75 & 0.041\\
 & \methodname{}-W &  & 7.47 & 0.032 & 9.79 & 0.032\\
\cmidrule(lr){2-7}
 & DecDEC & \multirow[c]{2}{*}{10\%} & 7.48 & 0.035 & 9.74 & 0.040\\
 & \methodname{}-W &  & 7.47 & 0.031 & 9.77 & 0.032\\
\cmidrule(lr){2-7}
 & DecDEC & \multirow[c]{2}{*}{20\%} & 7.47 & 0.033 & 9.73 & 0.039\\
 & \methodname{}-W &  & 7.45 & 0.029 & 9.76 & 0.031\\
\midrule
\multirow[c]{10}{*}[-4.8pt]{YAQA 4-bit} & Original base & & 7.33 & 0.014 & 9.79 & 0.016\\
 & Fine-tuned base & & 7.28 & 0.013 & 9.71 & 0.015\\
\cmidrule(lr){2-7}
 & DecDEC & \multirow[c]{2}{*}{2.5\%} & 7.27 & 0.011 & 9.70 & 0.014\\
 & \methodname{}-W &  & 7.27 & 0.010 & 9.73 & 0.011\\
\cmidrule(lr){2-7}
 & DecDEC & \multirow[c]{2}{*}{5\%} & 7.27 & 0.011 & 9.69 & 0.013\\
 & \methodname{}-W &  & 7.28 & 0.009 & 9.74 & 0.010\\
\cmidrule(lr){2-7}
 & DecDEC & \multirow[c]{2}{*}{10\%} & 7.27 & 0.011 & 9.69 & 0.013\\
 & \methodname{}-W &  & 7.27 & 0.009 & 9.74 & 0.010\\
\cmidrule(lr){2-7}
 & DecDEC & \multirow[c]{2}{*}{20\%} & 7.26 & 0.010 & 9.69 & 0.012\\
 & \methodname{}-W &  & 7.27 & 0.009 & 9.74 & 0.010\\
\bottomrule\end{tabular*}
\end{appendixcolumntable}

\subsection{Refinement and Cache Ablations}
\begin{figure}[!tb]
\centering
\includegraphics[width=\linewidth]{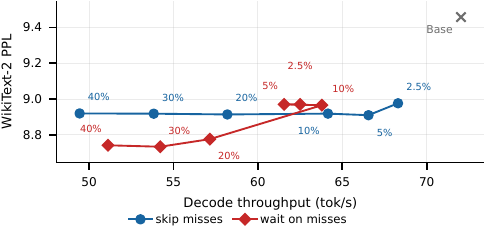}
\caption{Waiting versus skipping weight-cache misses on Llama-3.1-8B-Instruct with YAQA 2-bit weights, DGX Spark, and a 128 MiB cache. Curves plot WikiText-2 PPL against throughput. Points follow slowdown budgets of 2.5\%, 5\%, 10\%, 20\%, 30\%, and 40\%. Measured wait-policy throughput is non-monotonic across budgets.}
\label{fig:weight-wait-ablation}
\end{figure}

Figure~\ref{fig:weight-fitting-ablation} compares fitted and unfitted weight refinement at fixed configurations. Figure~\ref{fig:weight-wait-ablation} compares skipping and waiting on weight-cache misses across slowdown budgets.
Tables~\ref{tab:weight-cache-measurements} and~\ref{tab:paired-weight-ablations} give the cache measurements and paired technique ablations.

\begin{appendixcolumntable}\centering
\begingroup\setlength{\tabcolsep}{1.5pt}\renewcommand{\arraystretch}{1.0}\setlength{\aboverulesep}{.2ex}\setlength{\belowrulesep}{.25ex}\fontsize{7}{8.3}\selectfont
\caption{Weight-cache capacity and miss policy on DGX Spark: Llama-3.1-8B-Instruct, 5\% configuration, YAQA 2-bit and AWQ 3-bit. WikiText-2 PPL/KL and decode tok/s; capacities in MiB. Raw values behind Figure~\ref{main-fig:cache-ablation}(d,e).}\label{tab:weight-cache-measurements}
\begin{tabular*}{\linewidth}{@{\extracolsep{\fill}}@{}llcccccc@{}}
\toprule
 & & \multicolumn{3}{c}{\textbf{YAQA 2-bit}} & \multicolumn{3}{c}{\textbf{AWQ 3-bit}}\\
\cmidrule(lr){3-5}\cmidrule(lr){6-8}
\textbf{MiB} & \textbf{Miss policy} & \textbf{PPL} & \textbf{KL} & \textbf{tok/s} & \textbf{PPL} & \textbf{KL} & \textbf{tok/s}\\\midrule

64 & skip & 9.01 & 0.207 & 68.8 & 8.70 & 0.186 & 57.4\\
128 & skip & 8.94 & 0.200 & 69.2 & 8.64 & 0.179 & 57.6\\
256 & skip & 8.88 & 0.192 & 69.7 & 8.59 & 0.172 & 57.6\\
512 & skip & 8.80 & 0.183 & 69.4 & 8.53 & 0.166 & 57.4\\
128 & wait & 8.52 & 0.152 & 17.9 & 8.31 & 0.137 & 27.7\\
\bottomrule\end{tabular*}\endgroup
\end{appendixcolumntable}
\begin{table}[!htbp]\centering
\begingroup\setlength{\tabcolsep}{.25pt}\renewcommand{\arraystretch}{1.0}\setlength{\aboverulesep}{.2ex}\setlength{\belowrulesep}{.25ex}\fontsize{6.2}{7.6}\selectfont
\caption{Paired weight-refinement ablations on Llama-3.1-8B-Instruct, RTX 5090: WikiText-2 PPL/KL and empty-context decode tok/s. Pairs are tuned separately at a 5\% slowdown budget, except in the unchanged-settings pair: there, $s=0$ disables lead/lag in the final configuration while retaining its $k$, fitted refinements, MLP scheme, and GPU thread-block counts.}\label{tab:paired-weight-ablations}
\begin{tabular*}{\linewidth}{@{\extracolsep{\fill}}@{}llcccccc@{}}
\toprule
 & & \multicolumn{3}{c}{\textbf{YAQA 2-bit}} & \multicolumn{3}{c}{\textbf{AWQ 3-bit}}\\
\cmidrule(lr){3-5}\cmidrule(lr){6-8}
\textbf{Comparison} & \textbf{Setting} & \textbf{PPL} & \textbf{KL} & \textbf{tok/s} & \textbf{PPL} & \textbf{KL} & \textbf{tok/s}\\\midrule

\multirow[c]{2}{*}{\mbox{Fitting, $s=0$}} & Unfitted & 8.76 & 0.181 & 234.1 & 8.70 & 0.182 & 288.1\\
 & Fitted & 8.67 & 0.170 & 230.1 & 8.61 & 0.171 & 287.5\\
\midrule
\multirow[c]{2}{*}{\mbox{Fitting, tuned $s$}} & Unfitted & 8.68 & 0.174 & 234.9 & 8.50 & 0.158 & 290.2\\
 & Fitted & 8.59 & 0.162 & 234.3 & 8.38 & 0.145 & 289.0\\
\midrule
\multirow[c]{2}{*}{\mbox{MLP scheme}} & Channel-wise & 8.59 & 0.162 & 234.3 & 8.38 & 0.145 & 289.0\\
 & \mbox{Neuron bundles} & 8.40 & 0.139 & 234.4 & 8.56 & 0.165 & 289.3\\
\midrule
\multirow[c]{2}{*}{\mbox{Lead/lag, unfitted}} & $s=0$ & 8.76 & 0.181 & 234.1 & 8.70 & 0.182 & 288.1\\
 & Tuned $s$ & 8.68 & 0.174 & 234.9 & 8.50 & 0.158 & 290.2\\
\midrule
\multirow[c]{2}{*}{\mbox{Lead/lag, full method}} & $s=0$ & 8.57 & 0.157 & 232.2 & 8.57 & 0.168 & 286.8\\
 & Tuned $s$ & 8.40 & 0.139 & 234.4 & 8.38 & 0.145 & 289.0\\
\midrule
\multirow[c]{2}{*}{\mbox{Lead/lag, unchanged settings}} & $s=0$ & 8.34 & 0.131 & 182.2 & 8.33 & 0.141 & 271.1\\
 & Tuned $s$ & 8.40 & 0.139 & 234.4 & 8.38 & 0.145 & 289.0\\
\bottomrule\end{tabular*}
\endgroup
\end{table}

\subsection{Weight Refinement Tuner}
\label{sec:weight-tuning-configurations}
The tuner searches separately for each model, weight quantizer, and platform under a specified slowdown budget relative to the quantized base. It selects channel budgets $k$, fetch lead/lag distances $s$, the MLP refinement scheme, and GPU thread-block counts.

\textbf{Resource allocation.} We first profile base projections alongside refinement while varying its thread-block count and selection budget. The resulting timing table identifies launch configurations that balance refinement speed against interference with the base projections.

\textbf{Budget-constrained search.} For each combination of shift distances and channel-wise or neuron-wise MLP refinement, the tuner searches for the largest common channel budget $k$ that satisfies the slowdown limit in end-to-end decoding. For neuron-wise MLP refinement, it sets $k_n=\mathrm{round}(3k/4)$: each increment of $k_n$ fetches $4/3$ as many bytes as each increment of $k$ for channel-wise gate/up and down refinement. It also considers disabling refinement for projection sets when needed to meet the budget. On DGX Spark, we search skip-policy configurations only at $s=0$: skipping cache misses leaves no SSD wait for fetch lead/lag to hide.

\textbf{Accuracy selection and budget allocation.} The tuner compares feasible candidates by KL divergence from BF16 on held-out text. Starting from the lowest-KL candidate, it increases individual selection budgets while remeasuring end-to-end latency and retaining increases that satisfy the slowdown limit. These final per-projection budgets need not be equal.

\textbf{Refinement fitting during search.} Fitted refinement values depend on the selection budget (Section~\ref{main-sec:weight-streaming}). We approximate candidate KL during search using refinements pre-fitted at a ladder of power-of-two selection budgets, avoiding a separate fit for each candidate. Once the final per-projection budgets are selected, we fit refinement values at those exact budgets before quantizing them for inference.

\par\medskip\noindent\textbf{Tuned configurations.}
Table~\ref{tab:weight-tuning} lists tuned configurations across slowdown budgets and miss-policy sweeps. Each row displays one configuration. The Weights column gives the quantizer and bitwidth. Four-entry tuples give the query/key/value, attention-output, gate/up, and down configurations for channel-wise refinement. Three-entry tuples give query/key/value, attention-output, and neuron bundles. This order applies to selection counts ($k$, or $k_n$ for neuron bundles) and shift distances $s$. $n_{\mathrm{tb}}$ lists the GPU thread-block count shared by channel-wise kernels, followed by the count for the neuron-bundle kernel when used. Counts $k$ are per 1,024 input channels or neurons, and $s$ is in Transformer blocks. SSD-backed configurations use a 128 MiB cache; miss-policy sweeps identify skipping or waiting.
\par\bigskip
\gdef\weighttuningcaption{Tuned weight-refinement configurations. Model and platform headings identify each configuration block. The main DGX Spark skip-policy search uses only $s=0$; skip/wait policy sweeps are listed separately.}
{\fontsize{7.5}{8.5}\selectfont\setlength{\tabcolsep}{1.6pt}\renewcommand{\arraystretch}{1.0}\setlength{\aboverulesep}{.2ex}\setlength{\belowrulesep}{.25ex}
\tablecaption{\weighttuningcaption}\label{tab:weight-tuning}
\tablefirsthead{\toprule\multicolumn{5}{@{}p{\linewidth}@{}}{\textbf{Llama-\allowbreak{}3.1-\allowbreak{}8B-\allowbreak{}Instruct / DGX Spark}}\\\midrule\textbf{\boldmath{Weights}} & \textbf{\boldmath{Budget}} & \textbf{\boldmath{$k$}} & \textbf{\boldmath{$s$}} & \textbf{\boldmath{$n_{\mathrm{tb}}$}} \\\midrule}
\tablehead{\toprule\multicolumn{5}{@{}p{\linewidth}@{}}{\textbf{Llama-\allowbreak{}3.1-\allowbreak{}8B-\allowbreak{}Instruct / DGX Spark}}\\\midrule\textbf{\boldmath{Weights}} & \textbf{\boldmath{Budget}} & \textbf{\boldmath{$k$}} & \textbf{\boldmath{$s$}} & \textbf{\boldmath{$n_{\mathrm{tb}}$}} \\\midrule}
\tabletail{\bottomrule}
\tablelasttail{\bottomrule}
\appendixfirstgroupreserve{8}
\begin{supertabular*}{\linewidth}{@{}p{.20\linewidth}@{}p{.14\linewidth}@{}p{.25\linewidth}@{}p{.29\linewidth}@{}p{.12\linewidth}@{}}
\appendixgroupreserve{4}
\multirow[c]{4}{=}{AWQ 3} & {2.5\%} & {3,3,2} & {0,0,0} & {7,4} \\
 & {5\%} & {10,14,8} & {0,0,0} & {2,4} \\
 & {10\%} & {21,21,21,21} & {0,0,0,0} & {3} \\
 & {20\%} & {31,31,23} & {0,0,0} & {3,8} \\
\midrule
\appendixgroupreserve{4}
\multirow[c]{4}{=}{AWQ 4} & {2.5\%} & {5,6,0,5} & {0,0,0,0} & {3} \\
 & {5\%} & {16,16,14,16} & {0,0,0,0} & {4} \\
 & {10\%} & {28,28,0,28} & {0,0,0,0} & {3} \\
 & {20\%} & {34,33,33,33} & {0,0,0,0} & {6} \\
\midrule
\appendixgroupreserve{4}
\multirow[c]{4}{=}{LNQ+GQ 3} & {2.5\%} & {0,4,3} & {0,0,0} & {7,4} \\
 & {5\%} & {11,0,8} & {0,0,0} & {6,4} \\
 & {10\%} & {19,19,14} & {0,0,0} & {6,8} \\
 & {20\%} & {33,32,24} & {0,0,0} & {5,16} \\
\midrule
\appendixgroupreserve{4}
\multirow[c]{4}{=}{LNQ+GQ 4} & {2.5\%} & {8,8,0,8} & {0,0,0,0} & {7} \\
 & {5\%} & {14,0,9} & {0,0,0} & {5,4} \\
 & {10\%} & {23,24,16} & {0,0,0} & {7,8} \\
 & {20\%} & {40,40,29} & {0,0,0} & {8,8} \\
\midrule
\appendixgroupreserve{4}
\multirow[c]{4}{=}{SqLLM 3} & {2.5\%} & {9,8,6} & {0,0,0} & {6,4} \\
 & {5\%} & {10,14,9,9} & {0,0,0,0} & {3} \\
 & {10\%} & {18,18,19,18} & {0,0,0,0} & {4} \\
 & {20\%} & {28,29,28,27} & {0,0,0,0} & {3} \\
\midrule
\appendixgroupreserve{4}
\multirow[c]{4}{=}{SqLLM 4} & {2.5\%} & {0,15,15,15} & {0,0,0,0} & {4} \\
 & {5\%} & {10,9,10,10} & {0,0,0,0} & {6} \\
 & {10\%} & {21,21,21,21} & {0,0,0,0} & {5} \\
 & {20\%} & {33,33,33,33} & {0,0,0,0} & {6} \\
\midrule
\appendixgroupreserve{4}
\multirow[c]{4}{=}{YAQA 2} & {2.5\%} & {8,0,5} & {0,0,0} & {1,4} \\
 & {5\%} & {14,12,8} & {0,0,0} & {2,4} \\
 & {10\%} & {0,26,18} & {0,0,0} & {2,8} \\
 & {20\%} & {32,33,22} & {0,0,0} & {3,8} \\
\midrule
\appendixgroupreserve{4}
\multirow[c]{4}{=}{YAQA 3} & {2.5\%} & {5,5,4} & {0,0,0} & {1,8} \\
 & {5\%} & {16,16,12} & {0,0,0} & {2,4} \\
 & {10\%} & {28,28,21} & {0,0,0} & {3,8} \\
 & {20\%} & {35,35,26} & {0,0,0} & {3,4} \\
\midrule
\appendixgroupreserve{4}
\multirow[c]{4}{=}{YAQA 4} & {2.5\%} & {0,6,4} & {0,0,0} & {1,4} \\
 & {5\%} & {18,18,12} & {0,0,0} & {2,4} \\
 & {10\%} & {0,32,24} & {0,0,0} & {2,8} \\
 & {20\%} & {42,43,30} & {0,0,0} & {3,8} \cr
\end{supertabular*}
\par\vspace{5pt}
\tablefirsthead{\toprule\multicolumn{5}{@{}p{\linewidth}@{}}{\textbf{Llama-\allowbreak{}3.1-\allowbreak{}8B-\allowbreak{}Instruct / DGX Spark, skip-\allowbreak{}policy sweep}}\\\midrule\textbf{\boldmath{Weights}} & \textbf{\boldmath{Budget}} & \textbf{\boldmath{$k$}} & \textbf{\boldmath{$s$}} & \textbf{\boldmath{$n_{\mathrm{tb}}$}} \\\midrule}
\tablehead{\toprule\multicolumn{5}{@{}p{\linewidth}@{}}{\textbf{Llama-\allowbreak{}3.1-\allowbreak{}8B-\allowbreak{}Instruct / DGX Spark, skip-\allowbreak{}policy sweep}}\\\midrule\textbf{\boldmath{Weights}} & \textbf{\boldmath{Budget}} & \textbf{\boldmath{$k$}} & \textbf{\boldmath{$s$}} & \textbf{\boldmath{$n_{\mathrm{tb}}$}} \\\midrule}
\tabletail{\bottomrule}
\tablelasttail{\bottomrule}
\appendixfirstgroupreserve{8}
\begin{supertabular*}{\linewidth}{@{}p{.20\linewidth}@{}p{.14\linewidth}@{}p{.25\linewidth}@{}p{.29\linewidth}@{}p{.12\linewidth}@{}}
\appendixgroupreserve{6}
\multirow[c]{6}{=}{YAQA 2} & {2.5\%} & {6,12,4} & {0,0,0} & {3,4} \\
 & {5\%} & {0,17,12} & {0,0,0} & {5,4} \\
 & {10\%} & {24,23,18} & {0,0,0} & {2,8} \\
 & {20\%} & {0,37,28} & {0,0,0} & {4,8} \\
 & {30\%} & {0,40,30} & {0,0,0} & {13,4} \\
 & {40\%} & {0,46,32} & {0,0,0} & {13,4} \cr
\end{supertabular*}
\par\vspace{5pt}
\tablefirsthead{\toprule\multicolumn{5}{@{}p{\linewidth}@{}}{\textbf{Llama-\allowbreak{}3.1-\allowbreak{}8B-\allowbreak{}Instruct / DGX Spark, wait-\allowbreak{}policy sweep}}\\\midrule\textbf{\boldmath{Weights}} & \textbf{\boldmath{Budget}} & \textbf{\boldmath{$k$}} & \textbf{\boldmath{$s$}} & \textbf{\boldmath{$n_{\mathrm{tb}}$}} \\\midrule}
\tablehead{\toprule\multicolumn{5}{@{}p{\linewidth}@{}}{\textbf{Llama-\allowbreak{}3.1-\allowbreak{}8B-\allowbreak{}Instruct / DGX Spark, wait-\allowbreak{}policy sweep}}\\\midrule\textbf{\boldmath{Weights}} & \textbf{\boldmath{Budget}} & \textbf{\boldmath{$k$}} & \textbf{\boldmath{$s$}} & \textbf{\boldmath{$n_{\mathrm{tb}}$}} \\\midrule}
\tabletail{\bottomrule}
\tablelasttail{\bottomrule}
\appendixfirstgroupreserve{8}
\begin{supertabular*}{\linewidth}{@{}p{.20\linewidth}@{}p{.14\linewidth}@{}p{.25\linewidth}@{}p{.29\linewidth}@{}p{.12\linewidth}@{}}
\appendixgroupreserve{6}
\multirow[c]{6}{=}{YAQA 2} & {2.5\%} & {0,3,1} & {0,1.5,1.5} & {2,4} \\
 & {5\%} & {0,3,1} & {0,1.5,1.5} & {4,4} \\
 & {10\%} & {4,4,3} & {2.5,2,2} & {2,4} \\
 & {20\%} & {10,10,8} & {1.5,1.5,1.5} & {3,4} \\
 & {30\%} & {40,0,9} & {2,0,1.5} & {9,4} \\
 & {40\%} & {16,0,11} & {1.5,0,1.5} & {14,4} \cr
\end{supertabular*}
\par\vspace{5pt}
\tablefirsthead{\toprule\multicolumn{5}{@{}p{\linewidth}@{}}{\textbf{Llama-\allowbreak{}3.1-\allowbreak{}8B-\allowbreak{}Instruct / RTX 5070 Ti}}\\\midrule\textbf{\boldmath{Weights}} & \textbf{\boldmath{Budget}} & \textbf{\boldmath{$k$}} & \textbf{\boldmath{$s$}} & \textbf{\boldmath{$n_{\mathrm{tb}}$}} \\\midrule}
\tablehead{\toprule\multicolumn{5}{@{}p{\linewidth}@{}}{\textbf{Llama-\allowbreak{}3.1-\allowbreak{}8B-\allowbreak{}Instruct / RTX 5070 Ti}}\\\midrule\textbf{\boldmath{Weights}} & \textbf{\boldmath{Budget}} & \textbf{\boldmath{$k$}} & \textbf{\boldmath{$s$}} & \textbf{\boldmath{$n_{\mathrm{tb}}$}} \\\midrule}
\tabletail{\bottomrule}
\tablelasttail{\bottomrule}
\appendixfirstgroupreserve{8}
\begin{supertabular*}{\linewidth}{@{}p{.20\linewidth}@{}p{.14\linewidth}@{}p{.25\linewidth}@{}p{.29\linewidth}@{}p{.12\linewidth}@{}}
\appendixgroupreserve{4}
\multirow[c]{4}{=}{AWQ 3} & {2.5\%} & {3,10,2} & {0,0.5,0} & {3,4} \\
 & {5\%} & {18,0,18,18} & {1,0,0,0} & {4} \\
 & {10\%} & {37,37,37,37} & {1,0,0,0.5} & {6} \\
 & {20\%} & {45,40,39,40} & {0,0,0,0} & {7} \\
\midrule
\appendixgroupreserve{4}
\multirow[c]{4}{=}{AWQ 4} & {2.5\%} & {28,28,34,36} & {0.5,1,0,0.5} & {6} \\
 & {5\%} & {52,53,49,52} & {1,0.5,0,0.5} & {24} \\
 & {10\%} & {45,45,45,45} & {0,0.5,0,0} & {24} \\
 & {20\%} & {53,55,52,52} & {0,0,0,0} & {24} \\
\midrule
\appendixgroupreserve{4}
\multirow[c]{4}{=}{LNQ+GQ 3} & {2.5\%} & {6,8,3} & {0,1,0.5} & {4,4} \\
 & {5\%} & {63,63,33,61} & {1,0.5,0,0.5} & {16} \\
 & {10\%} & {29,28,20} & {0.5,0,0.5} & {16,16} \\
 & {20\%} & {60,61,58,60} & {1,0.5,0,0.5} & {16} \\
\midrule
\appendixgroupreserve{4}
\multirow[c]{4}{=}{LNQ+GQ 4} & {2.5\%} & {45,42,47,45} & {1,0.5,0,0} & {16} \\
 & {5\%} & {48,45,48,49} & {1,0,0,0} & {16} \\
 & {10\%} & {68,59,58,59} & {1,0,0,0.5} & {16} \\
 & {20\%} & {64,64,64,63} & {1,0,0,0} & {24} \\
\midrule
\appendixgroupreserve{4}
\multirow[c]{4}{=}{SqLLM 3} & {2.5\%} & {10,10,10,10} & {0.5,0.5,0,0} & {4} \\
 & {5\%} & {32,37,31,30} & {1,0.5,0,0} & {16} \\
 & {10\%} & {27,27,26,27} & {0,0,0,0} & {16} \\
 & {20\%} & {43,43,43,43} & {1,0,0,0} & {16} \\
\midrule
\appendixgroupreserve{4}
\multirow[c]{4}{=}{SqLLM 4} & {2.5\%} & {85,46,46,45} & {1,0.5,0,0} & {16} \\
 & {5\%} & {48,48,48,48} & {1,0,0,0} & {16} \\
 & {10\%} & {54,53,53,54} & {1,0,0,0} & {16} \\
 & {20\%} & {59,60,57,58} & {0,0,0,0} & {24} \\
\midrule
\appendixgroupreserve{4}
\multirow[c]{4}{=}{YAQA 2} & {2.5\%} & {15,23,12} & {1,1,0.5} & {6,4} \\
 & {5\%} & {28,28,16} & {1,1,0.5} & {6,4} \\
 & {10\%} & {37,39,25} & {1,0,0.5} & {7,16} \\
 & {20\%} & {97,49,35} & {0.5,0,0.5} & {14,32} \\
\midrule
\appendixgroupreserve{4}
\multirow[c]{4}{=}{YAQA 3} & {2.5\%} & {32,32,31,32} & {0.5,0.5,0,0} & {5} \\
 & {5\%} & {66,50,44,44} & {0,0.5,0,0.5} & {7} \\
 & {10\%} & {62,48,47,48} & {0,0,0,0} & {14} \\
 & {20\%} & {61,61,60,61} & {0,0,0,0} & {16} \\
\midrule
\appendixgroupreserve{4}
\multirow[c]{4}{=}{YAQA 4} & {2.5\%} & {59,60,58,58} & {1,0,0,0.5} & {7} \\
 & {5\%} & {82,60,58,60} & {0,1,0,0.5} & {7} \\
 & {10\%} & {73,77,70,70} & {1,0,0,0.5} & {23} \\
 & {20\%} & {87,87,86,87} & {0,0,0,0.5} & {28} \cr
\end{supertabular*}
\par\vspace{5pt}
\tablefirsthead{\toprule\multicolumn{5}{@{}p{\linewidth}@{}}{\textbf{Llama-\allowbreak{}3.1-\allowbreak{}8B-\allowbreak{}Instruct / RTX 5090}}\\\midrule\textbf{\boldmath{Weights}} & \textbf{\boldmath{Budget}} & \textbf{\boldmath{$k$}} & \textbf{\boldmath{$s$}} & \textbf{\boldmath{$n_{\mathrm{tb}}$}} \\\midrule}
\tablehead{\toprule\multicolumn{5}{@{}p{\linewidth}@{}}{\textbf{Llama-\allowbreak{}3.1-\allowbreak{}8B-\allowbreak{}Instruct / RTX 5090}}\\\midrule\textbf{\boldmath{Weights}} & \textbf{\boldmath{Budget}} & \textbf{\boldmath{$k$}} & \textbf{\boldmath{$s$}} & \textbf{\boldmath{$n_{\mathrm{tb}}$}} \\\midrule}
\tabletail{\bottomrule}
\tablelasttail{\bottomrule}
\appendixfirstgroupreserve{8}
\begin{supertabular*}{\linewidth}{@{}p{.20\linewidth}@{}p{.14\linewidth}@{}p{.25\linewidth}@{}p{.29\linewidth}@{}p{.12\linewidth}@{}}
\appendixgroupreserve{4}
\multirow[c]{4}{=}{AWQ 3} & {2.5\%} & {8,0,0,8} & {0.5,0,0,0} & {8} \\
 & {5\%} & {17,0,17,13} & {1,0,0,0} & {16} \\
 & {10\%} & {13,13,13,13} & {0,1,0,0} & {14} \\
 & {20\%} & {29,28,22,27} & {1,0,0,0} & {14} \\
\midrule
\appendixgroupreserve{4}
\multirow[c]{4}{=}{AWQ 4} & {2.5\%} & {10,0,10,10} & {0.5,0,0,0} & {16} \\
 & {5\%} & {20,0,18,13} & {0.5,0,0,0} & {16} \\
 & {10\%} & {30,39,26,28} & {0.5,0.5,0,0.5} & {24} \\
 & {20\%} & {53,54,46,46} & {0.5,1,1,0.5} & {24} \\
\midrule
\appendixgroupreserve{4}
\multirow[c]{4}{=}{LNQ+GQ 3} & {2.5\%} & {18,0,13,9} & {0.5,0,0,0} & {16} \\
 & {5\%} & {17,0,13,13} & {0.5,0,0,0} & {16} \\
 & {10\%} & {59,65,5} & {1,1,0.5} & {24,4} \\
 & {20\%} & {29,29,16} & {1,0,0.5} & {24,16} \\
\midrule
\appendixgroupreserve{4}
\multirow[c]{4}{=}{LNQ+GQ 4} & {2.5\%} & {25,0,15,11} & {0.5,0,0,0} & {11} \\
 & {5\%} & {21,0,21,18} & {0.5,0,0,0} & {14} \\
 & {10\%} & {23,25,21,22} & {1,1,0,0} & {16} \\
 & {20\%} & {40,32,31,32} & {1,0,0,0} & {16} \\
\midrule
\appendixgroupreserve{4}
\multirow[c]{4}{=}{SqLLM 3} & {2.5\%} & {17,0,9,9} & {0.5,0,0,0} & {16} \\
 & {5\%} & {17,0,13,13} & {0.5,0,0,0} & {16} \\
 & {10\%} & {10,10,10,10} & {0.5,0,0,0} & {16} \\
 & {20\%} & {26,26,20,22} & {1,0,0,0} & {16} \\
\midrule
\appendixgroupreserve{4}
\multirow[c]{4}{=}{SqLLM 4} & {2.5\%} & {16,0,16,9} & {1,0,0,0} & {11} \\
 & {5\%} & {22,0,21,21} & {0.5,0,0,0} & {24} \\
 & {10\%} & {19,19,19,19} & {0.5,0,0,0} & {24} \\
 & {20\%} & {29,29,28,29} & {0,0,0,0} & {38} \\
\midrule
\appendixgroupreserve{4}
\multirow[c]{4}{=}{YAQA 2} & {2.5\%} & {16,28,12} & {0,0,0.5} & {23,8} \\
 & {5\%} & {39,40,18} & {0,0,0.5} & {23,16} \\
 & {10\%} & {39,39,20} & {0,0,0.5} & {19,32} \\
 & {20\%} & {50,51,24} & {0.5,0,0.5} & {38,16} \\
\midrule
\appendixgroupreserve{4}
\multirow[c]{4}{=}{YAQA 3} & {2.5\%} & {29,28,26,26} & {1,0,0,0} & {56} \\
 & {5\%} & {45,46,20} & {0,0,0.5} & {56,16} \\
 & {10\%} & {51,51,24} & {0.5,0,0.5} & {38,32} \\
 & {20\%} & {53,53,49,49} & {0,0,0,0.5} & {38} \\
\midrule
\appendixgroupreserve{4}
\multirow[c]{4}{=}{YAQA 4} & {2.5\%} & {42,42,34,34} & {0,0.5,0,0} & {56} \\
 & {5\%} & {38,39,38,38} & {0,0,0,0.5} & {28} \\
 & {10\%} & {50,50,50,50} & {0.5,0.5,0.5,0.5} & {38} \\
 & {20\%} & {58,60,57,57} & {0,0,0,0.5} & {38} \cr
\end{supertabular*}
\par\vspace{5pt}
\tablefirsthead{\toprule\multicolumn{5}{@{}p{\linewidth}@{}}{\textbf{Qwen3-\allowbreak{}8B / DGX Spark}}\\\midrule\textbf{\boldmath{Weights}} & \textbf{\boldmath{Budget}} & \textbf{\boldmath{$k$}} & \textbf{\boldmath{$s$}} & \textbf{\boldmath{$n_{\mathrm{tb}}$}} \\\midrule}
\tablehead{\toprule\multicolumn{5}{@{}p{\linewidth}@{}}{\textbf{Qwen3-\allowbreak{}8B / DGX Spark}}\\\midrule\textbf{\boldmath{Weights}} & \textbf{\boldmath{Budget}} & \textbf{\boldmath{$k$}} & \textbf{\boldmath{$s$}} & \textbf{\boldmath{$n_{\mathrm{tb}}$}} \\\midrule}
\tabletail{\bottomrule}
\tablelasttail{\bottomrule}
\appendixfirstgroupreserve{8}
\begin{supertabular*}{\linewidth}{@{}p{.20\linewidth}@{}p{.14\linewidth}@{}p{.25\linewidth}@{}p{.29\linewidth}@{}p{.12\linewidth}@{}}
\appendixgroupreserve{4}
\multirow[c]{4}{=}{AWQ 3} & {2.5\%} & {0,8,2,2} & {0,0,0,0} & {3} \\
 & {5\%} & {0,16,16,16} & {0,0,0,0} & {5} \\
 & {10\%} & {25,21,20,21} & {0,0,0,0} & {3} \\
 & {20\%} & {28,27,28,27} & {0,0,0,0} & {5} \\
\midrule
\appendixgroupreserve{4}
\multirow[c]{4}{=}{AWQ 4} & {2.5\%} & {10,10,0,10} & {0,0,0,0} & {10} \\
 & {5\%} & {11,11,0,11} & {0,0,0,0} & {11} \\
 & {10\%} & {20,20,20,19} & {0,0,0,0} & {11} \\
 & {20\%} & {37,37,37,37} & {0,0,0,0} & {7} \\
\midrule
\appendixgroupreserve{4}
\multirow[c]{4}{=}{LNQ+GQ 3} & {2.5\%} & {4,0,3} & {0,0,0} & {7,4} \\
 & {5\%} & {13,10,7} & {0,0,0} & {3,4} \\
 & {10\%} & {0,24,18} & {0,0,0} & {3,8} \\
 & {20\%} & {34,34,26} & {0,0,0} & {5,8} \\
\midrule
\appendixgroupreserve{4}
\multirow[c]{4}{=}{LNQ+GQ 4} & {2.5\%} & {0,13,10} & {0,0,0} & {5,4} \\
 & {5\%} & {0,19,12} & {0,0,0} & {5,8} \\
 & {10\%} & {31,31,23} & {0,0,0} & {5,8} \\
 & {20\%} & {44,44,32} & {0,0,0} & {7,8} \\
\midrule
\appendixgroupreserve{4}
\multirow[c]{4}{=}{SqLLM 3} & {2.5\%} & {5,4,4,4} & {0,0,0,0} & {5} \\
 & {5\%} & {8,8,8,8} & {0,0,0,0} & {3} \\
 & {10\%} & {22,23,22,22} & {0,0,0,0} & {4} \\
 & {20\%} & {31,30,30,32} & {0,0,0,0} & {9} \\
\midrule
\appendixgroupreserve{4}
\multirow[c]{4}{=}{SqLLM 4} & {2.5\%} & {0,9,8,9} & {0,0,0,0} & {7} \\
 & {5\%} & {11,12,11,11} & {0,0,0,0} & {5} \\
 & {10\%} & {24,24,24,24} & {0,0,0,0} & {6} \\
 & {20\%} & {38,38,38,38} & {0,0,0,0} & {6} \\
\midrule
\appendixgroupreserve{4}
\multirow[c]{4}{=}{YAQA 2} & {2.5\%} & {5,7,4} & {0,0,0} & {2,4} \\
 & {5\%} & {11,13,8} & {0,0,0} & {5,4} \\
 & {10\%} & {27,29,19} & {0,0,0} & {2,8} \\
 & {20\%} & {32,33,24} & {0,0,0} & {3,8} \\
\midrule
\appendixgroupreserve{4}
\multirow[c]{4}{=}{YAQA 3} & {2.5\%} & {0,9,7} & {0,0,0} & {5,4} \\
 & {5\%} & {0,17,12} & {0,0,0} & {6,4} \\
 & {10\%} & {32,34,22} & {0,0,0} & {4,8} \\
 & {20\%} & {38,40,27} & {0,0,0} & {2,8} \\
\midrule
\appendixgroupreserve{4}
\multirow[c]{4}{=}{YAQA 4} & {2.5\%} & {0,21,15} & {0,0,0} & {7,4} \\
 & {5\%} & {14,14,11} & {0,0,0} & {3,4} \\
 & {10\%} & {30,29,21} & {0,0,0} & {3,8} \\
 & {20\%} & {42,42,32} & {0,0,0} & {2,8} \cr
\end{supertabular*}
\par\vspace{5pt}
\tablefirsthead{\toprule\multicolumn{5}{@{}p{\linewidth}@{}}{\textbf{Qwen3-\allowbreak{}8B / RTX 5070 Ti}}\\\midrule\textbf{\boldmath{Weights}} & \textbf{\boldmath{Budget}} & \textbf{\boldmath{$k$}} & \textbf{\boldmath{$s$}} & \textbf{\boldmath{$n_{\mathrm{tb}}$}} \\\midrule}
\tablehead{\toprule\multicolumn{5}{@{}p{\linewidth}@{}}{\textbf{Qwen3-\allowbreak{}8B / RTX 5070 Ti}}\\\midrule\textbf{\boldmath{Weights}} & \textbf{\boldmath{Budget}} & \textbf{\boldmath{$k$}} & \textbf{\boldmath{$s$}} & \textbf{\boldmath{$n_{\mathrm{tb}}$}} \\\midrule}
\tabletail{\bottomrule}
\tablelasttail{\bottomrule}
\appendixfirstgroupreserve{8}
\begin{supertabular*}{\linewidth}{@{}p{.20\linewidth}@{}p{.14\linewidth}@{}p{.25\linewidth}@{}p{.29\linewidth}@{}p{.12\linewidth}@{}}
\appendixgroupreserve{4}
\multirow[c]{4}{=}{AWQ 3} & {2.5\%} & {0,6,6,6} & {0,0.5,0,0.5} & {3} \\
 & {5\%} & {22,21,14} & {1,1,0.5} & {4,4} \\
 & {10\%} & {44,44,44,44} & {1,1,0,0.5} & {6} \\
 & {20\%} & {54,54,53,55} & {1,0,0,0.5} & {6} \\
\midrule
\appendixgroupreserve{4}
\multirow[c]{4}{=}{AWQ 4} & {2.5\%} & {19,19,19,19} & {0.5,1,0,0} & {6} \\
 & {5\%} & {35,41,35,41} & {0.5,1,0,0.5} & {8} \\
 & {10\%} & {61,61,60,61} & {1,0.5,0,0.5} & {24} \\
 & {20\%} & {63,63,62,62} & {0,0,1,0} & {24} \\
\midrule
\appendixgroupreserve{4}
\multirow[c]{4}{=}{LNQ+GQ 3} & {2.5\%} & {0,16,14} & {0,1,0.5} & {5,16} \\
 & {5\%} & {15,17,7} & {0.5,1,0} & {5,32} \\
 & {10\%} & {32,39,22} & {1,0.5,0.5} & {16,16} \\
 & {20\%} & {44,44,32} & {0.5,0,0.5} & {16,32} \\
\midrule
\appendixgroupreserve{4}
\multirow[c]{4}{=}{LNQ+GQ 4} & {2.5\%} & {27,28,21} & {0.5,1,0.5} & {11,16} \\
 & {5\%} & {46,46,24} & {0.5,0,0.5} & {16,16} \\
 & {10\%} & {44,45,29} & {0.5,0,0.5} & {16,16} \\
 & {20\%} & {54,57,40} & {0,0,0.5} & {16,32} \\
\midrule
\appendixgroupreserve{4}
\multirow[c]{4}{=}{SqLLM 3} & {2.5\%} & {11,11,8} & {0.5,1,0.5} & {5,16} \\
 & {5\%} & {36,40,24,37} & {0.5,0.5,0,0.5} & {5} \\
 & {10\%} & {42,42,42,42} & {1,0.5,1,0} & {16} \\
 & {20\%} & {51,51,50,51} & {1,0,1,0} & {16} \\
\midrule
\appendixgroupreserve{4}
\multirow[c]{4}{=}{SqLLM 4} & {2.5\%} & {38,0,37,37} & {1,0,0,0} & {16} \\
 & {5\%} & {45,42,45,47} & {1,0.5,0,0} & {16} \\
 & {10\%} & {44,45,44,45} & {0,0,0,0} & {16} \\
 & {20\%} & {62,62,61,61} & {1,0,0,0} & {16} \\
\midrule
\appendixgroupreserve{4}
\multirow[c]{4}{=}{YAQA 2} & {2.5\%} & {21,28,14} & {1,1,0.5} & {6,4} \\
 & {5\%} & {27,28,18} & {1,0,0.5} & {6,4} \\
 & {10\%} & {67,74,29} & {0.5,0.5,0.5} & {6,32} \\
 & {20\%} & {52,52,36} & {0,0,0.5} & {6,32} \\
\midrule
\appendixgroupreserve{4}
\multirow[c]{4}{=}{YAQA 3} & {2.5\%} & {26,31,17} & {0.5,1,0.5} & {5,4} \\
 & {5\%} & {32,37,24} & {0.5,1,0.5} & {6,16} \\
 & {10\%} & {72,50,34} & {1,0,0.5} & {6,32} \\
 & {20\%} & {56,57,42} & {0,0,0.5} & {16,32} \\
\midrule
\appendixgroupreserve{4}
\multirow[c]{4}{=}{YAQA 4} & {2.5\%} & {38,39,22} & {1,0,0.5} & {7,4} \\
 & {5\%} & {49,52,35} & {1,1,0.5} & {6,32} \\
 & {10\%} & {57,59,40} & {0.5,0,0.5} & {6,32} \\
 & {20\%} & {67,67,48} & {0,0,0.5} & {16,32} \cr
\end{supertabular*}
\par\vspace{5pt}
\tablefirsthead{\toprule\multicolumn{5}{@{}p{\linewidth}@{}}{\textbf{Qwen3-\allowbreak{}8B / RTX 5090}}\\\midrule\textbf{\boldmath{Weights}} & \textbf{\boldmath{Budget}} & \textbf{\boldmath{$k$}} & \textbf{\boldmath{$s$}} & \textbf{\boldmath{$n_{\mathrm{tb}}$}} \\\midrule}
\tablehead{\toprule\multicolumn{5}{@{}p{\linewidth}@{}}{\textbf{Qwen3-\allowbreak{}8B / RTX 5090}}\\\midrule\textbf{\boldmath{Weights}} & \textbf{\boldmath{Budget}} & \textbf{\boldmath{$k$}} & \textbf{\boldmath{$s$}} & \textbf{\boldmath{$n_{\mathrm{tb}}$}} \\\midrule}
\tabletail{\bottomrule}
\tablelasttail{\bottomrule}
\appendixfirstgroupreserve{8}
\begin{supertabular*}{\linewidth}{@{}p{.20\linewidth}@{}p{.14\linewidth}@{}p{.25\linewidth}@{}p{.29\linewidth}@{}p{.12\linewidth}@{}}
\appendixgroupreserve{4}
\multirow[c]{4}{=}{AWQ 3} & {2.5\%} & {2,0,2} & {0,0,0.5} & {12,32} \\
 & {5\%} & {17,0,16,10} & {1,0,0,0} & {16} \\
 & {10\%} & {60,60,32,32} & {1,0.5,1,0.5} & {12} \\
 & {20\%} & {48,48,44,48} & {1,0.5,1,0.5} & {16} \\
\midrule
\appendixgroupreserve{4}
\multirow[c]{4}{=}{AWQ 4} & {2.5\%} & {14,0,13,5} & {0.5,0,0,0} & {16} \\
 & {5\%} & {18,0,16,10} & {0.5,0,0,0} & {16} \\
 & {10\%} & {29,29,26,29} & {0.5,0.5,0,0.5} & {32} \\
 & {20\%} & {51,112,50,51} & {1,0.5,1,0.5} & {32} \\
\midrule
\appendixgroupreserve{4}
\multirow[c]{4}{=}{LNQ+GQ 3} & {2.5\%} & {0,16,12,5} & {0,0.5,0,0} & {32} \\
 & {5\%} & {0,11,11,11} & {0,0.5,0,0} & {32} \\
 & {10\%} & {13,13,10} & {0.5,0.5,0.5} & {32,16} \\
 & {20\%} & {27,31,20} & {0.5,0,0.5} & {32,16} \\
\midrule
\appendixgroupreserve{4}
\multirow[c]{4}{=}{LNQ+GQ 4} & {2.5\%} & {16,0,14,14} & {1,0,1,0} & {48} \\
 & {5\%} & {31,0,26,25} & {1,0,1,0} & {32} \\
 & {10\%} & {54,54,9} & {1,1,0.5} & {48,4} \\
 & {20\%} & {31,31,23} & {0,0,0.5} & {16,16} \\
\midrule
\appendixgroupreserve{4}
\multirow[c]{4}{=}{SqLLM 3} & {2.5\%} & {10,0,9,10} & {1,0,1,0} & {24} \\
 & {5\%} & {11,0,11,11} & {1,0,0,0} & {32} \\
 & {10\%} & {25,27,25,26} & {1,0.5,1,0.5} & {24} \\
 & {20\%} & {28,28,27,28} & {1,0,1,0} & {32} \\
\midrule
\appendixgroupreserve{4}
\multirow[c]{4}{=}{SqLLM 4} & {2.5\%} & {16,0,16,16} & {0.5,0,0,0} & {48} \\
 & {5\%} & {34,0,23,21} & {1,0,0,0} & {32} \\
 & {10\%} & {25,25,25,25} & {1,0.5,0,0.5} & {48} \\
 & {20\%} & {37,37,37,37} & {1,0,1,0} & {16} \\
\midrule
\appendixgroupreserve{4}
\multirow[c]{4}{=}{YAQA 2} & {2.5\%} & {11,10,8} & {0,0,0.5} & {7,16} \\
 & {5\%} & {22,20,16} & {0,0,0.5} & {7,16} \\
 & {10\%} & {40,41,22} & {0,0,0.5} & {20,32} \\
 & {20\%} & {47,46,26} & {0,0,0.5} & {32,32} \\
\midrule
\appendixgroupreserve{4}
\multirow[c]{4}{=}{YAQA 3} & {2.5\%} & {17,17,13} & {0,0,0.5} & {32,8} \\
 & {5\%} & {36,36,23} & {0,0,0.5} & {24,32} \\
 & {10\%} & {57,57,25} & {0.5,0,0.5} & {32,32} \\
 & {20\%} & {44,44,33} & {0.5,0,0.5} & {32,32} \\
\midrule
\appendixgroupreserve{4}
\multirow[c]{4}{=}{YAQA 4} & {2.5\%} & {19,18,14} & {0,0,0.5} & {5,4} \\
 & {5\%} & {42,42,26} & {0,0,0.5} & {8,16} \\
 & {10\%} & {44,45,28} & {0,0,0.5} & {32,32} \\
 & {20\%} & {47,48,36} & {0.5,0,0.5} & {32,16} \cr
\end{supertabular*}
\par\vspace{5pt}
\tablefirsthead{\toprule\multicolumn{4}{@{}p{\linewidth}@{}}{\textbf{Llama-\allowbreak{}3.3-\allowbreak{}70B-\allowbreak{}Instruct / RTX PRO 6000 / 5\% slowdown budget}}\\\midrule\textbf{\boldmath{Weights}} & \textbf{\boldmath{$k$}} & \textbf{\boldmath{$s$}} & \textbf{\boldmath{$n_{\mathrm{tb}}$}} \\\midrule}
\tablehead{\toprule\multicolumn{4}{@{}p{\linewidth}@{}}{\textbf{Llama-\allowbreak{}3.3-\allowbreak{}70B-\allowbreak{}Instruct / RTX PRO 6000 / 5\% slowdown budget}}\\\midrule\textbf{\boldmath{Weights}} & \textbf{\boldmath{$k$}} & \textbf{\boldmath{$s$}} & \textbf{\boldmath{$n_{\mathrm{tb}}$}} \\\midrule}
\tabletail{\bottomrule}
\tablelasttail{\bottomrule}
\appendixfirstgroupreserve{1}
\begin{supertabular*}{\linewidth}{@{}p{.25\linewidth}@{}p{.30\linewidth}@{}p{.30\linewidth}@{}p{.15\linewidth}@{}}
\appendixgroupreserve{1}
\multirow[c]{1}{=}{AWQ 3} & {25,25,25,25} & {0,0.5,0,0} & {8} \\
\appendixgroupreserve{1}
\multirow[c]{1}{=}{AWQ 4} & {30,30,31,29} & {1,0,0,0} & {40} \\
\appendixgroupreserve{1}
\multirow[c]{1}{=}{LNQ+GQ 3} & {13,13,10} & {0.5,0.5,0.5} & {17,32} \\
\appendixgroupreserve{1}
\multirow[c]{1}{=}{LNQ+GQ 4} & {24,24,24,24} & {0,0.5,0,0} & {28} \\
\appendixgroupreserve{1}
\multirow[c]{1}{=}{SqLLM 3} & {16,16,15,15} & {0,0,0,0} & {16} \\
\appendixgroupreserve{1}
\multirow[c]{1}{=}{SqLLM 4} & {24,24,24,24} & {1,0,0,0} & {28} \\
\appendixgroupreserve{1}
\multirow[c]{1}{=}{YAQA 2} & {32,32,24} & {0.5,0.5,1} & {38,8} \\
\appendixgroupreserve{1}
\multirow[c]{1}{=}{YAQA 3} & {42,37,33,34} & {0,0,0,0.5} & {32} \\
\appendixgroupreserve{1}
\multirow[c]{1}{=}{YAQA 4} & {42,45,38,42} & {0,0,0,0.5} & {32} \cr
\end{supertabular*}
\par\vspace{5pt}
\tablefirsthead{\toprule\multicolumn{4}{@{}p{\linewidth}@{}}{\textbf{Qwen3-\allowbreak{}32B / RTX 5090 / 5\% slowdown budget}}\\\midrule\textbf{\boldmath{Weights}} & \textbf{\boldmath{$k$}} & \textbf{\boldmath{$s$}} & \textbf{\boldmath{$n_{\mathrm{tb}}$}} \\\midrule}
\tablehead{\toprule\multicolumn{4}{@{}p{\linewidth}@{}}{\textbf{Qwen3-\allowbreak{}32B / RTX 5090 / 5\% slowdown budget}}\\\midrule\textbf{\boldmath{Weights}} & \textbf{\boldmath{$k$}} & \textbf{\boldmath{$s$}} & \textbf{\boldmath{$n_{\mathrm{tb}}$}} \\\midrule}
\tabletail{\bottomrule}
\tablelasttail{\bottomrule}
\appendixfirstgroupreserve{1}
\begin{supertabular*}{\linewidth}{@{}p{.25\linewidth}@{}p{.30\linewidth}@{}p{.30\linewidth}@{}p{.15\linewidth}@{}}
\appendixgroupreserve{1}
\multirow[c]{1}{=}{AWQ 3} & {18,20,18,18} & {0,0.5,0,0} & {11} \\
\appendixgroupreserve{1}
\multirow[c]{1}{=}{AWQ 4} & {19,19,19,17} & {0,0.5,0,0} & {12} \\
\appendixgroupreserve{1}
\multirow[c]{1}{=}{LNQ+GQ 3} & {16,18,17,18} & {0,0.5,0,0} & {14} \\
\appendixgroupreserve{1}
\multirow[c]{1}{=}{LNQ+GQ 4} & {28,26,28,26} & {1,0,0,0} & {16} \\
\appendixgroupreserve{1}
\multirow[c]{1}{=}{SqLLM 3} & {16,18,16,16} & {0,0.5,0,0} & {14} \\
\appendixgroupreserve{1}
\multirow[c]{1}{=}{SqLLM 4} & {27,27,26,27} & {1,0,0,0} & {16} \\
\appendixgroupreserve{1}
\multirow[c]{1}{=}{YAQA 2} & {41,41,15} & {0.5,0,0.5} & {50,20} \\
\appendixgroupreserve{1}
\multirow[c]{1}{=}{YAQA 3} & {201,32,19} & {0.5,0,0.5} & {50,20} \\
\appendixgroupreserve{1}
\multirow[c]{1}{=}{YAQA 4} & {27,28,20} & {0.5,0,0.5} & {25,20} \cr
\end{supertabular*}
\par\vspace{5pt}
}

\section{Extended KV Evaluation}
\label{sec:appendix-kv}
We extend Section~\ref{main-sec:eval-kv} across KV precisions and context lengths. In configuration labels, K and V give resident bits per key and value element; +4/+4 gives their additional refinement bits. The fetch budget is the fraction of eligible token groups selected. Main host-RAM configurations use two-token groups. Main SSD-backed configurations use four-token groups and skip refinement on RAM-cache misses.
\begin{figure}[!t]
\centering
\includegraphics[width=\columnwidth]{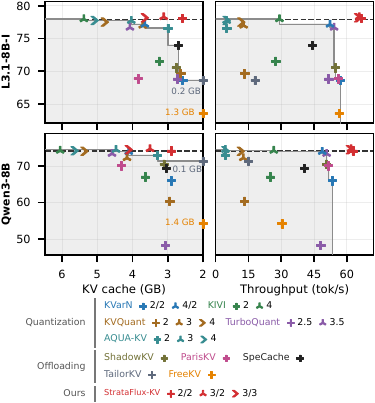}
\caption{KV-only refinement: RULER accuracy at 128K (13 tasks, 96 samples each) versus GPU-resident KV-cache size and RTX 5090 throughput at the same context. \methodname{}-KV uses its dedicated-GPU configuration. Gray shading extends below the baseline frontiers; dashed lines mark BF16 accuracy. TailorKV\textquotesingle{}s size is labeled at the KV-cache axis\textquotesingle{}s right edge.}
\label{fig:kv-downstream}
\end{figure}

\subsection{Accuracy and Context Length}
Figure~\ref{fig:kv-downstream} compares KV-cache size and decode throughput against RULER accuracy at 128K.
\begin{appendixflowtable}
\centering
\caption{KV-only refinement: GPU-resident KV-cache size and PG-19 PPL/KL at 128K, with RULER accuracy at 8K--128K. $^{*}$GB counts only KV memory. Baseline sizes are for RTX 5090; \methodname{}-KV headings identify the platform. DGX Spark sizes include its RAM cache. KVQ and TQ denote KVQuant and TurboQuant.}
\label{tab:kv-accuracy-ruler}
\begingroup\fontsize{7}{8.5}\selectfont\setlength{\tabcolsep}{.6pt}\renewcommand{\arraystretch}{1.0}\setlength{\aboverulesep}{.2ex}\setlength{\belowrulesep}{.25ex}
\tablehead{\toprule
\multicolumn{9}{@{}l}{\textbf{Llama-3.1-8B-Instruct}}\\
\cmidrule(lr){1-9}
\textbf{Method} & \textbf{GB\rlap{$^{*}$}} & \textbf{PPL} & \textbf{KL} & \textbf{R8} & \textbf{R16} & \textbf{R32} & \textbf{R64} & \textbf{R128}\\\midrule}
\tabletail{\bottomrule}\tablelasttail{\bottomrule}
\begin{supertabular*}{\linewidth}{@{\extracolsep{\fill}}@{}p{.28\linewidth}cccccccc@{}}
\multicolumn{1}{@{}p{.28\linewidth}}{BF16} & 17.2 & 10.32 & 0.000 & 95.1 & 92.9 & 87.1 & 85.8 & 77.9\\
\multicolumn{1}{@{}p{.28\linewidth}}{KVarN 2/2} & 2.6 & 10.98 & 0.067 & 94.4 & 90.4 & 85.5 & 82.5 & 68.6\\
\multicolumn{1}{@{}p{.28\linewidth}}{KVQ 2-bit} & 2.6 & 11.71 & 0.125 & 93.4 & 88.8 & 84.8 & 82.3 & 69.6\\
\multicolumn{1}{@{}p{.28\linewidth}}{TQ 2.5-bit} & 2.7 & 10.98 & 0.055 & 93.6 & 90.8 & 84.4 & 82.9 & 68.8\\
\multicolumn{1}{@{}p{.28\linewidth}}{KIVI-2} & 3.2 & 10.90 & 0.059 & 93.8 & 91.7 & 85.6 & 83.2 & 71.6\\
\multicolumn{1}{@{}p{.28\linewidth}}{KVarN 4/2} & 3.7 & 10.38 & 0.008 & 94.8 & 92.8 & 87.1 & 85.9 & 77.2\\
\multicolumn{1}{@{}p{.28\linewidth}}{KVQ 3-bit} & 3.7 & 10.56 & 0.024 & 94.8 & 92.2 & 86.9 & 85.3 & 77.0\\
\multicolumn{1}{@{}p{.28\linewidth}}{TQ 3.5-bit} & 4.1 & 10.43 & 0.010 & 94.9 & 92.6 & 87.1 & 85.3 & 76.7\\
\multicolumn{1}{@{}p{.28\linewidth}}{KVQ 4-bit} & 4.8 & 10.38 & 0.007 & 95.1 & 93.0 & 87.1 & 85.7 & 77.5\\
\multicolumn{1}{@{}p{.28\linewidth}}{KIVI-4} & 5.4 & 10.33 & 0.002 & 95.3 & 93.1 & 87.0 & 85.8 & 78.0\\
\multicolumn{1}{@{}p{.28\linewidth}}{TailorKV-2} & 0.2 & 11.21 & 0.107 & 94.1 & 92.5 & 88.4 & 81.7 & 68.6\\
\multicolumn{1}{@{}p{.28\linewidth}}{FreeKV} & 1.3 & 11.62 & 0.126 & 93.7 & 87.8 & 80.8 & 75.4 & 63.6\\
\multicolumn{1}{@{}p{.28\linewidth}}{SpeCache} & 2.7 & 10.68 & 0.035 & 95.0 & 92.6 & 87.1 & 85.5 & 74.0\\
\multicolumn{1}{@{}p{.28\linewidth}}{ShadowKV} & 2.8 & 11.24 & 0.086 & 90.5 & 88.4 & 84.4 & 81.0 & 70.6\\
\multicolumn{1}{@{}p{.28\linewidth}}{ParisKV} & 3.8 & 13.07 & 0.282 & 93.1 & 90.2 & 85.6 & 78.5 & 68.9\\
\midrule
\appendixgroupreserve{5}
\multicolumn{9}{@{}l}{\textbf{\methodname{}-KV / DGX Spark}}\\
\cmidrule(lr){1-9}
\appendixgroupreserve{3}
2/2 & 2.7 & 10.47 & 0.011 & 95.2 & 92.7 & 87.5 & 85.1 & 75.2\\
3/2 & 3.2 & 10.34 & 0.002 & 95.0 & 93.2 & 87.1 & 85.8 & 77.6\\
3/3 & 3.8 & 10.33 & 0.001 & 95.2 & 93.2 & 87.0 & 85.8 & 77.5\\
\midrule
\appendixgroupreserve{5}
\multicolumn{9}{@{}l}{\textbf{\methodname{}-KV / RTX 5090}}\\
\cmidrule(lr){1-9}
2/2 & 2.6 & 10.40 & 0.003 & 95.1 & 92.9 & 87.1 & 86.0 & 78.1\\
3/2 & 3.1 & 10.33 & 0.001 & 95.1 & 93.1 & 87.0 & 85.8 & 78.3\\
3/3 & 3.7 & 10.33 & 0.001 & 95.0 & 93.0 & 86.9 & 85.8 & 78.2\cr
\end{supertabular*}\endgroup\par\vspace{3pt}
\begingroup\fontsize{7}{8.5}\selectfont\setlength{\tabcolsep}{.6pt}\renewcommand{\arraystretch}{1.0}\setlength{\aboverulesep}{.2ex}\setlength{\belowrulesep}{.25ex}
\tablehead{\toprule
\multicolumn{9}{@{}l}{\textbf{Qwen3-8B}}\\
\cmidrule(lr){1-9}
\textbf{Method} & \textbf{GB\rlap{$^{*}$}} & \textbf{PPL} & \textbf{KL} & \textbf{R8} & \textbf{R16} & \textbf{R32} & \textbf{R64} & \textbf{R128}\\\midrule}
\tabletail{\bottomrule}\tablelasttail{\bottomrule}
\begin{supertabular*}{\linewidth}{@{\extracolsep{\fill}}@{}p{.28\linewidth}cccccccc@{}}
\multicolumn{1}{@{}p{.28\linewidth}}{BF16} & 19.3 & 12.89 & 0.000 & 93.8 & 93.8 & 92.1 & 80.6 & 74.1\\
\multicolumn{1}{@{}p{.28\linewidth}}{KVarN 2/2} & 2.9 & 14.40 & 0.130 & 93.0 & 91.9 & 89.0 & 72.6 & 66.0\\
\multicolumn{1}{@{}p{.28\linewidth}}{KVQ 2-bit} & 2.9 & 14.40 & 0.167 & 91.2 & 90.2 & 85.1 & 67.7 & 60.4\\
\multicolumn{1}{@{}p{.28\linewidth}}{TQ 2.5-bit} & 3.1 & 36.02 & 1.115 & 90.2 & 86.8 & 81.3 & 58.8 & 48.2\\
\multicolumn{1}{@{}p{.28\linewidth}}{KIVI-2} & 3.6 & 13.85 & 0.106 & 92.9 & 92.1 & 88.8 & 73.0 & 67.0\\
\multicolumn{1}{@{}p{.28\linewidth}}{KVarN 4/2} & 4.1 & 13.04 & 0.014 & 93.8 & 93.6 & 91.9 & 79.7 & 73.9\\
\multicolumn{1}{@{}p{.28\linewidth}}{KVQ 3-bit} & 4.2 & 13.21 & 0.041 & 93.6 & 93.1 & 91.9 & 79.1 & 72.2\\
\multicolumn{1}{@{}p{.28\linewidth}}{TQ 3.5-bit} & 4.6 & 13.99 & 0.102 & 94.0 & 93.1 & 92.0 & 78.6 & 73.4\\
\multicolumn{1}{@{}p{.28\linewidth}}{KVQ 4-bit} & 5.4 & 12.96 & 0.011 & 93.8 & 93.6 & 91.9 & 79.9 & 74.0\\
\multicolumn{1}{@{}p{.28\linewidth}}{KIVI-4} & 6.1 & 12.90 & 0.004 & 94.0 & 93.5 & 92.1 & 80.8 & 74.2\\
\multicolumn{1}{@{}p{.28\linewidth}}{TailorKV-2} & 0.1 & 13.66 & 0.098 & 93.7 & 93.1 & 90.5 & 78.1 & 71.4\\
\multicolumn{1}{@{}p{.28\linewidth}}{FreeKV} & 1.4 & 14.68 & 0.184 & 91.1 & 86.4 & 82.2 & 68.7 & 54.2\\
\multicolumn{1}{@{}p{.28\linewidth}}{SpeCache} & 3.0 & 13.96 & 0.106 & 93.8 & 93.6 & 91.5 & 77.0 & 69.3\\
\multicolumn{1}{@{}p{.28\linewidth}}{ShadowKV} & 3.1 & 15.10 & 0.187 & 89.4 & 87.3 & 82.8 & 74.7 & 70.5\\
\multicolumn{1}{@{}p{.28\linewidth}}{ParisKV} & 4.3 & 14.11 & 0.154 & 93.3 & 91.6 & 88.5 & 77.4 & 70.1\\
\midrule
\appendixgroupreserve{5}
\multicolumn{9}{@{}l}{\textbf{\methodname{}-KV / DGX Spark}}\\
\cmidrule(lr){1-9}
\appendixgroupreserve{3}
2/2 & 3.0 & 13.32 & 0.030 & 94.0 & 93.4 & 91.6 & 77.2 & 71.4\\
3/2 & 3.6 & 12.98 & 0.005 & 93.7 & 93.7 & 92.0 & 79.5 & 74.0\\
3/3 & 4.2 & 12.93 & 0.004 & 93.8 & 93.8 & 92.2 & 79.8 & 74.2\\
\midrule
\appendixgroupreserve{5}
\multicolumn{9}{@{}l}{\textbf{\methodname{}-KV / RTX 5090}}\\
\cmidrule(lr){1-9}
2/2 & 2.9 & 12.98 & 0.007 & 94.0 & 93.8 & 91.8 & 79.9 & 74.2\\
3/2 & 3.5 & 12.92 & 0.002 & 93.9 & 93.8 & 92.0 & 80.4 & 74.7\\
3/3 & 4.1 & 12.90 & 0.001 & 93.9 & 93.8 & 92.1 & 80.3 & 74.5\cr
\end{supertabular*}\endgroup\par\vspace{3pt}
\end{appendixflowtable}

Tables~\ref{tab:kv-accuracy-ruler} and~\ref{tab:kv_quantization_accuracy-1} compare 128K accuracy, memory, and throughput; Tables~\ref{tab:kv_downstream-1} and~\ref{tab:kv_downstream-2} report task accuracy. Tables~\ref{tab:kv_long_context-3} and~\ref{tab:kv_long_context-4} give context-length sweeps.
\appendixfirstgroupreserve{8}
\begin{appendixflowtable}
\centering
\caption{KV-only refinement: PG-19 PPL/KL and RTX 5090 throughput at 128K. \methodname{}-KV fetches +4/+4 bits from host RAM for 1/128 of two-token groups.}
\label{tab:kv_quantization_accuracy-1}
\begingroup\fontsize{7}{8.5}\selectfont\setlength{\tabcolsep}{.65pt}\renewcommand{\arraystretch}{1.0}\setlength{\aboverulesep}{.2ex}\setlength{\belowrulesep}{.25ex}
\tablehead{\toprule
 & \multicolumn{3}{c}{\textbf{L3.1-8B-I}} & \multicolumn{3}{c}{\textbf{Qwen3-8B}}\\
\cmidrule(lr){2-4}\cmidrule(lr){5-7}
\textbf{KV configuration} & \textbf{PPL} & \textbf{KL} & \textbf{tok/s} & \textbf{PPL} & \textbf{KL} & \textbf{tok/s}\\\midrule}
\tabletail{\bottomrule}\tablelasttail{\bottomrule}
\begin{supertabular*}{\linewidth}{@{\extracolsep{\fill}}@{}p{.39\linewidth}cccccc@{}}
BF16 KV & 10.32 & 0.000 & 47.6 & 12.89 & 0.000 & OOM\\
KVarN 2/2 & 10.98 & 0.067 & 57.1 & 14.40 & 0.130 & 53.4\\
KIVI-2 & 10.90 & 0.059 & 27.4 & 13.85 & 0.106 & 25.0\\
KVQuant 2-bit & 11.71 & 0.125 & 13.3 & 14.40 & 0.167 & 13.2\\
TurboQuant 2.5-bit & 10.98 & 0.055 & 51.7 & 36.02 & 1.115 & 48.1\\
AQUA-KV 2-bit & 10.66 & 0.030 & 5.1 & 13.14 & 0.029 & 4.5\\
KVQuant 3-bit & 10.56 & 0.024 & 12.8 & 13.21 & 0.041 & 12.7\\
TurboQuant 3.5-bit & 10.43 & 0.010 & 54.2 & 13.99 & 0.102 & 50.7\\
AQUA-KV 3-bit & 10.39 & 0.006 & 5.0 & 12.98 & 0.017 & 4.5\\
KVarN 4/2 & 10.38 & 0.008 & 52.3 & 13.04 & 0.014 & 48.8\\
KIVI-4 & 10.33 & 0.002 & 29.2 & 12.90 & 0.004 & 26.6\\
KVQuant 4-bit & 10.38 & 0.007 & 11.8 & 12.96 & 0.011 & 11.5\\
AQUA-KV 4-bit & 10.32 & 0.002 & 5.0 & 12.95 & 0.016 & 4.5\\
ShadowKV & 11.24 & 0.086 & 54.8 & 15.10 & 0.187 & 50.7\\
ParisKV & 13.07 & 0.282 & 56.0 & 14.11 & 0.154 & 51.6\\
TailorKV-2 & 11.21 & 0.107 & 18.0 & 13.66 & 0.098 & 15.0\\
FreeKV & 11.62 & 0.126 & 56.4 & 14.68 & 0.184 & 30.3\\
SpeCache & 10.68 & 0.035 & 44.3 & 13.96 & 0.106 & 40.5\\
\methodname{}-KV 2/2 & 10.40 & 0.003 & 66.6 & 12.98 & 0.007 & 62.7\\
\methodname{}-KV 3/2 & 10.33 & 0.001 & 65.7 & 12.92 & 0.002 & 61.9\\
\methodname{}-KV 3/3 & 10.33 & 0.001 & 65.1 & 12.90 & 0.001 & 61.4\cr
\end{supertabular*}\endgroup
\end{appendixflowtable}

\begin{appendixflowtable}
\centering
\caption{KV-only refinement: quantized KV task accuracy. R4--R128 give RULER accuracy at 4K--128K; LB32 uses LongBench prompts truncated to 32,128 tokens; M500/AIME report MATH500/AIME accuracy. Platform headings distinguish \methodname{}-KV configurations; baseline accuracies are shared across devices.}
\label{tab:kv_downstream-1}
\begingroup\fontsize{7}{8.5}\selectfont\setlength{\tabcolsep}{.65pt}\renewcommand{\arraystretch}{1.0}\setlength{\aboverulesep}{.2ex}\setlength{\belowrulesep}{.25ex}
\tablehead{\toprule
\multicolumn{8}{@{}l}{\textbf{Llama-3.1-8B-Instruct}}\\
\cmidrule(lr){1-8}
\textbf{Configuration} & \textbf{R4} & \textbf{R8} & \textbf{R16} & \textbf{R32} & \textbf{R64} & \textbf{R128} & \textbf{LB32}\\\midrule}
\tabletail{\bottomrule}\tablelasttail{\bottomrule}
\begin{supertabular*}{\linewidth}{@{\extracolsep{\fill}}@{}p{.35\linewidth}ccccccc@{}}
\multicolumn{1}{@{}p{.35\linewidth}}{BF16 KV} & 95.7 & 95.1 & 92.9 & 87.1 & 85.8 & 77.9 & 49.7\\
\multicolumn{1}{@{}p{.35\linewidth}}{KVarN 2/2} & 94.4 & 94.4 & 90.4 & 85.5 & 82.5 & 68.6 & 49.3\\
\multicolumn{1}{@{}p{.35\linewidth}}{KIVI-2} & 94.7 & 93.8 & 91.7 & 85.6 & 83.2 & 71.6 & 49.2\\
\multicolumn{1}{@{}p{.35\linewidth}}{KVQuant 2-bit} & 94.8 & 93.4 & 88.8 & 84.8 & 82.3 & 69.6 & 48.8\\
\multicolumn{1}{@{}p{.35\linewidth}}{TurboQuant 2.5-bit} & 94.5 & 93.6 & 90.8 & 84.4 & 82.9 & 68.8 & 49.0\\
\multicolumn{1}{@{}p{.35\linewidth}}{AQUA-KV 2-bit} & 95.4 & 94.8 & 92.8 & 86.5 & 85.7 & 76.6 & 49.3\\
\multicolumn{1}{@{}p{.35\linewidth}}{KVQuant 3-bit} & 95.6 & 94.8 & 92.2 & 86.9 & 85.3 & 77.0 & 49.5\\
\multicolumn{1}{@{}p{.35\linewidth}}{TurboQuant 3.5-bit} & 95.7 & 94.9 & 92.6 & 87.1 & 85.3 & 76.7 & 49.5\\
\multicolumn{1}{@{}p{.35\linewidth}}{AQUA-KV 3-bit} & 95.6 & 95.0 & 92.8 & 87.0 & 85.8 & 77.7 & 49.6\\
\multicolumn{1}{@{}p{.35\linewidth}}{KVarN 4/2} & 95.5 & 94.8 & 92.8 & 87.1 & 85.9 & 77.2 & 49.6\\
\multicolumn{1}{@{}p{.35\linewidth}}{KIVI-4} & 95.6 & 95.3 & 93.1 & 87.0 & 85.8 & 78.0 & 49.6\\
\multicolumn{1}{@{}p{.35\linewidth}}{KVQuant 4-bit} & 95.7 & 95.1 & 93.0 & 87.1 & 85.7 & 77.5 & 49.6\\
\multicolumn{1}{@{}p{.35\linewidth}}{AQUA-KV 4-bit} & 95.7 & 95.1 & 92.9 & 87.0 & 85.8 & 77.8 & 49.5\\
\midrule
\appendixgroupreserve{5}
\multicolumn{8}{@{}l}{\textbf{\methodname{}-KV / DGX Spark}}\\
\cmidrule(lr){1-8}
\appendixgroupreserve{3}
2/2 & 95.8 & 95.2 & 92.7 & 87.5 & 85.1 & 75.2 & 49.6\\
3/2 & 95.7 & 95.0 & 93.2 & 87.1 & 85.8 & 77.6 & 49.5\\
3/3 & 95.8 & 95.2 & 93.2 & 87.0 & 85.8 & 77.5 & 49.6\\
\midrule
\appendixgroupreserve{5}
\multicolumn{8}{@{}l}{\textbf{\methodname{}-KV / RTX 5090}}\\
\cmidrule(lr){1-8}
2/2 & 95.9 & 95.1 & 92.9 & 87.1 & 86.0 & 78.1 & 49.6\\
3/2 & 95.7 & 95.1 & 93.1 & 87.0 & 85.8 & 78.3 & 49.5\\
3/3 & 95.7 & 95.0 & 93.0 & 86.9 & 85.8 & 78.2 & 49.5\cr
\end{supertabular*}\endgroup\par\vspace{3pt}
\begingroup\fontsize{7}{8.5}\selectfont\setlength{\tabcolsep}{.65pt}\renewcommand{\arraystretch}{1.0}\setlength{\aboverulesep}{.2ex}\setlength{\belowrulesep}{.25ex}
\tablehead{\toprule
\multicolumn{10}{@{}l}{\textbf{Qwen3-8B}}\\
\cmidrule(lr){1-10}
\textbf{Configuration} & \textbf{R4} & \textbf{R8} & \textbf{R16} & \textbf{R32} & \textbf{R64} & \textbf{R128} & \textbf{LB32} & \textbf{M500} & \textbf{AIME}\\\midrule}
\tabletail{\bottomrule}\tablelasttail{\bottomrule}
\begin{supertabular*}{\linewidth}{@{\extracolsep{\fill}}@{}lccccccccc@{}}
\multicolumn{1}{@{}l}{BF16 KV} & 95.2 & 93.8 & 93.8 & 92.1 & 80.6 & 74.1 & 49.4 & 97.2 & 71.0\\
\multicolumn{1}{@{}l}{KVarN 2/2} & 94.2 & 93.0 & 91.9 & 89.0 & 72.6 & 66.0 & 48.6 & 96.2 & 66.0\\
\multicolumn{1}{@{}l}{KIVI-2} & 94.0 & 92.9 & 92.1 & 88.8 & 73.0 & 67.0 & 48.8 & 96.0 & 64.8\\
\multicolumn{1}{@{}l}{KVQuant 2-bit} & 93.5 & 91.2 & 90.2 & 85.1 & 67.7 & 60.4 & 48.0 & 94.4 & 54.0\\
\multicolumn{1}{@{}l}{TurboQuant 2.5-bit} & 92.3 & 90.2 & 86.8 & 81.3 & 58.8 & 48.2 & 48.4 & 79.4 & 30.0\\
\multicolumn{1}{@{}l}{AQUA-KV 2-bit} & 94.9 & 93.7 & 93.4 & 91.9 & 79.8 & 72.9 & 49.2 & 96.0 & 70.4\\
\multicolumn{1}{@{}l}{KVQuant 3-bit} & 94.9 & 93.6 & 93.1 & 91.9 & 79.1 & 72.2 & 49.2 & 96.2 & 70.8\\
\multicolumn{1}{@{}l}{TurboQuant 3.5-bit} & 95.1 & 94.0 & 93.1 & 92.0 & 78.6 & 73.4 & 49.4 & 97.0 & 69.8\\
\multicolumn{1}{@{}l}{AQUA-KV 3-bit} & 95.3 & 94.1 & 93.6 & 92.2 & 80.5 & 74.5 & 49.3 & 97.2 & 71.9\\
\multicolumn{1}{@{}l}{KVarN 4/2} & 95.4 & 93.8 & 93.6 & 91.9 & 79.7 & 73.9 & 49.3 & 96.2 & 71.0\\
\multicolumn{1}{@{}l}{KIVI-4} & 95.0 & 94.0 & 93.5 & 92.1 & 80.8 & 74.2 & 49.4 & 96.0 & 71.3\\
\multicolumn{1}{@{}l}{KVQuant 4-bit} & 95.3 & 93.8 & 93.6 & 91.9 & 79.9 & 74.0 & 49.2 & 96.6 & 72.1\\
\multicolumn{1}{@{}l}{AQUA-KV 4-bit} & 95.2 & 93.9 & 93.6 & 92.1 & 80.7 & 74.2 & 49.3 & 95.6 & 70.2\\
\midrule
\appendixgroupreserve{5}
\multicolumn{10}{@{}l}{\textbf{\methodname{}-KV / DGX Spark}}\\
\cmidrule(lr){1-10}
\appendixgroupreserve{3}
2/2 & 95.4 & 94.0 & 93.4 & 91.6 & 77.2 & 71.4 & 49.0 & 96.0 & 68.3\\
3/2 & 95.2 & 93.7 & 93.7 & 92.0 & 79.5 & 74.0 & 49.4 & 95.0 & 70.2\\
3/3 & 95.3 & 93.8 & 93.8 & 92.2 & 79.8 & 74.2 & 49.3 & 96.2 & 69.4\\
\midrule
\appendixgroupreserve{5}
\multicolumn{10}{@{}l}{\textbf{\methodname{}-KV / RTX 5090}}\\
\cmidrule(lr){1-10}
2/2 & 95.5 & 94.0 & 93.8 & 91.8 & 79.9 & 74.2 & 49.3 & 95.6 & 69.2\\
3/2 & 95.3 & 93.9 & 93.8 & 92.0 & 80.4 & 74.7 & 49.3 & 96.2 & 70.0\\
3/3 & 95.3 & 93.9 & 93.8 & 92.1 & 80.3 & 74.5 & 49.3 & 97.0 & 71.5\cr
\end{supertabular*}\endgroup\par\vspace{3pt}
\end{appendixflowtable}

\begin{appendixflowtable}
\centering
\caption{KV-only refinement: selective KV task accuracy. R8--R128 give RULER accuracy at 8K--128K; LB32 uses LongBench prompts truncated to 32,128 tokens. M500/AIME report Qwen3-8B MATH500/AIME accuracy. ShadowKV and TailorKV do not compress KV produced during decoding and are omitted from the M500/AIME comparison. Platform headings distinguish \methodname{}-KV configurations; baseline accuracies are shared across devices.}
\label{tab:kv_downstream-2}
\begingroup\fontsize{7}{8.5}\selectfont\setlength{\tabcolsep}{.65pt}\renewcommand{\arraystretch}{1.0}\setlength{\aboverulesep}{.2ex}\setlength{\belowrulesep}{.25ex}
\tablehead{\toprule
\multicolumn{7}{@{}l}{\textbf{Llama-3.1-8B-Instruct}}\\
\cmidrule(lr){1-7}
\textbf{Configuration} & \textbf{R8} & \textbf{R16} & \textbf{R32} & \textbf{R64} & \textbf{R128} & \textbf{LB32}\\\midrule}
\tabletail{\bottomrule}\tablelasttail{\bottomrule}
\appendixfirstgroupreserve{5}
\begin{supertabular*}{\linewidth}{@{\extracolsep{\fill}}@{}p{.35\linewidth}cccccc@{}}
\multicolumn{1}{@{}p{.35\linewidth}}{BF16 KV} & 95.1 & 92.9 & 87.1 & 85.8 & 77.9 & 49.7\\
\multicolumn{1}{@{}p{.35\linewidth}}{ShadowKV} & 90.5 & 88.4 & 84.4 & 81.0 & 70.6 & 49.3\\
\multicolumn{1}{@{}p{.35\linewidth}}{ParisKV} & 93.1 & 90.2 & 85.6 & 78.5 & 68.9 & 49.6\\
\multicolumn{1}{@{}p{.35\linewidth}}{TailorKV-2} & 94.1 & 92.5 & 88.4 & 81.7 & 68.6 & 48.8\\
\multicolumn{1}{@{}p{.35\linewidth}}{TailorKV} & 93.6 & 91.8 & 87.3 & 77.8 & 65.2 & 48.6\\
\multicolumn{1}{@{}p{.35\linewidth}}{FreeKV} & 93.7 & 87.8 & 80.8 & 75.4 & 63.6 & 49.6\\
\multicolumn{1}{@{}p{.35\linewidth}}{SpeCache} & 95.0 & 92.6 & 87.1 & 85.5 & 74.0 & 49.5\\
\multicolumn{1}{@{}p{.35\linewidth}}{KVSwap} & 83.9 & 78.4 & 73.3 & 69.0 & 49.0 & 48.5\\
\midrule
\appendixgroupreserve{5}
\multicolumn{7}{@{}l}{\textbf{\methodname{}-KV / DGX Spark}}\\
\cmidrule(lr){1-7}
\appendixgroupreserve{3}
2/2 & 95.2 & 92.7 & 87.5 & 85.1 & 75.2 & 49.6\\
3/2 & 95.0 & 93.2 & 87.1 & 85.8 & 77.6 & 49.5\\
3/3 & 95.2 & 93.2 & 87.0 & 85.8 & 77.5 & 49.6\\
\midrule
\appendixgroupreserve{5}
\multicolumn{7}{@{}l}{\textbf{\methodname{}-KV / RTX 5090}}\\
\cmidrule(lr){1-7}
2/2 & 95.1 & 92.9 & 87.1 & 86.0 & 78.1 & 49.6\\
3/2 & 95.1 & 93.1 & 87.0 & 85.8 & 78.3 & 49.5\\
3/3 & 95.0 & 93.0 & 86.9 & 85.8 & 78.2 & 49.5\cr
\end{supertabular*}\endgroup\par\vspace{3pt}
\begingroup\fontsize{7}{8.5}\selectfont\setlength{\tabcolsep}{.65pt}\renewcommand{\arraystretch}{1.0}\setlength{\aboverulesep}{.2ex}\setlength{\belowrulesep}{.25ex}
\tablehead{\toprule
\multicolumn{9}{@{}l}{\textbf{Qwen3-8B}}\\
\cmidrule(lr){1-9}
\textbf{Configuration} & \textbf{R8} & \textbf{R16} & \textbf{R32} & \textbf{R64} & \textbf{R128} & \textbf{LB32} & \textbf{M500} & \textbf{AIME}\\\midrule}
\tabletail{\bottomrule}\tablelasttail{\bottomrule}
\appendixfirstgroupreserve{5}
\begin{supertabular*}{\linewidth}{@{\extracolsep{\fill}}@{}p{.24\linewidth}cccccccc@{}}
\multicolumn{1}{@{}p{.24\linewidth}}{BF16 KV} & 93.8 & 93.8 & 92.1 & 80.6 & 74.1 & 49.4 & 97.2 & 71.0\\
\multicolumn{1}{@{}p{.24\linewidth}}{ShadowKV} & 89.4 & 87.3 & 82.8 & 74.7 & 70.5 & 49.3 & \multicolumn{1}{c}{--} & \multicolumn{1}{c}{--}\\
\multicolumn{1}{@{}p{.24\linewidth}}{ParisKV} & 93.3 & 91.6 & 88.5 & 77.4 & 70.1 & 49.3 & 95.4 & 64.2\\
\multicolumn{1}{@{}p{.24\linewidth}}{TailorKV-2} & 93.7 & 93.1 & 90.5 & 78.1 & 71.4 & 49.1 & \multicolumn{1}{c}{--} & \multicolumn{1}{c}{--}\\
\multicolumn{1}{@{}p{.24\linewidth}}{FreeKV} & 91.1 & 86.4 & 82.2 & 68.7 & 54.2 & 49.1 & 96.8 & 63.5\\
\multicolumn{1}{@{}p{.24\linewidth}}{SpeCache} & 93.8 & 93.6 & 91.5 & 77.0 & 69.3 & 49.2 & 96.6 & 70.2\\
\multicolumn{1}{@{}p{.24\linewidth}}{KVSwap} & 88.6 & 84.0 & 78.3 & 68.6 & 59.6 & 49.1 & 94.0 & 52.7\\
\midrule
\appendixgroupreserve{5}
\multicolumn{9}{@{}l}{\textbf{\methodname{}-KV / DGX Spark}}\\
\cmidrule(lr){1-9}
\appendixgroupreserve{3}
2/2 & 94.0 & 93.4 & 91.6 & 77.2 & 71.4 & 49.0 & 96.0 & 68.3\\
3/2 & 93.7 & 93.7 & 92.0 & 79.5 & 74.0 & 49.4 & 95.0 & 70.2\\
3/3 & 93.8 & 93.8 & 92.2 & 79.8 & 74.2 & 49.3 & 96.2 & 69.4\\
\midrule
\appendixgroupreserve{5}
\multicolumn{9}{@{}l}{\textbf{\methodname{}-KV / RTX 5090}}\\
\cmidrule(lr){1-9}
2/2 & 94.0 & 93.8 & 91.8 & 79.9 & 74.2 & 49.3 & 95.6 & 69.2\\
3/2 & 93.9 & 93.8 & 92.0 & 80.4 & 74.7 & 49.3 & 96.2 & 70.0\\
3/3 & 93.9 & 93.8 & 92.1 & 80.3 & 74.5 & 49.3 & 97.0 & 71.5\cr
\end{supertabular*}\endgroup\par\vspace{3pt}
\end{appendixflowtable}

\begin{appendixflowtable}
\caption{KV-only refinement: PG-19 context sweep on Llama-3.1-8B-Instruct. $^{*}$GB counts only the resident KV representation and, on DGX Spark, its refinement RAM cache; weights and backing-tier storage are excluded. Throughput is tok/s at the column's context length.}\label{tab:kv_long_context-3}\label{tab:kv_long_context-1}

\begingroup\fontsize{6.8}{8.3}\selectfont\setlength{\tabcolsep}{.65pt}\renewcommand{\arraystretch}{1.0}
\tablehead{\toprule
\textbf{Configuration} & \textbf{Metric} & \textbf{8K} & \textbf{16K} & \textbf{32K} & \textbf{64K} & \textbf{128K}\\\midrule}
\tabletail{\bottomrule}\tablelasttail{\bottomrule}
\begin{supertabular*}{\linewidth}{@{\extracolsep{\fill}}@{}p{.24\linewidth}p{.17\linewidth}ccccc@{}}
\appendixgroupreserve{5}
\multirow[c]{5}{=}{BF16 KV} & PPL & 10.34 & 10.29 & 10.27 & 10.17 & 10.32\\
 & KL & 0.000 & 0.000 & 0.000 & 0.000 & 0.000\\
 & GB$^{*}$ & 1.07 & 2.15 & 4.30 & 8.59 & 17.18\\
 & Spark tok/s & 14.9 & 14.1 & 12.5 & 9.9 & 7.3\\
 & 5090 tok/s & 96.1 & 90.1 & 80.2 & 65.0 & 47.6\\
\midrule
\appendixgroupreserve{6}
\multicolumn{7}{@{}l}{\textbf{Quantization}}\\
\cmidrule(lr){1-7}
\appendixgroupreserve{4}
\multirow[c]{4}{=}{KVarN 2/2} & PPL & 10.74 & 10.73 & 10.74 & 10.64 & 10.98\\
 & KL & 0.045 & 0.049 & 0.051 & 0.052 & 0.067\\
 & Spark tok/s & 15.3 & 14.7 & 13.9 & 13.2 & 11.3\\
 & 5090 tok/s & 92.9 & 88.0 & 79.7 & 72.1 & 57.1\\
\midrule
\appendixgroupreserve{4}
\multirow[c]{4}{=}{KIVI-2} & PPL & 10.82 & 10.84 & 10.81 & 10.69 & 10.90\\
 & KL & 0.055 & 0.058 & 0.059 & 0.059 & 0.059\\
 & Spark tok/s & 14.7 & 14.0 & 12.9 & 11.3 & 8.8\\
 & 5090 tok/s & 82.1 & 72.5 & 58.9 & 42.7 & 27.4\\
\midrule
\appendixgroupreserve{4}
\multirow[c]{4}{=}{KVQuant 2-bit} & PPL & 11.15 & 11.11 & 11.12 & 11.04 & 11.71\\
 & KL & 0.084 & 0.086 & 0.087 & 0.089 & 0.125\\
 & Spark tok/s & 12.9 & 11.6 & 9.5 & 7.0 & 3.9\\
 & 5090 tok/s & 60.2 & 48.8 & 36.2 & 22.6 & 13.3\\
\midrule
\appendixgroupreserve{4}
\multirow[c]{4}{=}{TurboQuant 2.5-bit} & PPL & 10.77 & 10.76 & 10.78 & 10.67 & 10.98\\
 & KL & 0.040 & 0.042 & 0.045 & 0.045 & 0.055\\
 & Spark tok/s & 15.0 & 14.5 & 13.6 & 12.4 & 10.2\\
 & 5090 tok/s & 88.9 & 85.1 & 77.2 & 65.9 & 51.7\\
\midrule
\appendixgroupreserve{4}
\multirow[c]{4}{=}{AQUA-KV 2-bit} & PPL & 10.54 & 10.51 & 10.54 & 10.48 & 10.66\\
 & KL & 0.022 & 0.021 & 0.024 & 0.030 & 0.030\\
 & Spark tok/s & 9.8 & 6.9 & 4.5 & 2.7 & 1.5\\
 & 5090 tok/s & 44.4 & 30.6 & 17.9 & 9.1 & 5.1\\
\midrule
\appendixgroupreserve{4}
\multirow[c]{4}{=}{KVQuant 3-bit} & PPL & 10.52 & 10.47 & 10.46 & 10.35 & 10.56\\
 & KL & 0.020 & 0.021 & 0.021 & 0.021 & 0.024\\
 & Spark tok/s & 12.9 & 11.4 & 9.1 & 6.4 & 3.9\\
 & 5090 tok/s & 59.5 & 48.1 & 35.0 & 21.9 & 12.8\\
\midrule
\appendixgroupreserve{4}
\multirow[c]{4}{=}{TurboQuant 3.5-bit} & PPL & 10.42 & 10.37 & 10.35 & 10.26 & 10.43\\
 & KL & 0.007 & 0.007 & 0.008 & 0.008 & 0.010\\
 & Spark tok/s & 15.0 & 14.5 & 13.7 & 12.5 & 10.4\\
 & 5090 tok/s & 89.1 & 85.7 & 78.7 & 67.9 & 54.2\\
\midrule
\appendixgroupreserve{4}
\multirow[c]{4}{=}{AQUA-KV 3-bit} & PPL & 10.38 & 10.35 & 10.34 & 10.24 & 10.39\\
 & KL & 0.005 & 0.006 & 0.006 & 0.007 & 0.006\\
 & Spark tok/s & 9.7 & 6.8 & 4.5 & 2.6 & 1.5\\
 & 5090 tok/s & 44.4 & 30.5 & 17.9 & 9.1 & 5.0\\
\midrule
\appendixgroupreserve{4}
\multirow[c]{4}{=}{KVarN 4/2} & PPL & 10.38 & 10.34 & 10.32 & 10.22 & 10.38\\
 & KL & 0.007 & 0.007 & 0.007 & 0.007 & 0.008\\
 & Spark tok/s & 15.1 & 14.5 & 13.5 & 12.6 & 10.4\\
 & 5090 tok/s & 91.8 & 86.2 & 76.9 & 68.3 & 52.3\\
\midrule
\appendixgroupreserve{4}
\multirow[c]{4}{=}{KIVI-4} & PPL & 10.35 & 10.30 & 10.28 & 10.19 & 10.33\\
 & KL & 0.002 & 0.002 & 0.002 & 0.002 & 0.002\\
 & Spark tok/s & 14.7 & 14.0 & 12.9 & 11.1 & 8.6\\
 & 5090 tok/s & 83.4 & 74.4 & 61.0 & 44.8 & 29.2\\
\midrule
\appendixgroupreserve{4}
\multirow[c]{4}{=}{KVQuant 4-bit} & PPL & 10.38 & 10.34 & 10.32 & 10.22 & 10.38\\
 & KL & 0.006 & 0.006 & 0.006 & 0.006 & 0.007\\
 & Spark tok/s & 12.5 & 10.8 & 8.4 & 5.8 & 3.3\\
 & 5090 tok/s & 57.3 & 46.0 & 33.0 & 20.3 & 11.8\\
\midrule
\appendixgroupreserve{4}
\multirow[c]{4}{=}{AQUA-KV 4-bit} & PPL & 10.35 & 10.31 & 10.29 & 10.19 & 10.32\\
 & KL & 0.002 & 0.002 & 0.002 & 0.002 & 0.002\\
 & Spark tok/s & 9.7 & 6.8 & 4.4 & 2.6 & 1.5\\
 & 5090 tok/s & 44.2 & 30.4 & 17.8 & 9.1 & 5.0\\
\midrule
\appendixgroupreserve{6}
\multicolumn{7}{@{}l}{\textbf{KV offloading / DGX Spark}}\\
\cmidrule(lr){1-7}
\appendixgroupreserve{4}
\multirow[c]{4}{=}{KVSwap} & PPL & 10.87 & 10.99 & 11.10 & 11.09 & 16.27\\
 & KL & 0.056 & 0.072 & 0.087 & 0.097 & 0.477\\
 & GB$^{*}$ & 0.12 & 0.19 & 0.32 & 0.59 & 1.13\\
 & tok/s & 8.7 & 8.2 & 9.2 & 8.8 & 8.7\\
\midrule
\appendixgroupreserve{6}
\multicolumn{7}{@{}l}{\textbf{KV offloading / RTX 5090}}\\
\cmidrule(lr){1-7}
\appendixgroupreserve{4}
\multirow[c]{4}{=}{ShadowKV} & PPL & 10.69 & 10.66 & 10.65 & 10.60 & 11.24\\
 & KL & 0.042 & 0.044 & 0.045 & 0.048 & 0.086\\
 & GB$^{*}$ & 0.23 & 0.40 & 0.74 & 1.41 & 2.76\\
 & tok/s & 88.3 & 86.9 & 84.6 & 67.9 & 54.8\\
\midrule
\appendixgroupreserve{4}
\multirow[c]{4}{=}{ParisKV} & PPL & 10.59 & 10.61 & 10.62 & 10.60 & 13.07\\
 & KL & 0.045 & 0.060 & 0.072 & 0.085 & 0.282\\
 & GB$^{*}$ & 0.32 & 0.55 & 1.02 & 1.96 & 3.84\\
 & tok/s & 77.6 & 75.5 & 72.9 & 67.2 & 56.0\\
\midrule
\appendixgroupreserve{4}
\multirow[c]{4}{=}{TailorKV-2} & PPL & 10.38 & 10.36 & 10.35 & 10.25 & 11.21\\
 & KL & 0.018 & 0.027 & 0.035 & 0.040 & 0.107\\
 & GB$^{*}$ & 0.03 & 0.04 & 0.05 & 0.09 & 0.16\\
 & tok/s & 36.3 & 32.3 & 34.9 & 23.2 & 18.0\\
\midrule
\appendixgroupreserve{4}
\multirow[c]{4}{=}{FreeKV} & PPL & 10.45 & 10.50 & 10.58 & 10.61 & 11.62\\
 & KL & 0.015 & 0.026 & 0.038 & 0.052 & 0.126\\
 & GB$^{*}$ & 0.33 & 0.39 & 0.53 & 0.79 & 1.32\\
 & tok/s & 58.0 & 56.3 & 50.8 & 49.6 & 56.4\\
\midrule
\appendixgroupreserve{4}
\multirow[c]{4}{=}{SpeCache} & PPL & 10.45 & 10.42 & 10.42 & 10.34 & 10.68\\
 & KL & 0.013 & 0.016 & 0.018 & 0.021 & 0.035\\
 & GB$^{*}$ & 0.18 & 0.35 & 0.69 & 1.36 & 2.70\\
 & tok/s & 78.8 & 76.4 & 67.7 & 58.1 & 44.3\\
\midrule
\appendixgroupreserve{2}
\multirow[c]{2}{=}{TailorKV-1} & PPL & 11.46 & 11.59 & 11.58 & 11.60 & 13.71\\
 & KL & 0.128 & 0.149 & 0.157 & 0.173 & 0.322\\
\midrule
\appendixgroupreserve{6}
\multicolumn{7}{@{}l}{\textbf{\methodname{}-KV / DGX Spark}}\\
\cmidrule(lr){1-7}
\appendixgroupreserve{4}
\multirow[c]{4}{=}{\methodname{}-KV 2/2} & PPL & 10.42 & 10.39 & 10.37 & 10.27 & 10.47\\
 & KL & 0.007 & 0.006 & 0.006 & 0.006 & 0.011\\
 & GB$^{*}$ & 0.32 & 0.48 & 0.80 & 1.44 & 2.71\\
 & tok/s & 14.9 & 14.7 & 14.2 & 13.4 & 11.4\\
\midrule
\appendixgroupreserve{4}
\multirow[c]{4}{=}{\methodname{}-KV 3/2} & PPL & 10.35 & 10.31 & 10.29 & 10.19 & 10.34\\
 & KL & 0.002 & 0.002 & 0.002 & 0.002 & 0.002\\
 & GB$^{*}$ & 0.36 & 0.55 & 0.93 & 1.71 & 3.25\\
 & tok/s & 14.8 & 14.6 & 14.1 & 13.1 & 11.0\\
\midrule
\appendixgroupreserve{4}
\multirow[c]{4}{=}{\methodname{}-KV 3/3} & PPL & 10.35 & 10.30 & 10.28 & 10.18 & 10.33\\
 & KL & 0.001 & 0.001 & 0.001 & 0.001 & 0.001\\
 & GB$^{*}$ & 0.39 & 0.61 & 1.07 & 1.97 & 3.79\\
 & tok/s & 14.8 & 14.5 & 14.0 & 12.8 & 10.7\\
\midrule
\appendixgroupreserve{6}
\multicolumn{7}{@{}l}{\textbf{\methodname{}-KV / RTX 5090}}\\
\cmidrule(lr){1-7}
\appendixgroupreserve{4}
\multirow[c]{4}{=}{\methodname{}-KV 2/2} & PPL & 10.41 & 10.37 & 10.35 & 10.25 & 10.40\\
 & KL & 0.004 & 0.004 & 0.003 & 0.002 & 0.003\\
 & GB$^{*}$ & 0.19 & 0.35 & 0.67 & 1.30 & 2.58\\
 & tok/s & 88.0 & 85.4 & 81.8 & 75.7 & 66.6\\
\midrule
\appendixgroupreserve{4}
\multirow[c]{4}{=}{\methodname{}-KV 3/2} & PPL & 10.35 & 10.30 & 10.29 & 10.18 & 10.33\\
 & KL & 0.001 & 0.001 & 0.001 & 0.001 & 0.001\\
 & GB$^{*}$ & 0.22 & 0.41 & 0.80 & 1.57 & 3.12\\
 & tok/s & 87.5 & 84.9 & 81.2 & 75.1 & 65.7\\
\midrule
\appendixgroupreserve{4}
\multirow[c]{4}{=}{\methodname{}-KV 3/3} & PPL & 10.35 & 10.30 & 10.28 & 10.18 & 10.33\\
 & KL & 0.001 & 0.001 & 0.001 & 0.001 & 0.001\\
 & GB$^{*}$ & 0.25 & 0.48 & 0.93 & 1.84 & 3.65\\
 & tok/s & 87.5 & 84.6 & 80.8 & 74.4 & 65.1\cr
\end{supertabular*}\endgroup
\end{appendixflowtable}

\begin{appendixflowtable}
\caption{KV-only refinement: PG-19 context sweep on Qwen3-8B. $^{*}$GB counts only the resident KV representation and, on DGX Spark, its refinement RAM cache; weights and backing-tier storage are excluded. Throughput is tok/s at the column's context length.}\label{tab:kv_long_context-4}\label{tab:kv_long_context-2}

\begingroup\fontsize{6.8}{8.3}\selectfont\setlength{\tabcolsep}{.65pt}\renewcommand{\arraystretch}{1.0}
\tablehead{\toprule
\textbf{Configuration} & \textbf{Metric} & \textbf{8K} & \textbf{16K} & \textbf{32K} & \textbf{64K} & \textbf{128K}\\\midrule}
\tabletail{\bottomrule}\tablelasttail{\bottomrule}
\begin{supertabular*}{\linewidth}{@{\extracolsep{\fill}}@{}p{.24\linewidth}p{.17\linewidth}ccccc@{}}
\appendixgroupreserve{5}
\multirow[c]{5}{=}{BF16 KV} & PPL & 13.23 & 12.91 & 12.64 & 12.99 & 12.89\\
 & KL & 0.000 & 0.000 & 0.000 & 0.000 & 0.000\\
 & GB$^{*}$ & 1.21 & 2.42 & 4.83 & 9.66 & 19.33\\
 & Spark tok/s & 14.7 & 13.7 & 12.1 & 9.4 & 6.7\\
 & 5090 tok/s & 93.8 & 87.5 & 77.0 & 61.4 & OOM\\
\midrule
\appendixgroupreserve{6}
\multicolumn{7}{@{}l}{\textbf{Quantization}}\\
\cmidrule(lr){1-7}
\appendixgroupreserve{4}
\multirow[c]{4}{=}{KVarN 2/2} & PPL & 13.77 & 13.47 & 13.11 & 14.44 & 14.40\\
 & KL & 0.062 & 0.065 & 0.059 & 0.131 & 0.130\\
 & Spark tok/s & 14.9 & 14.4 & 13.6 & 12.7 & 10.7\\
 & 5090 tok/s & 89.9 & 84.9 & 76.4 & 68.5 & 53.4\\
\midrule
\appendixgroupreserve{4}
\multirow[c]{4}{=}{KIVI-2} & PPL & 13.56 & 13.25 & 13.01 & 13.88 & 13.85\\
 & KL & 0.048 & 0.048 & 0.051 & 0.090 & 0.106\\
 & Spark tok/s & 14.4 & 13.7 & 12.5 & 10.7 & 8.3\\
 & 5090 tok/s & 78.6 & 69.1 & 55.3 & 39.4 & 25.0\\
\midrule
\appendixgroupreserve{4}
\multirow[c]{4}{=}{KVQuant 2-bit} & PPL & 14.48 & 14.18 & 13.97 & 14.39 & 14.40\\
 & KL & 0.117 & 0.124 & 0.129 & 0.158 & 0.167\\
 & Spark tok/s & 12.8 & 11.1 & 8.8 & 6.3 & 4.1\\
 & 5090 tok/s & 57.6 & 47.3 & 34.7 & 21.7 & 13.2\\
\midrule
\appendixgroupreserve{4}
\multirow[c]{4}{=}{TurboQuant 2.5-bit} & PPL & 20.66 & 21.20 & 21.82 & 33.85 & 36.02\\
 & KL & 0.582 & 0.631 & 0.685 & 1.047 & 1.115\\
 & Spark tok/s & 14.7 & 14.2 & 13.2 & 11.9 & 9.6\\
 & 5090 tok/s & 86.0 & 81.9 & 73.6 & 62.2 & 48.1\\
\midrule
\appendixgroupreserve{4}
\multirow[c]{4}{=}{AQUA-KV 2-bit} & PPL & 13.40 & 13.11 & 12.88 & 13.21 & 13.14\\
 & KL & 0.012 & 0.013 & 0.013 & 0.028 & 0.029\\
 & Spark tok/s & 9.3 & 6.5 & 4.1 & 2.4 & 1.4\\
 & 5090 tok/s & 41.1 & 27.9 & 16.2 & 8.1 & 4.5\\
\midrule
\appendixgroupreserve{4}
\multirow[c]{4}{=}{KVQuant 3-bit} & PPL & 13.47 & 13.20 & 12.90 & 13.29 & 13.21\\
 & KL & 0.026 & 0.027 & 0.028 & 0.039 & 0.041\\
 & Spark tok/s & 12.6 & 10.8 & 8.5 & 6.4 & 3.9\\
 & 5090 tok/s & 56.6 & 45.8 & 33.5 & 21.3 & 12.7\\
\midrule
\appendixgroupreserve{4}
\multirow[c]{4}{=}{TurboQuant 3.5-bit} & PPL & 13.80 & 13.56 & 13.12 & 14.32 & 13.99\\
 & KL & 0.057 & 0.067 & 0.051 & 0.118 & 0.102\\
 & Spark tok/s & 14.7 & 14.2 & 13.3 & 12.0 & 9.9\\
 & 5090 tok/s & 86.3 & 82.4 & 75.4 & 64.3 & 50.7\\
\midrule
\appendixgroupreserve{4}
\multirow[c]{4}{=}{AQUA-KV 3-bit} & PPL & 13.27 & 12.96 & 12.69 & 13.08 & 12.98\\
 & KL & 0.005 & 0.006 & 0.006 & 0.017 & 0.017\\
 & Spark tok/s & 9.3 & 6.4 & 4.1 & 2.4 & 1.4\\
 & 5090 tok/s & 41.0 & 27.8 & 16.1 & 8.1 & 4.5\\
\midrule
\appendixgroupreserve{4}
\multirow[c]{4}{=}{KVarN 4/2} & PPL & 13.32 & 13.00 & 12.72 & 13.15 & 13.04\\
 & KL & 0.008 & 0.008 & 0.009 & 0.014 & 0.014\\
 & Spark tok/s & 14.8 & 14.2 & 13.2 & 12.0 & 9.8\\
 & 5090 tok/s & 88.8 & 82.9 & 73.3 & 64.7 & 48.8\\
\midrule
\appendixgroupreserve{4}
\multirow[c]{4}{=}{KIVI-4} & PPL & 13.23 & 12.92 & 12.64 & 13.01 & 12.90\\
 & KL & 0.002 & 0.002 & 0.002 & 0.004 & 0.004\\
 & Spark tok/s & 14.4 & 13.7 & 12.5 & 10.6 & 8.0\\
 & 5090 tok/s & 80.1 & 70.8 & 57.5 & 41.4 & 26.6\\
\midrule
\appendixgroupreserve{4}
\multirow[c]{4}{=}{KVQuant 4-bit} & PPL & 13.30 & 12.97 & 12.69 & 13.04 & 12.96\\
 & KL & 0.007 & 0.007 & 0.007 & 0.011 & 0.011\\
 & Spark tok/s & 12.3 & 10.3 & 8.2 & 5.6 & 3.4\\
 & 5090 tok/s & 54.5 & 43.5 & 31.4 & 19.3 & 11.5\\
\midrule
\appendixgroupreserve{4}
\multirow[c]{4}{=}{AQUA-KV 4-bit} & PPL & 13.24 & 12.93 & 12.65 & 13.05 & 12.95\\
 & KL & 0.004 & 0.004 & 0.004 & 0.016 & 0.016\\
 & Spark tok/s & 9.3 & 6.4 & 4.0 & 2.4 & 1.3\\
 & 5090 tok/s & 41.0 & 27.8 & 16.1 & 8.1 & 4.5\\
\midrule
\appendixgroupreserve{6}
\multicolumn{7}{@{}l}{\textbf{KV offloading / DGX Spark}}\\
\cmidrule(lr){1-7}
\appendixgroupreserve{4}
\multirow[c]{4}{=}{KVSwap} & PPL & 14.09 & 14.10 & 14.21 & 15.64 & 15.82\\
 & KL & 0.074 & 0.101 & 0.133 & 0.259 & 0.289\\
 & GB$^{*}$ & 0.14 & 0.21 & 0.36 & 0.66 & 1.27\\
 & tok/s & 7.4 & 7.7 & 7.7 & 7.6 & 7.5\\
\midrule
\appendixgroupreserve{6}
\multicolumn{7}{@{}l}{\textbf{KV offloading / RTX 5090}}\\
\cmidrule(lr){1-7}
\appendixgroupreserve{4}
\multirow[c]{4}{=}{ShadowKV} & PPL & 15.78 & 14.85 & 13.92 & 15.65 & 15.10\\
 & KL & 0.229 & 0.193 & 0.145 & 0.214 & 0.187\\
 & GB$^{*}$ & 0.26 & 0.45 & 0.83 & 1.59 & 3.11\\
 & tok/s & 84.9 & 83.3 & 81.1 & 62.5 & 50.7\\
\midrule
\appendixgroupreserve{4}
\multirow[c]{4}{=}{ParisKV} & PPL & 13.34 & 13.10 & 12.87 & 14.09 & 14.11\\
 & KL & 0.050 & 0.064 & 0.070 & 0.140 & 0.154\\
 & GB$^{*}$ & 0.36 & 0.62 & 1.15 & 2.21 & 4.32\\
 & tok/s & 70.8 & 71.6 & 68.9 & 62.8 & 51.6\\
\midrule
\appendixgroupreserve{4}
\multirow[c]{4}{=}{TailorKV-2} & PPL & 13.09 & 12.85 & 12.58 & 13.62 & 13.66\\
 & KL & 0.017 & 0.030 & 0.037 & 0.082 & 0.098\\
 & GB$^{*}$ & 0.03 & 0.03 & 0.04 & 0.05 & 0.07\\
 & tok/s & 24.5 & 25.7 & 21.8 & 16.7 & 15.0\\
\midrule
\appendixgroupreserve{4}
\multirow[c]{4}{=}{FreeKV} & PPL & 13.45 & 13.34 & 13.30 & 14.37 & 14.68\\
 & KL & 0.021 & 0.040 & 0.062 & 0.146 & 0.184\\
 & GB$^{*}$ & 0.36 & 0.43 & 0.58 & 0.86 & 1.42\\
 & tok/s & 47.3 & 38.9 & 35.5 & 31.5 & 30.3\\
\midrule
\appendixgroupreserve{5}
\multirow[c]{4}{=}{SpeCache} & PPL & 13.43 & 13.14 & 12.94 & 13.74 & 13.96\\
 & KL & 0.026 & 0.030 & 0.038 & 0.080 & 0.106\\
 & GB$^{*}$ & 0.21 & 0.39 & 0.77 & 1.53 & 3.04\\
 & tok/s & 73.4 & 70.9 & 63.0 & 53.1 & 40.5\\
\midrule
\appendixgroupreserve{6}
\multicolumn{7}{@{}l}{\textbf{\methodname{}-KV / DGX Spark}}\\
\cmidrule(lr){1-7}
\appendixgroupreserve{4}
\multirow[c]{4}{=}{\methodname{}-KV 2/2} & PPL & 13.39 & 13.09 & 12.82 & 13.34 & 13.32\\
 & KL & 0.011 & 0.010 & 0.009 & 0.028 & 0.030\\
 & GB$^{*}$ & 0.35 & 0.53 & 0.88 & 1.60 & 3.04\\
 & tok/s & 14.6 & 14.4 & 13.9 & 13.0 & 10.9\\
\midrule
\appendixgroupreserve{4}
\multirow[c]{4}{=}{\methodname{}-KV 3/2} & PPL & 13.28 & 12.96 & 12.69 & 13.07 & 12.98\\
 & KL & 0.002 & 0.002 & 0.002 & 0.005 & 0.005\\
 & GB$^{*}$ & 0.38 & 0.60 & 1.03 & 1.90 & 3.64\\
 & tok/s & 14.6 & 14.3 & 13.7 & 12.6 & 10.5\\
\midrule
\appendixgroupreserve{4}
\multirow[c]{4}{=}{\methodname{}-KV 3/3} & PPL & 13.25 & 12.93 & 12.66 & 13.03 & 12.93\\
 & KL & 0.001 & 0.001 & 0.001 & 0.003 & 0.004\\
 & GB$^{*}$ & 0.42 & 0.67 & 1.18 & 2.20 & 4.24\\
 & tok/s & 14.5 & 14.2 & 13.7 & 12.4 & 10.2\\
\midrule
\appendixgroupreserve{6}
\multicolumn{7}{@{}l}{\textbf{\methodname{}-KV / RTX 5090}}\\
\cmidrule(lr){1-7}
\appendixgroupreserve{4}
\multirow[c]{4}{=}{\methodname{}-KV 2/2} & PPL & 13.39 & 13.07 & 12.80 & 13.07 & 12.98\\
 & KL & 0.007 & 0.005 & 0.004 & 0.009 & 0.007\\
 & GB$^{*}$ & 0.21 & 0.39 & 0.75 & 1.47 & 2.90\\
 & tok/s & 84.7 & 82.2 & 78.5 & 72.3 & 62.7\\
\midrule
\appendixgroupreserve{4}
\multirow[c]{4}{=}{\methodname{}-KV 3/2} & PPL & 13.28 & 12.96 & 12.68 & 13.02 & 12.92\\
 & KL & 0.002 & 0.001 & 0.001 & 0.002 & 0.002\\
 & GB$^{*}$ & 0.25 & 0.47 & 0.90 & 1.77 & 3.50\\
 & tok/s & 84.2 & 81.6 & 77.9 & 71.5 & 61.9\\
\midrule
\appendixgroupreserve{4}
\multirow[c]{4}{=}{\methodname{}-KV 3/3} & PPL & 13.25 & 12.93 & 12.66 & 12.99 & 12.90\\
 & KL & 0.001 & 0.001 & 0.001 & 0.001 & 0.001\\
 & GB$^{*}$ & 0.29 & 0.54 & 1.05 & 2.07 & 4.11\\
 & tok/s & 84.1 & 81.3 & 77.4 & 70.9 & 61.4\cr
\end{supertabular*}\endgroup
\end{appendixflowtable}

\begin{minipage}{\linewidth}
\setlength{\parskip}{6pt}
\subsection{Alternative Attention Architectures and MoE Models}
\begin{appendixcolumntable}
\centering
\caption{KV-only refinement: PG-19 perplexity/KL and RULER accuracy at 32K, plus MATH500 accuracy for Qwen3.5-9B. The comparison covers hybrid linear/full attention (Qwen3.5-9B) and MoE models with standard attention (Qwen3-30B-A3B) or latent attention (DeepSeek-V2-Lite). Refinement applies only to full attention. For DeepSeek, 2/2 and 3/3 indicate 2- and 3-bit latent precision; 3/2 is inapplicable because the shared latent representation uses one bitwidth.}
\label{tab:model_coverage-1}
\begingroup\fontsize{6.7}{8.1}\selectfont\setlength{\tabcolsep}{.35pt}\renewcommand{\arraystretch}{1.0}\setlength{\aboverulesep}{.2ex}\setlength{\belowrulesep}{.25ex}
\begin{tabular*}{\linewidth}{@{\extracolsep{\fill}}@{}llcccccccccc@{}}
\toprule
 & & \multicolumn{4}{c}{\textbf{Qwen3.5-9B}} & \multicolumn{3}{c}{\textbf{Qwen3-30B-A3B}} & \multicolumn{3}{c}{\textbf{DeepSeek-V2-Lite}}\\
\cmidrule(lr){3-6}\cmidrule(lr){7-9}\cmidrule(lr){10-12}
\textbf{K/V} & \textbf{Method} & \textbf{PPL} & \textbf{KL} & \textbf{R32} & \textbf{M500} & \textbf{PPL} & \textbf{KL} & \textbf{R32} & \textbf{PPL} & \textbf{KL} & \textbf{R32}\\\midrule
 & BF16 & 9.48 & 0.000 & 96.6 & 85.6 & 9.29 & 0.000 & 94.1 & 9.12 & 0.000 & 67.5\\
\midrule
\multirow[c]{2}{*}{2/2} & Base & 9.71 & 0.027 & 96.3 & 83.0 & 10.06 & 0.086 & 89.9 & 9.59 & 0.138 & 59.9\\
 & \methodname{}-KV & 9.54 & 0.004 & 96.5 & 88.0 & 9.55 & 0.021 & 93.6 & 9.17 & 0.017 & 67.2\\
\midrule
\multirow[c]{2}{*}{3/2} & Base & 9.62 & 0.014 & 96.5 & 84.6 & 9.42 & 0.016 & 93.9 & \multicolumn{3}{c}{\multirow[c]{2}{*}{--}}\\
 & \methodname{}-KV & 9.51 & 0.003 & 96.5 & 86.6 & 9.35 & 0.004 & 93.9 & \multicolumn{3}{c}{}\\
\midrule
\multirow[c]{2}{*}{3/3} & Base & 9.50 & 0.004 & 96.5 & 86.6 & 9.34 & 0.009 & 93.9 & 9.16 & 0.026 & 66.5\\
 & \methodname{}-KV & 9.48 & 0.001 & 96.5 & 89.6 & 9.32 & 0.002 & 94.0 & 9.13 & 0.006 & 67.2\\
\bottomrule\end{tabular*}\endgroup\par\vspace{3pt}
\end{appendixcolumntable}

\end{minipage}
Table~\ref{tab:model_coverage-1} groups the three architectures; Tables~\ref{tab:coverage_kv-1}--\ref{tab:coverage_kv-3} give per-model task results.
\begin{appendixcolumntable}
\centering
\caption{KV-only refinement: accuracy on Qwen3.5-9B. R4--R32 are RULER context lengths in thousands of tokens. M500 is MATH500. KVQuant uses C4 calibration and retains 1\% outliers. TurboQuant presets mix bitwidths across channels. \methodname{}-KV uses two-token groups, 1/128 selection, and no cache misses on dedicated GPUs; Base disables refinement.}
\label{tab:coverage_kv-1}
\fontsize{7}{8.5}\selectfont\setlength{\tabcolsep}{.9pt}\renewcommand{\arraystretch}{1.0}\setlength{\aboverulesep}{.2ex}\setlength{\belowrulesep}{.25ex}
\begin{tabular*}{\linewidth}{@{\extracolsep{\fill}}@{}lcccccc@{}}
\toprule
\textbf{Method} & \textbf{Bits} & \textbf{R4} & \textbf{R8} & \textbf{R16} & \textbf{R32} & \textbf{M500}\\\midrule
BF16 KV & 16/16 & 97.1 & 97.1 & 96.6 & 96.6 & 85.6\\
\midrule
\multirow[c]{2}{*}{KIVI} & 2 & 97.2 & 97.0 & 96.7 & 96.4 & 81.6\\
 & 4 & 97.1 & 97.1 & 96.6 & 96.5 & 85.2\\
\midrule
\multirow[c]{3}{*}{KVQuant} & 2 & 96.9 & 97.0 & 96.6 & 96.6 & 94.2\\
 & 3 & 97.1 & 97.0 & 96.7 & 96.7 & 86.2\\
 & 4 & 97.1 & 97.1 & 96.6 & 96.5 & 89.0\\
\midrule
\multirow[c]{2}{*}{TurboQuant} & 2.5 & 97.1 & 96.9 & 96.5 & 96.4 & 88.8\\
 & 3.5 & 97.1 & 97.2 & 96.6 & 96.5 & 86.2\\
\midrule
\multirow[c]{3}{*}{AQUA-KV} & 2 & 97.1 & 97.0 & 96.9 & 96.9 & 87.4\\
 & 3 & 97.1 & 97.1 & 96.7 & 96.7 & 83.8\\
 & 4 & 97.1 & 97.1 & 96.7 & 96.7 & 84.6\\
\midrule
Base & \multirow[c]{2}{*}{2/2} & 97.1 & 97.0 & 96.5 & 96.3 & 83.0\\
\methodname{}-KV &  & 97.1 & 97.2 & 96.6 & 96.5 & 88.0\\
\cmidrule(lr){1-7}
Base & \multirow[c]{2}{*}{3/2} & 97.1 & 97.1 & 96.4 & 96.5 & 84.6\\
\methodname{}-KV &  & 97.1 & 97.2 & 96.6 & 96.5 & 86.6\\
\cmidrule(lr){1-7}
Base & \multirow[c]{2}{*}{3/3} & 97.1 & 97.2 & 96.6 & 96.5 & 86.6\\
\methodname{}-KV &  & 97.1 & 97.1 & 96.6 & 96.5 & 89.6\\
\bottomrule
\end{tabular*}
\end{appendixcolumntable}

\begin{appendixcolumntable}
\centering

\centering
\caption{KV-only refinement: accuracy on DeepSeek-V2-Lite. R4--R32 are RULER context lengths in thousands of tokens. KVQuant uses C4 calibration and retains 1\% outliers. TurboQuant presets mix bitwidths across channels. \methodname{}-KV uses two-token groups, 1/128 selection, and no cache misses on dedicated GPUs; Base disables refinement. \methodname{}-KV bitwidths give latent precision, rather than separate K/V widths.}
\label{tab:coverage_kv-2}
\begingroup\fontsize{7}{8.5}\selectfont\setlength{\tabcolsep}{.9pt}\renewcommand{\arraystretch}{1.0}\setlength{\aboverulesep}{.2ex}\setlength{\belowrulesep}{.25ex}
\begin{tabular*}{\linewidth}{@{\extracolsep{\fill}}@{}lccccc@{}}
\toprule
\textbf{Method} & \textbf{Bits} & \textbf{R4} & \textbf{R8} & \textbf{R16} & \textbf{R32}\\\midrule
BF16 KV & 16/16 & 91.1 & 85.2 & 77.8 & 67.5\\
\midrule
\multirow[c]{2}{*}{KIVI} & 2 & 86.7 & 78.6 & 69.8 & 58.5\\
 & 4 & 90.4 & 85.3 & 77.7 & 67.3\\
\midrule
\multirow[c]{3}{*}{KVQuant} & 2 & 86.4 & 75.4 & 65.4 & 53.2\\
 & 3 & 90.2 & 83.1 & 76.2 & 64.0\\
 & 4 & 90.9 & 84.4 & 78.1 & 66.9\\
\midrule
\multirow[c]{2}{*}{TurboQuant} & 2.5 & 84.4 & 77.3 & 69.0 & 52.5\\
 & 3.5 & 90.9 & 84.3 & 77.0 & 66.0\\
\midrule
\multirow[c]{3}{*}{AQUA-KV} & 2 & 90.7 & 82.5 & 74.8 & 63.8\\
 & 3 & 91.2 & 84.8 & 77.5 & 65.8\\
 & 4 & 90.9 & 85.1 & 78.4 & 66.7\\
\midrule
Base & \multirow[c]{2}{*}{2} & 86.8 & 79.2 & 70.9 & 59.9\\
\methodname{}-KV &  & 90.4 & 84.5 & 78.8 & 67.2\\
\cmidrule(lr){1-6}
Base & \multirow[c]{2}{*}{3} & 90.6 & 84.1 & 77.5 & 66.5\\
\methodname{}-KV &  & 90.7 & 85.2 & 78.2 & 67.2\\
\bottomrule
\end{tabular*}
\endgroup
\end{appendixcolumntable}

\begin{appendixcolumntable}
\centering
\caption{KV-only refinement: accuracy on Qwen3-30B-A3B. R4--R32 are RULER context lengths in thousands of tokens. KVQuant uses C4 calibration and retains 1\% outliers. TurboQuant presets mix bitwidths across channels. \methodname{}-KV uses two-token groups, 1/128 selection, and no cache misses on dedicated GPUs; Base disables refinement.}
\label{tab:coverage_kv-3}
\begingroup\fontsize{7}{8.5}\selectfont\setlength{\tabcolsep}{.9pt}\renewcommand{\arraystretch}{1.0}\setlength{\aboverulesep}{.2ex}\setlength{\belowrulesep}{.25ex}
\begin{tabular*}{\linewidth}{@{\extracolsep{\fill}}@{}lccccc@{}}
\toprule
\textbf{Method} & \textbf{Bits} & \textbf{R4} & \textbf{R8} & \textbf{R16} & \textbf{R32}\\\midrule
BF16 KV & 16/16 & 95.6 & 95.8 & 95.1 & 94.1\\
\midrule
\multirow[c]{2}{*}{KIVI} & 2 & 94.6 & 95.0 & 93.1 & 92.3\\
 & 4 & 95.5 & 95.9 & 94.8 & 94.1\\
\midrule
\multirow[c]{3}{*}{KVQuant} & 2 & 93.9 & 93.9 & 91.9 & 90.7\\
 & 3 & 95.3 & 95.3 & 94.6 & 94.2\\
 & 4 & 95.5 & 95.6 & 94.8 & 94.2\\
\midrule
\multirow[c]{2}{*}{TurboQuant} & 2.5 & 93.0 & 91.4 & 91.1 & 85.2\\
 & 3.5 & 95.3 & 95.7 & 94.7 & 93.6\\
\midrule
\multirow[c]{3}{*}{AQUA-KV} & 2 & 95.6 & 95.7 & 94.8 & 94.1\\
 & 3 & 95.6 & 95.8 & 94.7 & 94.1\\
 & 4 & 95.7 & 95.8 & 94.7 & 94.2\\
\midrule
Base & \multirow[c]{2}{*}{2/2} & 93.4 & 93.3 & 92.0 & 89.9\\
\methodname{}-KV &  & 95.6 & 95.8 & 95.0 & 93.6\\
\cmidrule(lr){1-6}
Base & \multirow[c]{2}{*}{3/2} & 95.5 & 95.6 & 94.9 & 93.9\\
\methodname{}-KV &  & 95.7 & 95.9 & 94.8 & 93.9\\
\cmidrule(lr){1-6}
Base & \multirow[c]{2}{*}{3/3} & 95.6 & 95.7 & 95.1 & 93.9\\
\methodname{}-KV &  & 95.6 & 95.8 & 94.9 & 94.0\\
\bottomrule
\end{tabular*}
\endgroup

\end{appendixcolumntable}

\appendixfirstgroupreserve{13}
\subsection{KV Representation and Selection Controls}
Tables~\ref{tab:kv-controls-llama} and~\ref{tab:kv-controls-qwen} vary the retained FP16 tokens and the attention-score aggregation used for token-group selection.
\begingroup\let\controlsfirstreserve\appendixfirstgroupreserve\renewcommand{\appendixfirstgroupreserve}[1]{\controlsfirstreserve{1}}
\begin{appendixflowtable}\centering
\begingroup\setlength{\tabcolsep}{.8pt}\renewcommand{\arraystretch}{1.0}\setlength{\aboverulesep}{.2ex}\setlength{\belowrulesep}{.25ex}\fontsize{7}{8.5}\selectfont
\caption{KV representation and selection controls: Llama-3.1-8B-Instruct, PG-19 at 32K and 128K. Four-token groups, 1/128 fetch fraction. We retain the first 128 tokens in FP16 following KVarN~\citep{muller2026kvarn} and select by the maximum score over query heads; controls change one setting.}\label{tab:kv-controls-llama}
\tablehead{\toprule
\begin{tabular}[b]{@{}l@{}}\textbf{GPU resident}\\\textbf{KV}\end{tabular} & \textbf{Configuration} & \textbf{32K PPL} & \textbf{KL} & \textbf{128K PPL} & \textbf{KL}\\\midrule}
\tabletail{\bottomrule}\tablelasttail{\bottomrule}
\begin{supertabular*}{\linewidth}{@{\extracolsep{\fill}}@{}p{.19\linewidth}p{.35\linewidth}cccc@{}}
\appendixgroupreserve{3}
\multirow[c]{3}{=}{2/2} & Reference & 10.36 & 0.004 & 10.42 & 0.003\\
 & First 128 tokens quantized & 10.36 & 0.004 & 10.41 & 0.004\\
 & Mean over query heads & 10.36 & 0.004 & 10.41 & 0.003\\
\midrule
\appendixgroupreserve{3}
\multirow[c]{3}{=}{3/2} & Reference & 10.29 & 0.001 & 10.33 & 0.001\\
 & First 128 tokens quantized & 10.29 & 0.001 & 10.33 & 0.001\\
 & Mean over query heads & 10.29 & 0.001 & 10.33 & 0.001\\
\midrule
\appendixgroupreserve{3}
\multirow[c]{3}{=}{3/3} & Reference & 10.29 & 0.001 & 10.33 & 0.001\\
 & First 128 tokens quantized & 10.29 & 0.001 & 10.33 & 0.001\\
 & Mean over query heads & 10.28 & 0.001 & 10.33 & 0.001\\
\end{supertabular*}\endgroup
\end{appendixflowtable}
\begin{appendixflowtable}\centering
\begingroup\setlength{\tabcolsep}{.8pt}\renewcommand{\arraystretch}{1.0}\setlength{\aboverulesep}{.2ex}\setlength{\belowrulesep}{.25ex}\fontsize{7}{8.5}\selectfont
\caption{KV representation and selection controls: Qwen3-8B, PG-19 at 32K and 128K. Four-token groups, 1/128 fetch fraction. Reference shows our configuration, where we retain the first 128 tokens in FP16 following KVarN~\citep{muller2026kvarn} and select token groups using the maximum score over query heads. Each control changes one of these settings.}\label{tab:kv-controls-qwen}
\tablehead{\toprule
\begin{tabular}[b]{@{}l@{}}\textbf{GPU resident}\\\textbf{KV}\end{tabular} & \textbf{Configuration} & \textbf{32K PPL} & \textbf{KL} & \textbf{128K PPL} & \textbf{KL}\\\midrule}
\tabletail{\bottomrule}\tablelasttail{\bottomrule}
\begin{supertabular*}{\linewidth}{@{\extracolsep{\fill}}@{}p{.19\linewidth}p{.35\linewidth}cccc@{}}
\appendixgroupreserve{3}
\multirow[c]{3}{=}{2/2} & Reference & 12.81 & 0.006 & 13.01 & 0.009\\
 & First 128 tokens quantized & 12.81 & 0.006 & 13.00 & 0.010\\
 & Mean over query heads & 12.80 & 0.006 & 13.01 & 0.009\\
\midrule
\appendixgroupreserve{3}
\multirow[c]{3}{=}{3/2} & Reference & 12.69 & 0.001 & 12.93 & 0.002\\
 & First 128 tokens quantized & 12.69 & 0.002 & 12.93 & 0.002\\
 & Mean over query heads & 12.68 & 0.001 & 12.93 & 0.002\\
\midrule
\appendixgroupreserve{3}
\multirow[c]{3}{=}{3/3} & Reference & 12.66 & 0.001 & 12.90 & 0.002\\
 & First 128 tokens quantized & 12.66 & 0.001 & 12.90 & 0.002\\
 & Mean over query heads & 12.66 & 0.001 & 12.90 & 0.002\\
\end{supertabular*}\endgroup
\end{appendixflowtable}

\endgroup

\subsection{Fetch Budgets, RAM Caches, and Miss Policies}
\begin{figure}[H]
\centering
\centering
\includegraphics[width=\columnwidth]{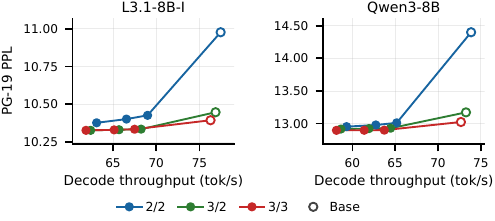}
\caption{KV fetch-fraction sweep: PG-19 PPL versus RTX 5090 throughput at 128K. Curves identify resident K/V bitwidths; fractions 1/256, 1/128, and 1/64 run from highest to lowest throughput. \methodname{}-KV uses two-token groups and +4/+4 bits. Open markers show the matching bases without refinement.}
\label{fig:kv-fetch-budget}
\end{figure}

\begin{figure}[H]
\centering
\centering
\includegraphics[width=\columnwidth]{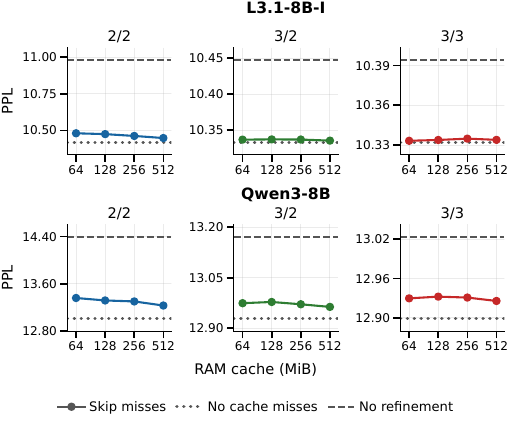}
\caption{KV RAM-cache capacity by resident bitwidth on DGX Spark: PG-19 PPL at 128K with four-token groups, +4/+4 bits, 1/128 selection, and skipped misses. Dotted lines assume no cache misses; dashed lines show the matching no-refinement bases. }
\label{fig:kv-cache-capacity}
\end{figure}

Figure~\ref{fig:kv-fetch-budget} varies the host-RAM fetch fraction on RTX 5090; Figure~\ref{fig:kv-cache-capacity} varies RAM-cache capacity on DGX Spark. Table~\ref{tab:kv-wait-cache} reports throughput when waiting on cache misses.
Table~\ref{tab:kv_ssd_tradeoff-1} compares SSD-backed configurations; Tables~\ref{tab:kv-cache-grid-llama}--\ref{tab:kv-nomiss-qwen} report cache-capacity and no-cache-miss comparisons.
\begin{appendixflowtable}
\centering
\caption{KV-only refinement: PG-19 PPL/KL and DGX Spark throughput at 128K. \methodname{}-KV uses +4/+4 bits, 1/128 selection, four-token groups, and an SSD backing tier with a 128 MiB RAM cache and skipped misses.}
\label{tab:kv_ssd_tradeoff-1}
\begingroup\fontsize{7}{8.5}\selectfont\setlength{\tabcolsep}{.65pt}\renewcommand{\arraystretch}{1.0}\setlength{\aboverulesep}{.2ex}\setlength{\belowrulesep}{.25ex}
\tablehead{\toprule
 & \multicolumn{3}{c}{\textbf{L3.1-8B-I}} & \multicolumn{3}{c}{\textbf{Qwen3-8B}}\\
\cmidrule(lr){2-4}\cmidrule(lr){5-7}
\textbf{Configuration} & \textbf{PPL} & \textbf{KL} & \textbf{tok/s} & \textbf{PPL} & \textbf{KL} & \textbf{tok/s}\\\midrule}
\tabletail{\bottomrule}\tablelasttail{\bottomrule}
\begin{supertabular*}{\linewidth}{@{\extracolsep{\fill}}@{}p{.39\linewidth}cccccc@{}}
BF16 KV & 10.32 & 0.000 & 7.3 & 12.89 & 0.000 & 6.7\\
KVarN 2/2 & 10.98 & 0.067 & 11.3 & 14.40 & 0.130 & 10.7\\
KIVI-2 & 10.90 & 0.059 & 8.8 & 13.85 & 0.106 & 8.3\\
KVQuant 2-bit & 11.71 & 0.125 & 3.9 & 14.40 & 0.167 & 4.1\\
TurboQuant 2.5-bit & 10.98 & 0.055 & 10.2 & 36.02 & 1.115 & 9.6\\
AQUA-KV 2-bit & 10.66 & 0.030 & 1.5 & 13.14 & 0.029 & 1.4\\
KVQuant 3-bit & 10.56 & 0.024 & 3.9 & 13.21 & 0.041 & 3.9\\
TurboQuant 3.5-bit & 10.43 & 0.010 & 10.4 & 13.99 & 0.102 & 9.9\\
AQUA-KV 3-bit & 10.39 & 0.006 & 1.5 & 12.98 & 0.017 & 1.4\\
KVarN 4/2 & 10.38 & 0.008 & 10.4 & 13.04 & 0.014 & 9.8\\
KIVI-4 & 10.33 & 0.002 & 8.6 & 12.90 & 0.004 & 8.0\\
KVQuant 4-bit & 10.38 & 0.007 & 3.3 & 12.96 & 0.011 & 3.4\\
AQUA-KV 4-bit & 10.32 & 0.002 & 1.5 & 12.95 & 0.016 & 1.3\\
\methodname{}-KV 2/2 & 10.47 & 0.011 & 11.4 & 13.32 & 0.030 & 10.9\\
\methodname{}-KV 3/2 & 10.34 & 0.002 & 11.0 & 12.98 & 0.005 & 10.5\\
\methodname{}-KV 3/3 & 10.33 & 0.001 & 10.7 & 12.93 & 0.004 & 10.2\cr
\end{supertabular*}\endgroup\par\vspace{3pt}
\end{appendixflowtable}

\begingroup\setlength{\tabcolsep}{3pt}\renewcommand{\arraystretch}{1.08}\fontsize{8}{9.5}\selectfont
\def\appendixmatrixcaption{}
\appendixfirstgroupreserve{13}
\begin{appendixflowtable}
\caption{SSD-backed KV refinement on DGX Spark: Llama-3.1-8B-Instruct, PG-19 at 128K. PPL, KL, and decode tok/s by resident bits, fetched group fraction, and RAM-cache capacity. The 1/128, 128 MiB configuration repeats Table~\ref{tab:kv_ssd_tradeoff-1}.}\label{tab:kv-cache-grid-llama}

\begingroup\fontsize{7.5}{9}\selectfont\setlength{\tabcolsep}{.65pt}\renewcommand{\arraystretch}{1.0}\setlength{\aboverulesep}{.2ex}\setlength{\belowrulesep}{.25ex}
\tablehead{\toprule
\textbf{GPU resident} & \textbf{Fetch} & \multicolumn{3}{c}{\textbf{64 MiB}} & \multicolumn{3}{c}{\textbf{128 MiB}}\\
\cmidrule(lr){3-5}\cmidrule(lr){6-8}
\textbf{KV} & \textbf{fraction} & \textbf{PPL} & \textbf{KL} & \textbf{tok/s} & \textbf{PPL} & \textbf{KL} & \textbf{tok/s}\\\midrule}
\tabletail{\bottomrule}\tablelasttail{\bottomrule}
\appendixfirstgroupreserve{3}
\begin{supertabular*}{\linewidth}{@{\extracolsep{\fill}}@{}p{.21\linewidth}p{.14\linewidth}cccccc@{}}
\appendixgroupreserve{3}
\multirow[c]{3}{=}{2/2} & 1/256 & 10.50 & 0.014 & 11.5 & 10.50 & 0.013 & 11.5\\
 & 1/128 & 10.48 & 0.013 & 11.2 & 10.47 & 0.011 & 11.4\\
 & 1/64 & 10.47 & 0.013 & 10.6 & 10.46 & 0.010 & 10.8\\
\midrule
\appendixgroupreserve{3}
\multirow[c]{3}{=}{3/2} & 1/256 & 10.34 & 0.003 & 11.1 & 10.34 & 0.002 & 11.1\\
 & 1/128 & 10.34 & 0.002 & 10.7 & 10.34 & 0.002 & 11.0\\
 & 1/64 & 10.34 & 0.002 & 10.3 & 10.34 & 0.002 & 10.4\\
\midrule
\appendixgroupreserve{3}
\multirow[c]{3}{=}{3/3} & 1/256 & 10.33 & 0.002 & 10.8 & 10.34 & 0.001 & 10.8\\
 & 1/128 & 10.33 & 0.001 & 10.5 & 10.33 & 0.001 & 10.7\\
 & 1/64 & 10.33 & 0.002 & 10.0 & 10.33 & 0.001 & 10.2\cr
\end{supertabular*}\endgroup\par\vspace{3pt}
\begingroup\fontsize{7.5}{9}\selectfont\setlength{\tabcolsep}{.65pt}\renewcommand{\arraystretch}{1.0}\setlength{\aboverulesep}{.2ex}\setlength{\belowrulesep}{.25ex}
\tablehead{\toprule
\textbf{GPU resident} & \textbf{Fetch} & \multicolumn{3}{c}{\textbf{256 MiB}} & \multicolumn{3}{c}{\textbf{512 MiB}}\\
\cmidrule(lr){3-5}\cmidrule(lr){6-8}
\textbf{KV} & \textbf{fraction} & \textbf{PPL} & \textbf{KL} & \textbf{tok/s} & \textbf{PPL} & \textbf{KL} & \textbf{tok/s}\\\midrule}
\tabletail{\bottomrule}\tablelasttail{\bottomrule}
\appendixfirstgroupreserve{3}
\begin{supertabular*}{\linewidth}{@{\extracolsep{\fill}}@{}p{.21\linewidth}p{.14\linewidth}cccccc@{}}
\appendixgroupreserve{3}
\multirow[c]{3}{=}{2/2} & 1/256 & 10.49 & 0.011 & 11.5 & 10.49 & 0.010 & 11.5\\
 & 1/128 & 10.46 & 0.009 & 11.4 & 10.45 & 0.007 & 11.3\\
 & 1/64 & 10.44 & 0.008 & 11.1 & 10.43 & 0.006 & 11.0\\
\midrule
\appendixgroupreserve{3}
\multirow[c]{3}{=}{3/2} & 1/256 & 10.34 & 0.002 & 11.1 & 10.34 & 0.002 & 11.1\\
 & 1/128 & 10.34 & 0.002 & 10.9 & 10.34 & 0.002 & 10.9\\
 & 1/64 & 10.33 & 0.002 & 10.6 & 10.33 & 0.001 & 10.6\\
\midrule
\appendixgroupreserve{3}
\multirow[c]{3}{=}{3/3} & 1/256 & 10.34 & 0.001 & 10.8 & 10.34 & 0.001 & 10.7\\
 & 1/128 & 10.33 & 0.001 & 10.6 & 10.33 & 0.001 & 10.6\\
 & 1/64 & 10.33 & 0.001 & 10.3 & 10.33 & 0.001 & 10.3\cr
\end{supertabular*}\endgroup\par\vspace{3pt}
\end{appendixflowtable}
\appendixfirstgroupreserve{13}
\begin{appendixflowtable}
\caption{KV refinement with no cache misses: Llama-3.1-8B-Instruct, PG-19 at 128K. PPL/KL by fetched group fraction, using four-token groups.}\label{tab:kv-nomiss-llama}
\begingroup\fontsize{7.5}{9}\selectfont\setlength{\tabcolsep}{1pt}\renewcommand{\arraystretch}{1.0}\setlength{\aboverulesep}{.2ex}\setlength{\belowrulesep}{.25ex}
\begin{tabular*}{\linewidth}{@{\extracolsep{\fill}}@{}lcccccc@{}}
\toprule
 & \multicolumn{6}{c}{\textbf{GPU resident KV}}\\
\cmidrule(lr){2-7}
 & \multicolumn{2}{c}{\textbf{2/2}} & \multicolumn{2}{c}{\textbf{3/2}} & \multicolumn{2}{c}{\textbf{3/3}}\\
\cmidrule(lr){2-3}\cmidrule(lr){4-5}\cmidrule(lr){6-7}
\textbf{Fetch fraction} & \textbf{PPL} & \textbf{KL} & \textbf{PPL} & \textbf{KL} & \textbf{PPL} & \textbf{KL}\\\midrule
1/256 & 10.44 & 0.005 & 10.34 & 0.001 & 10.34 & 0.001\\
1/128 & 10.42 & 0.003 & 10.33 & 0.001 & 10.33 & 0.001\\
1/64 & 10.39 & 0.003 & 10.33 & 0.001 & 10.33 & 0.001\\
\bottomrule\end{tabular*}\endgroup
\end{appendixflowtable}
\appendixfirstgroupreserve{13}
\begin{appendixflowtable}
\caption{SSD-backed KV refinement on DGX Spark: Qwen3-8B, PG-19 at 128K. PPL, KL, and decode tok/s by resident bits, fetched group fraction, and RAM-cache capacity. The 1/128, 128 MiB configuration repeats Table~\ref{tab:kv_ssd_tradeoff-1}.}\label{tab:kv-cache-grid-qwen}

\begingroup\fontsize{7.5}{9}\selectfont\setlength{\tabcolsep}{.65pt}\renewcommand{\arraystretch}{1.0}\setlength{\aboverulesep}{.2ex}\setlength{\belowrulesep}{.25ex}
\tablehead{\toprule
\textbf{GPU resident} & \textbf{Fetch} & \multicolumn{3}{c}{\textbf{64 MiB}} & \multicolumn{3}{c}{\textbf{128 MiB}}\\
\cmidrule(lr){3-5}\cmidrule(lr){6-8}
\textbf{KV} & \textbf{fraction} & \textbf{PPL} & \textbf{KL} & \textbf{tok/s} & \textbf{PPL} & \textbf{KL} & \textbf{tok/s}\\\midrule}
\tabletail{\bottomrule}\tablelasttail{\bottomrule}
\appendixfirstgroupreserve{3}
\begin{supertabular*}{\linewidth}{@{\extracolsep{\fill}}@{}p{.21\linewidth}p{.14\linewidth}cccccc@{}}
\appendixgroupreserve{3}
\multirow[c]{3}{=}{2/2} & 1/256 & 13.41 & 0.038 & 11.1 & 13.37 & 0.033 & 11.1\\
 & 1/128 & 13.36 & 0.036 & 10.6 & 13.32 & 0.030 & 10.9\\
 & 1/64 & 14.28 & 0.119 & 10.0 & 13.28 & 0.028 & 10.3\\
\midrule
\appendixgroupreserve{3}
\multirow[c]{3}{=}{3/2} & 1/256 & 12.99 & 0.007 & 10.6 & 12.99 & 0.006 & 10.7\\
 & 1/128 & 12.97 & 0.006 & 10.1 & 12.98 & 0.005 & 10.5\\
 & 1/64 & 13.14 & 0.022 & 9.7 & 12.97 & 0.005 & 9.9\\
\midrule
\appendixgroupreserve{3}
\multirow[c]{3}{=}{3/3} & 1/256 & 12.94 & 0.005 & 10.3 & 12.94 & 0.004 & 10.3\\
 & 1/128 & 12.93 & 0.004 & 9.8 & 12.93 & 0.004 & 10.2\\
 & 1/64 & 13.00 & 0.013 & 9.4 & 12.93 & 0.004 & 9.6\cr
\end{supertabular*}\endgroup\par\vspace{3pt}
\begingroup\fontsize{7.5}{9}\selectfont\setlength{\tabcolsep}{.65pt}\renewcommand{\arraystretch}{1.0}\setlength{\aboverulesep}{.2ex}\setlength{\belowrulesep}{.25ex}
\tablehead{\toprule
\textbf{GPU resident} & \textbf{Fetch} & \multicolumn{3}{c}{\textbf{256 MiB}} & \multicolumn{3}{c}{\textbf{512 MiB}}\\
\cmidrule(lr){3-5}\cmidrule(lr){6-8}
\textbf{KV} & \textbf{fraction} & \textbf{PPL} & \textbf{KL} & \textbf{tok/s} & \textbf{PPL} & \textbf{KL} & \textbf{tok/s}\\\midrule}
\tabletail{\bottomrule}\tablelasttail{\bottomrule}
\appendixfirstgroupreserve{3}
\begin{supertabular*}{\linewidth}{@{\extracolsep{\fill}}@{}p{.21\linewidth}p{.14\linewidth}cccccc@{}}
\appendixgroupreserve{3}
\multirow[c]{3}{=}{2/2} & 1/256 & 13.34 & 0.030 & 11.0 & 13.28 & 0.026 & 11.0\\
 & 1/128 & 13.30 & 0.026 & 10.9 & 13.23 & 0.021 & 10.9\\
 & 1/64 & 13.25 & 0.023 & 10.5 & 13.20 & 0.018 & 10.4\\
\midrule
\appendixgroupreserve{3}
\multirow[c]{3}{=}{3/2} & 1/256 & 12.98 & 0.005 & 10.6 & 12.97 & 0.004 & 10.6\\
 & 1/128 & 12.97 & 0.005 & 10.4 & 12.96 & 0.004 & 10.4\\
 & 1/64 & 12.96 & 0.004 & 10.1 & 12.96 & 0.003 & 10.0\\
\midrule
\appendixgroupreserve{3}
\multirow[c]{3}{=}{3/3} & 1/256 & 12.94 & 0.004 & 10.3 & 12.93 & 0.003 & 10.2\\
 & 1/128 & 12.93 & 0.003 & 10.1 & 12.93 & 0.003 & 10.0\\
 & 1/64 & 12.93 & 0.003 & 9.8 & 12.93 & 0.003 & 9.7\cr
\end{supertabular*}\endgroup\par\vspace{3pt}
\end{appendixflowtable}
\appendixfirstgroupreserve{13}
\begin{appendixflowtable}
\caption{KV refinement with no cache misses: Qwen3-8B, PG-19 at 128K. PPL/KL by fetched group fraction, using four-token groups.}\label{tab:kv-nomiss-qwen}
\begingroup\fontsize{7.5}{9}\selectfont\setlength{\tabcolsep}{1pt}\renewcommand{\arraystretch}{1.0}\setlength{\aboverulesep}{.2ex}\setlength{\belowrulesep}{.25ex}
\begin{tabular*}{\linewidth}{@{\extracolsep{\fill}}@{}lcccccc@{}}
\toprule
 & \multicolumn{6}{c}{\textbf{GPU resident KV}}\\
\cmidrule(lr){2-7}
 & \multicolumn{2}{c}{\textbf{2/2}} & \multicolumn{2}{c}{\textbf{3/2}} & \multicolumn{2}{c}{\textbf{3/3}}\\
\cmidrule(lr){2-3}\cmidrule(lr){4-5}\cmidrule(lr){6-7}
\textbf{Fetch fraction} & \textbf{PPL} & \textbf{KL} & \textbf{PPL} & \textbf{KL} & \textbf{PPL} & \textbf{KL}\\\midrule
1/256 & 13.05 & 0.013 & 12.93 & 0.002 & 12.90 & 0.002\\
1/128 & 13.01 & 0.009 & 12.93 & 0.002 & 12.90 & 0.002\\
1/64 & 12.97 & 0.006 & 12.92 & 0.002 & 12.90 & 0.001\\
\bottomrule\end{tabular*}\endgroup
\end{appendixflowtable}

\endgroup

\begin{appendixcolumntable}
\centering
\caption{KV-only refinement: DGX Spark throughput (tok/s) when waiting on cache misses at 128K, with four-token groups. Columns give RAM-cache capacities (MiB). Matching PPL/KL appears in Tables~\ref{tab:kv-nomiss-llama} and~\ref{tab:kv-nomiss-qwen}.}
\label{tab:kv-wait-cache}
\setlength{\aboverulesep}{.2ex}\setlength{\belowrulesep}{.25ex}
\fontsize{7.5}{9}\selectfont\setlength{\tabcolsep}{.65pt}\renewcommand{\arraystretch}{1.0}\setlength{\aboverulesep}{.2ex}\setlength{\belowrulesep}{.25ex}
\begin{tabular*}{\linewidth}{@{\extracolsep{\fill}}@{}p{.21\linewidth}p{.14\linewidth}cccccccc@{}}
\toprule
\textbf{GPU resident} & \textbf{Fetch} & \multicolumn{4}{c}{\textbf{L3.1-8B-I}} & \multicolumn{4}{c}{\textbf{Qwen3-8B}}\\
\cmidrule(lr){3-6}\cmidrule(lr){7-10}
\textbf{KV} & \textbf{fraction} & \textbf{64} & \textbf{128} & \textbf{256} & \textbf{512} & \textbf{64} & \textbf{128} & \textbf{256} & \textbf{512}\\\midrule
\multirow[c]{3}{*}{2/2} & 1/256 & 8.6 & 8.8 & 9.1 & 9.2 & 7.8 & 8.5 & 9.3 & 8.5\\
 & 1/128 & 6.8 & 8.1 & 8.7 & 9.2 & 6.5 & 7.7 & 8.6 & 8.4\\
 & 1/64 & 4.0 & 5.8 & 7.5 & 8.3 & 2.9 & 5.5 & 7.1 & 8.1\\
\midrule
\multirow[c]{3}{*}{3/2} & 1/256 & 8.2 & 8.6 & 9.0 & 9.0 & 7.8 & 8.4 & 8.9 & 8.3\\
 & 1/128 & 7.0 & 7.9 & 8.4 & 8.8 & 6.5 & 7.3 & 8.0 & 8.2\\
 & 1/64 & 3.9 & 6.0 & 7.5 & 8.0 & 3.1 & 5.5 & 6.9 & 8.2\\
\midrule
\multirow[c]{3}{*}{3/3} & 1/256 & 8.2 & 8.6 & 8.9 & 8.9 & 7.8 & 8.2 & 8.4 & 8.6\\
 & 1/128 & 6.9 & 7.8 & 8.2 & 8.6 & 6.2 & 7.1 & 7.9 & 8.5\\
 & 1/64 & 4.0 & 5.8 & 7.2 & 7.9 & 3.0 & 5.4 & 7.0 & 7.4\\
\bottomrule
\end{tabular*}

\end{appendixcolumntable}

\begin{minipage}{\linewidth}
\setlength{\parskip}{6pt}
\subsection{Refinement with 2-bit Keys and 1-bit Values}
\label{sec:kv-two-one}
Table~\ref{tab:resident-two-one} compares Base and \methodname{}-KV with 2-bit keys and 1-bit values.
\begingroup\setlength{\tabcolsep}{3pt}\renewcommand{\arraystretch}{1.08}\fontsize{8}{9.5}\selectfont
\def\appendixmatrixcaption{}
\begin{appendixcolumntable}\caption{KV refinement with resident 2/1 bits at 128K: PG-19 PPL/KL, RULER accuracy, Qwen3-8B MATH500 accuracy, RTX 5090 decode tok/s at 128K, and memory. $^{*}$KV memory only. \methodname{}-KV uses +0/+1 bits, two-token groups, and a 1/16 fetch fraction. In our Llama-3.1-8B-Instruct 32K sweep, +0/+1 yields the lowest PPL/KL among tested fetched-bitwidth pairs at matched payload.}\label{tab:resident-two-one}

\begingroup\fontsize{6.8}{8.3}\selectfont\setlength{\tabcolsep}{.6pt}\renewcommand{\arraystretch}{1.0}\setlength{\aboverulesep}{.2ex}\setlength{\belowrulesep}{.25ex}
\begin{tabular*}{\linewidth}{@{\extracolsep{\fill}}@{}lccccccccccc@{}}
\toprule
 & \multicolumn{5}{c}{\textbf{L3.1-8B-I}} & \multicolumn{6}{c}{\textbf{Qwen3-8B}}\\
\cmidrule(lr){2-6}\cmidrule(lr){7-12}
\textbf{Method} & \textbf{PPL} & \textbf{KL} & \textbf{R128} & \textbf{tok/s} & \textbf{GB\rlap{$^{*}$}} & \textbf{PPL} & \textbf{KL} & \textbf{R128} & \textbf{M500} & \textbf{tok/s} & \textbf{GB\rlap{$^{*}$}}\\\midrule
BF16 & 10.32 & 0.000 & 77.9 & 47.6 & 17.2 & 12.89 & 0.000 & 74.1 & 97.2 & OOM & 19.3\\
\midrule
KVarN 2/1 & 31.29 & 1.180 & 40.7 & 77.8 & 2.0 & 112.6 & 2.318 & 18.0 & 28.6 & 74.2 & 2.3\\
\methodname{}-KV 2/1 & 11.65 & 0.135 & 64.1 & 63.5 & 2.0 & 15.16 & 0.181 & 60.4 & 93.4 & 59.8 & 2.3\cr
\bottomrule\end{tabular*}\endgroup
\end{appendixcolumntable}

\endgroup

\end{minipage}

\begin{figure}[!t]
\centering
\includegraphics[width=\columnwidth,height=.79\textheight,keepaspectratio]{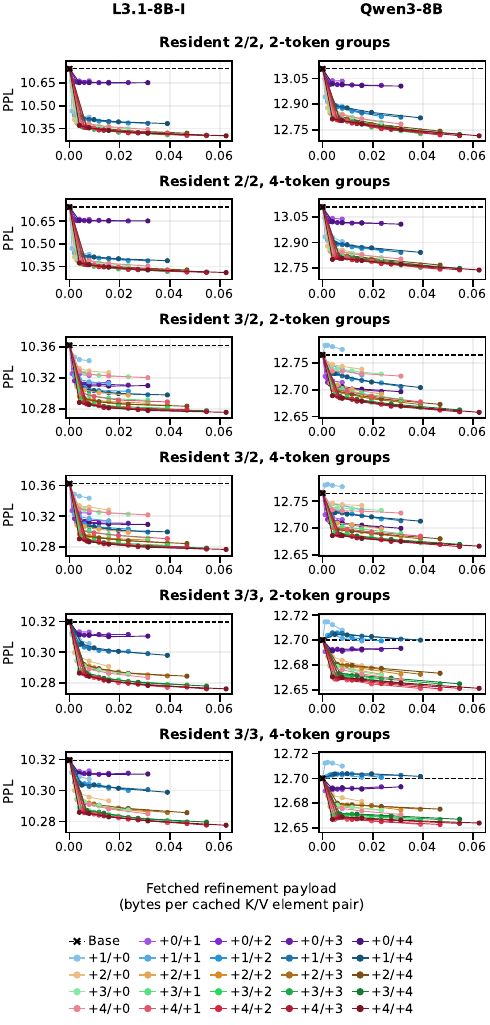}
\caption{KV refinement payload versus PG-19 PPL at 32K: left, Llama-3.1-8B-Instruct; right, Qwen3-8B. Rows compare resident 2/2, 3/2, and 3/3 with two- and four-token groups. Curves vary fetched key/value bits independently from 0 to 4, excluding 0/0; zero leaves that component unrefined. Color identifies fetched key bits; darker shades indicate more fetched value bits. Each curve shows all registered selected-group fractions from 1/256 to 1/16. Payload bytes per cached K/V element pair equal fetched key plus value bits, multiplied by the selected fraction and divided by eight. Dashed lines mark unrefined bases.}
\label{fig:kv-settings-sweep}
\label{fig:kv-settings-sweep-qwen}
\end{figure}

\begin{minipage}{\linewidth}
\setlength{\parskip}{6pt}
\subsection{KV Refinement Settings Sweep}
\label{sec:kv-settings-sweep}
Figure~\ref{fig:kv-settings-sweep} varies fetched key/value bitwidths and the selected token-group fraction for two- and four-token groups. +4/+4 performs well across evaluated fetch budgets and resident precisions, supporting its use in our main comparisons.
\end{minipage}

\section{Joint Configurations and Task Results}
\label{sec:appendix-joint}
These comparisons include the four configurations evaluated in Section~\ref{main-sec:eval-joint}. Table~\ref{tab:joint_kv_ablation-1} reports all four weight/KV refinement combinations at each resident format; Figure~\ref{main-fig:component-ablation} shows the YAQA 3-bit, 3/3 comparison.
\subsection{Configuration Comparisons}
Tables~\ref{tab:joint_accuracy-1}, \ref{tab:joint_pairs-1}, and~\ref{tab:joint_kv_ablation-1} compare inference-system configurations, static weight/KV pairs, and the four weight/KV refinement combinations.
\begin{appendixcolumntable}
\centering
\caption{Joint configurations at 128K: PG-19 PPL/KL, GPU memory (GB), and decode tok/s. $^{*}$DGX Spark; $^{\dagger}$RTX 5090. \methodname{} uses RTX 5090\textquotesingle{}s 5\% weight configurations. GB counts stored weights and KV. W/A/KV: weight/activation/cache bits; bpw: bits/weight; FP8: 8-bit floating point.}
\label{tab:joint_accuracy-1}
\begingroup\fontsize{6.2}{7.2}\selectfont\setlength{\tabcolsep}{.7pt}\renewcommand{\arraystretch}{1.0}\setlength{\aboverulesep}{.2ex}\setlength{\belowrulesep}{.25ex}
\begin{tabular*}{\linewidth}{@{\extracolsep{\fill}}@{}lcccccccccc@{}}
\toprule
 & \multicolumn{5}{c}{\textbf{L3.1-8B-I}} & \multicolumn{5}{c}{\textbf{Qwen3-8B}}\\
\cmidrule(lr){2-6}\cmidrule(lr){7-11}
\textbf{Configuration} & \textbf{PPL} & \textbf{KL} & \textbf{GB} & \textbf{tok/s\rlap{$^{*}$}} & \textbf{tok/s\rlap{$^{\dagger}$}} & \textbf{PPL} & \textbf{KL} & \textbf{GB} & \textbf{tok/s\rlap{$^{*}$}} & \textbf{tok/s\rlap{$^{\dagger}$}}\\\midrule
BF16 & 10.32 & 0.000 & 33.2 & 7.3 & 47.6 & 12.89 & 0.000 & 35.7 & 6.7 & OOM\\
\midrule
vLLM AWQ4 + FP8 KV & 11.03 & 0.074 & 14.3 & 16.8 & 106.8 & 13.49 & 0.078 & 15.8 & 15.3 & 96.6\\
EXL3 3.0 bpw + 4/4 KV & 11.39 & 0.107 & 8.9 & 25.6 & 99.6 & 14.00 & 0.125 & 9.8 & 23.6 & 91.3\\
EXL3 4.0 bpw + 4/4 KV & 10.78 & 0.051 & 9.8 & 23.6 & 101.7 & 13.73 & 0.086 & 10.6 & 21.9 & 92.9\\
EXL3 4.0 bpw + 2/2 KV & 36.26 & 1.364 & 7.6 & 25.7 & 104.7 & 46.43 & 1.394 & 8.2 & 24.0 & 95.6\\
llama.cpp Q4\_K\_M + q8\_0 KV & 10.59 & 0.027 & 14.0 & 12.7 & 63.1 & 13.17 & 0.027 & 15.3 & 11.5 & 57.2\\
llama.cpp Q4\_K\_M + q4\_0 KV & 10.88 & 0.062 & 9.8 & 14.4 & 64.6 & 13.94 & 0.110 & 10.5 & 13.0 & 58.7\\
llama.cpp Q3\_K\_M + q8\_0 KV & 11.13 & 0.073 & 13.1 & 13.0 & 64.2 & 13.75 & 0.092 & 14.4 & 11.9 & 58.1\\
llama.cpp Q3\_K\_M + q4\_0 KV & 11.46 & 0.109 & 8.9 & 15.0 & 65.9 & 14.64 & 0.172 & 9.6 & 13.5 & 59.6\\
QServe W4A8KV4 & 11.34 & 0.105 & 10.3 & 11.2 & 42.6 & 14.67 & 0.163 & 11.2 & 9.9 & 37.2\\
\bottomrule\end{tabular*}\endgroup
\par\vspace{1pt}
\begingroup\fontsize{6.2}{7.2}\selectfont\setlength{\tabcolsep}{.7pt}\renewcommand{\arraystretch}{1.0}\setlength{\aboverulesep}{.2ex}\setlength{\belowrulesep}{.25ex}
\begin{tabular*}{\linewidth}{@{\extracolsep{\fill}}@{}lcccccccc@{}}
\toprule
 & \multicolumn{4}{c}{\textbf{L3.1-8B-I}} & \multicolumn{4}{c}{\textbf{Qwen3-8B}}\\
\cmidrule(lr){2-5}\cmidrule(lr){6-9}
\textbf{Configuration} & \textbf{PPL} & \textbf{KL} & \textbf{GB} & \textbf{tok/s\rlap{$^{\dagger}$}} & \textbf{PPL} & \textbf{KL} & \textbf{GB} & \textbf{tok/s\rlap{$^{\dagger}$}}\\\midrule
BF16 & 10.32 & 0.000 & 33.2 & 47.6 & 12.89 & 0.000 & 35.7 & OOM\\
\midrule
\methodname{} / YAQA 2-bit / 2/2 & 13.79 & 0.312 & 6.4 & 106.6 & 14.91 & 0.165 & 7.1 & 97.6\\
\methodname{} / YAQA 2-bit / 3/2 & 13.76 & 0.313 & 7.0 & 104.2 & 14.88 & 0.162 & 7.7 & 95.4\\
\methodname{} / YAQA 2-bit / 3/3 & 13.74 & 0.311 & 7.5 & 102.1 & 14.84 & 0.161 & 8.3 & 93.3\\
\methodname{} / YAQA 3-bit / 2/2 & 10.80 & 0.045 & 7.3 & 102.0 & 13.25 & 0.038 & 8.0 & 91.7\\
\methodname{} / YAQA 3-bit / 3/2 & 10.74 & 0.044 & 7.8 & 100.3 & 13.20 & 0.034 & 8.6 & 90.3\\
\methodname{} / YAQA 3-bit / 3/3 & 10.72 & 0.042 & 8.4 & 98.3 & 13.17 & 0.034 & 9.2 & 88.2\\
\methodname{} / LNQ+GQ 3-bit / 2/2 & 12.98 & 0.241 & 7.3 & 117.9 & 17.46 & 0.292 & 8.0 & 105.1\\
\methodname{} / LNQ+GQ 3-bit / 3/2 & 12.95 & 0.242 & 7.9 & 115.8 & 17.29 & 0.286 & 8.6 & 103.3\\
\methodname{} / LNQ+GQ 3-bit / 3/3 & 12.95 & 0.242 & 8.4 & 113.1 & 17.27 & 0.285 & 9.2 & 100.6\\
\bottomrule\end{tabular*}\endgroup
\end{appendixcolumntable}

\appendixfirstgroupreserve{13}
\begin{appendixflowtable}
\centering
\caption{Joint refinement: static weight/KV pairs at 128K: PG-19 PPL/KL, GPU memory (GB), and RTX 5090 tokens/s. GB counts stored weights and KV. Neither weight nor KV refinement is enabled.}
\label{tab:joint_pairs-1}\label{tab:joint_pairs-2}
\begingroup\fontsize{6.7}{8.2}\selectfont\setlength{\tabcolsep}{.65pt}\renewcommand{\arraystretch}{1.0}\setlength{\aboverulesep}{.2ex}\setlength{\belowrulesep}{.25ex}
\tablehead{\toprule
 & & \multicolumn{4}{c}{\textbf{L3.1-8B-I}} & \multicolumn{4}{c}{\textbf{Qwen3-8B}}\\
\cmidrule(lr){3-6}\cmidrule(lr){7-10}
\textbf{Weight base} & \textbf{KV configuration} & \textbf{PPL} & \textbf{KL} & \textbf{GB} & \textbf{tok/s} & \textbf{PPL} & \textbf{KL} & \textbf{GB} & \textbf{tok/s}\\\midrule}
\tabletail{\bottomrule}\tablelasttail{\bottomrule}
\begin{supertabular*}{\linewidth}{@{\extracolsep{\fill}}@{}llcccccccc@{}}
BF16 & 16/16 & 10.32 & 0.000 & 33.2 & 47.6 & 12.89 & 0.000 & 35.7 & OOM\\
\midrule
\appendixgroupreserve{7}
\multirow[c]{7}{*}{YAQA w2} & KIVI-2 & 18.55 & 0.645 & 7.1 & 32.9 & 18.84 & 0.403 & 7.9 & 29.4\\
 & KIVI-4 & 15.97 & 0.477 & 9.2 & 35.2 & 16.59 & 0.257 & 10.3 & 31.5\\
 & KVQuant 2bit-1\% & 19.79 & 0.708 & 6.5 & 13.7 & 18.92 & 0.414 & 7.2 & 14.2\\
 & KVQuant 3bit-1\% & 16.64 & 0.521 & 7.6 & 13.2 & 17.09 & 0.290 & 8.4 & 13.5\\
 & KVQuant 4bit-1\% & 16.08 & 0.486 & 8.6 & 12.1 & 16.63 & 0.262 & 9.6 & 12.2\\
 & KVarN 2/2 & 18.67 & 0.656 & 6.4 & 84.7 & 19.48 & 0.434 & 7.1 & 76.6\\
 & KVarN 4/2 & 16.18 & 0.494 & 7.5 & 74.7 & 16.83 & 0.271 & 8.3 & 67.5\\
\midrule
\appendixgroupreserve{7}
\multirow[c]{7}{*}{YAQA w3} & KIVI-2 & 11.62 & 0.130 & 8.0 & 32.4 & 14.60 & 0.155 & 8.7 & 29.1\\
 & KIVI-4 & 10.90 & 0.059 & 10.1 & 34.6 & 13.42 & 0.049 & 11.2 & 31.1\\
 & KVQuant 2bit-1\% & 12.41 & 0.192 & 7.4 & 14.2 & 14.98 & 0.196 & 8.1 & 14.3\\
 & KVQuant 3bit-1\% & 11.17 & 0.084 & 8.4 & 13.6 & 13.68 & 0.081 & 9.3 & 13.6\\
 & KVQuant 4bit-1\% & 10.95 & 0.064 & 9.5 & 12.5 & 13.46 & 0.054 & 10.5 & 12.3\\
 & KVarN 2/2 & 11.75 & 0.146 & 7.3 & 81.7 & 15.15 & 0.185 & 8.0 & 74.4\\
 & KVarN 4/2 & 10.98 & 0.068 & 8.4 & 72.4 & 13.59 & 0.059 & 9.2 & 65.7\\
\midrule
\appendixgroupreserve{7}
\multirow[c]{7}{*}{YAQA w4} & KIVI-2 & 11.17 & 0.079 & 8.8 & 31.9 & 14.01 & 0.118 & 9.6 & 28.6\\
 & KIVI-4 & 10.54 & 0.018 & 11.0 & 34.1 & 13.04 & 0.017 & 12.0 & 30.6\\
 & KVQuant 2bit-1\% & 11.88 & 0.140 & 8.2 & 14.2 & 14.51 & 0.174 & 8.9 & 14.2\\
 & KVQuant 3bit-1\% & 10.77 & 0.041 & 9.3 & 13.7 & 13.33 & 0.051 & 10.1 & 13.5\\
 & KVQuant 4bit-1\% & 10.58 & 0.023 & 10.4 & 12.4 & 13.08 & 0.023 & 11.3 & 12.2\\
 & KVarN 2/2 & 11.24 & 0.090 & 8.2 & 79.0 & 14.58 & 0.143 & 8.9 & 71.8\\
 & KVarN 4/2 & 10.60 & 0.025 & 9.3 & 70.3 & 13.18 & 0.026 & 10.1 & 63.7\\
\midrule
\appendixgroupreserve{7}
\multirow[c]{7}{*}{LNQ+GQ w3} & KIVI-2 & 15.26 & 0.427 & 8.0 & 33.7 & 20.27 & 0.452 & 8.8 & 29.9\\
 & KIVI-4 & 13.70 & 0.306 & 10.1 & 36.2 & 18.20 & 0.331 & 11.2 & 32.3\\
 & KVQuant 2bit-1\% & 16.52 & 0.506 & 7.4 & 14.4 & 20.93 & 0.491 & 8.1 & 14.0\\
 & KVQuant 3bit-1\% & 14.18 & 0.342 & 8.5 & 13.8 & 18.78 & 0.366 & 9.3 & 13.4\\
 & KVQuant 4bit-1\% & 13.81 & 0.315 & 9.5 & 12.6 & 18.34 & 0.339 & 10.5 & 12.2\\
 & KVarN 2/2 & 15.39 & 0.442 & 7.3 & 90.9 & 20.76 & 0.473 & 8.0 & 81.6\\
 & KVarN 4/2 & 13.80 & 0.317 & 8.4 & 79.9 & 18.38 & 0.341 & 9.2 & 71.6\\
\midrule
\appendixgroupreserve{7}
\multirow[c]{7}{*}{LNQ+GQ w4} & KIVI-2 & 11.52 & 0.117 & 8.9 & 33.0 & 14.40 & 0.143 & 9.6 & 29.5\\
 & KIVI-4 & 10.79 & 0.046 & 11.0 & 35.5 & 13.30 & 0.040 & 12.1 & 31.7\\
 & KVQuant 2bit-1\% & 12.34 & 0.180 & 8.3 & 14.5 & 14.83 & 0.189 & 9.0 & 14.4\\
 & KVQuant 3bit-1\% & 11.04 & 0.068 & 9.3 & 13.9 & 13.60 & 0.073 & 10.2 & 13.7\\
 & KVQuant 4bit-1\% & 10.84 & 0.050 & 10.4 & 12.6 & 13.35 & 0.046 & 11.4 & 12.4\\
 & KVarN 2/2 & 11.62 & 0.129 & 8.2 & 86.1 & 14.91 & 0.167 & 8.9 & 77.8\\
 & KVarN 4/2 & 10.84 & 0.053 & 9.3 & 76.1 & 13.46 & 0.050 & 10.1 & 68.7\cr
\end{supertabular*}\endgroup
\end{appendixflowtable}

\begin{appendixflowtable}
\centering
\caption{Weight/KV component combinations at 128K: PG-19 PPL/KL, GPU memory (GB), and RTX 5090 tok/s. Each four-row group fixes resident formats and uses the 5\% weight configuration. All groups use the same CUDA attention implementation. GB counts weights and KV.}
\label{tab:joint_kv_ablation-1}\label{tab:joint_kv_ablation-2}
\begingroup\fontsize{6.7}{8.1}\selectfont\setlength{\tabcolsep}{.3pt}\renewcommand{\arraystretch}{1.0}\setlength{\aboverulesep}{.2ex}\setlength{\belowrulesep}{.25ex}
\tablehead{\toprule
 & & \multicolumn{2}{c}{\textbf{\methodname{}}} & \multicolumn{4}{c}{\textbf{L3.1-8B-I}} & \multicolumn{4}{c}{\textbf{Qwen3-8B}}\\
\cmidrule(lr){3-4}\cmidrule(lr){5-8}\cmidrule(lr){9-12}
\textbf{Weights} & \textbf{K/V} & \textbf{Weight} & \textbf{KV} & \textbf{PPL} & \textbf{KL} & \textbf{GB} & \textbf{tok/s} & \textbf{PPL} & \textbf{KL} & \textbf{GB} & \textbf{tok/s}\\\midrule}
\tabletail{\bottomrule}\tablelasttail{\bottomrule}
\begin{supertabular*}{\linewidth}{@{\extracolsep{\fill}}@{}llcccccccccc@{}}
\appendixgroupreserve{4}
\multirow[c]{4}{*}{YAQA 2-bit} & \multirow[c]{4}{*}{2/2} & -- & -- & 18.67 & 0.656 & \multirow[c]{4}{*}{6.4} & 137.5 & 19.48 & 0.434 & \multirow[c]{4}{*}{7.1} & 125.7\\
 &  & $\checkmark$ & -- & 15.57 & 0.458 &  & 133.5 & 17.02 & 0.311 &  & 123.3\\
 &  & -- & $\checkmark$ & 15.95 & 0.472 &  & 109.2 & 16.66 & 0.255 &  & 99.2\\
 &  & $\checkmark$ & $\checkmark$ & 13.79 & 0.312 &  & 106.6 & 14.91 & 0.165 &  & 97.6\\
\cmidrule(lr){2-12}
\appendixgroupreserve{4}
\multirow[c]{4}{*}{YAQA 2-bit} & \multirow[c]{4}{*}{3/2} & -- & -- & 16.42 & 0.511 & \multirow[c]{4}{*}{7.0} & 134.6 & 17.06 & 0.287 & \multirow[c]{4}{*}{7.7} & 123.0\\
 &  & $\checkmark$ & -- & 14.15 & 0.348 &  & 130.0 & 15.23 & 0.189 &  & 120.0\\
 &  & -- & $\checkmark$ & 15.91 & 0.473 &  & 107.2 & 16.62 & 0.255 &  & 97.4\\
 &  & $\checkmark$ & $\checkmark$ & 13.76 & 0.313 &  & 104.2 & 14.88 & 0.162 &  & 95.4\\
\cmidrule(lr){2-12}
\appendixgroupreserve{4}
\multirow[c]{4}{*}{YAQA 2-bit} & \multirow[c]{4}{*}{3/3} & -- & -- & 16.21 & 0.496 & \multirow[c]{4}{*}{7.5} & 132.5 & 16.85 & 0.274 & \multirow[c]{4}{*}{8.3} & 120.8\\
 &  & $\checkmark$ & -- & 13.94 & 0.329 &  & 128.1 & 15.05 & 0.179 &  & 118.3\\
 &  & -- & $\checkmark$ & 15.91 & 0.472 &  & 105.2 & 16.56 & 0.253 &  & 95.4\\
 &  & $\checkmark$ & $\checkmark$ & 13.74 & 0.311 &  & 102.1 & 14.84 & 0.161 &  & 93.3\\
\midrule
\appendixgroupreserve{4}
\multirow[c]{4}{*}{YAQA 3-bit} & \multirow[c]{4}{*}{2/2} & -- & -- & 11.75 & 0.146 & \multirow[c]{4}{*}{7.3} & 130.7 & 15.15 & 0.185 & \multirow[c]{4}{*}{8.0} & 120.2\\
 &  & $\checkmark$ & -- & 11.52 & 0.124 &  & 126.6 & 14.83 & 0.167 &  & 114.6\\
 &  & -- & $\checkmark$ & 10.96 & 0.058 &  & 105.2 & 13.48 & 0.048 &  & 95.8\\
 &  & $\checkmark$ & $\checkmark$ & 10.80 & 0.045 &  & 102.0 & 13.25 & 0.038 &  & 91.7\\
\cmidrule(lr){2-12}
\appendixgroupreserve{4}
\multirow[c]{4}{*}{YAQA 3-bit} & \multirow[c]{4}{*}{3/2} & -- & -- & 11.06 & 0.076 & \multirow[c]{4}{*}{7.8} & 128.1 & 13.71 & 0.071 & \multirow[c]{4}{*}{8.6} & 117.8\\
 &  & $\checkmark$ & -- & 10.89 & 0.060 &  & 123.5 & 13.46 & 0.058 &  & 112.2\\
 &  & -- & $\checkmark$ & 10.90 & 0.058 &  & 103.0 & 13.43 & 0.046 &  & 93.6\\
 &  & $\checkmark$ & $\checkmark$ & 10.74 & 0.044 &  & 100.3 & 13.20 & 0.034 &  & 90.3\\
\cmidrule(lr){2-12}
\appendixgroupreserve{4}
\multirow[c]{4}{*}{YAQA 3-bit} & \multirow[c]{4}{*}{3/3} & -- & -- & 10.98 & 0.067 & \multirow[c]{4}{*}{8.4} & 125.8 & 13.56 & 0.061 & \multirow[c]{4}{*}{9.2} & 115.9\\
 &  & $\checkmark$ & -- & 10.81 & 0.051 &  & 122.0 & 13.32 & 0.049 &  & 110.7\\
 &  & -- & $\checkmark$ & 10.89 & 0.057 &  & 100.8 & 13.39 & 0.045 &  & 92.0\\
 &  & $\checkmark$ & $\checkmark$ & 10.72 & 0.042 &  & 98.3 & 13.17 & 0.034 &  & 88.2\\
\midrule
\appendixgroupreserve{4}
\multirow[c]{4}{*}{YAQA 4-bit} & \multirow[c]{4}{*}{2/2} & -- & -- & 11.24 & 0.090 & \multirow[c]{4}{*}{8.2} & 123.9 & 14.58 & 0.143 & \multirow[c]{4}{*}{8.9} & 113.8\\
 &  & $\checkmark$ & -- & 11.17 & 0.085 &  & 119.8 & 14.47 & 0.137 &  & 111.1\\
 &  & -- & $\checkmark$ & 10.60 & 0.018 &  & 100.3 & 13.10 & 0.018 &  & 91.2\\
 &  & $\checkmark$ & $\checkmark$ & 10.55 & 0.015 &  & 97.5 & 13.04 & 0.015 &  & 90.1\\
\cmidrule(lr){2-12}
\appendixgroupreserve{4}
\multirow[c]{4}{*}{YAQA 4-bit} & \multirow[c]{4}{*}{3/2} & -- & -- & 10.66 & 0.031 & \multirow[c]{4}{*}{8.7} & 121.6 & 13.31 & 0.038 & \multirow[c]{4}{*}{9.5} & 111.4\\
 &  & $\checkmark$ & -- & 10.61 & 0.027 &  & 117.1 & 13.23 & 0.034 &  & 108.3\\
 &  & -- & $\checkmark$ & 10.53 & 0.017 &  & 98.3 & 13.05 & 0.014 &  & 89.6\\
 &  & $\checkmark$ & $\checkmark$ & 10.49 & 0.014 &  & 95.9 & 12.98 & 0.011 &  & 87.5\\
\cmidrule(lr){2-12}
\appendixgroupreserve{4}
\multirow[c]{4}{*}{YAQA 4-bit} & \multirow[c]{4}{*}{3/3} & -- & -- & 10.59 & 0.025 & \multirow[c]{4}{*}{9.3} & 119.7 & 13.16 & 0.028 & \multirow[c]{4}{*}{10.1} & 110.0\\
 &  & $\checkmark$ & -- & 10.55 & 0.021 &  & 115.7 & 13.09 & 0.025 &  & 107.0\\
 &  & -- & $\checkmark$ & 10.53 & 0.017 &  & 96.6 & 13.03 & 0.014 &  & 87.9\\
 &  & $\checkmark$ & $\checkmark$ & 10.48 & 0.013 &  & 94.6 & 12.96 & 0.011 &  & 85.8\\
\midrule
\appendixgroupreserve{4}
\multirow[c]{4}{*}{LNQ+GQ 3-bit} & \multirow[c]{4}{*}{2/2} & -- & -- & 15.39 & 0.442 & \multirow[c]{4}{*}{7.3} & 152.2 & 20.76 & 0.473 & \multirow[c]{4}{*}{8.0} & 137.6\\
 &  & $\checkmark$ & -- & 14.48 & 0.372 &  & 149.4 & 19.61 & 0.423 &  & 134.6\\
 &  & -- & $\checkmark$ & 13.68 & 0.301 &  & 119.2 & 18.42 & 0.336 &  & 107.0\\
 &  & $\checkmark$ & $\checkmark$ & 12.98 & 0.241 &  & 117.9 & 17.46 & 0.292 &  & 105.1\\
\cmidrule(lr){2-12}
\appendixgroupreserve{4}
\multirow[c]{4}{*}{LNQ+GQ 3-bit} & \multirow[c]{4}{*}{3/2} & -- & -- & 13.96 & 0.330 & \multirow[c]{4}{*}{7.9} & 150.0 & 18.61 & 0.354 & \multirow[c]{4}{*}{8.6} & 135.6\\
 &  & $\checkmark$ & -- & 13.23 & 0.269 &  & 147.2 & 17.63 & 0.308 &  & 132.1\\
 &  & -- & $\checkmark$ & 13.64 & 0.302 &  & 117.0 & 18.23 & 0.330 &  & 104.8\\
 &  & $\checkmark$ & $\checkmark$ & 12.95 & 0.242 &  & 115.8 & 17.29 & 0.286 &  & 103.3\\
\cmidrule(lr){2-12}
\appendixgroupreserve{4}
\multirow[c]{4}{*}{LNQ+GQ 3-bit} & \multirow[c]{4}{*}{3/3} & -- & -- & 13.83 & 0.318 & \multirow[c]{4}{*}{8.4} & 147.1 & 18.41 & 0.343 & \multirow[c]{4}{*}{9.2} & 133.1\\
 &  & $\checkmark$ & -- & 13.11 & 0.258 &  & 144.4 & 17.46 & 0.298 &  & 130.1\\
 &  & -- & $\checkmark$ & 13.64 & 0.301 &  & 114.1 & 18.20 & 0.329 &  & 102.6\\
 &  & $\checkmark$ & $\checkmark$ & 12.95 & 0.242 &  & 113.1 & 17.27 & 0.285 &  & 100.6\\
\midrule
\appendixgroupreserve{4}
\multirow[c]{4}{*}{LNQ+GQ 4-bit} & \multirow[c]{4}{*}{2/2} & -- & -- & 11.62 & 0.129 & \multirow[c]{4}{*}{8.2} & 140.2 & 14.91 & 0.167 & \multirow[c]{4}{*}{8.9} & 127.6\\
 &  & $\checkmark$ & -- & 11.52 & 0.120 &  & 137.7 & 14.83 & 0.161 &  & 124.6\\
 &  & -- & $\checkmark$ & 10.84 & 0.044 &  & 111.7 & 13.36 & 0.040 &  & 101.2\\
 &  & $\checkmark$ & $\checkmark$ & 10.77 & 0.038 &  & 110.4 & 13.30 & 0.035 &  & 99.5\\
\cmidrule(lr){2-12}
\appendixgroupreserve{4}
\multirow[c]{4}{*}{LNQ+GQ 4-bit} & \multirow[c]{4}{*}{3/2} & -- & -- & 10.90 & 0.060 & \multirow[c]{4}{*}{8.8} & 138.4 & 13.59 & 0.062 & \multirow[c]{4}{*}{9.5} & 125.8\\
 &  & $\checkmark$ & -- & 10.82 & 0.054 &  & 135.6 & 13.53 & 0.056 &  & 122.9\\
 &  & -- & $\checkmark$ & 10.78 & 0.044 &  & 110.0 & 13.31 & 0.038 &  & 99.5\\
 &  & $\checkmark$ & $\checkmark$ & 10.71 & 0.038 &  & 109.2 & 13.25 & 0.032 &  & 97.7\\
\cmidrule(lr){2-12}
\appendixgroupreserve{4}
\multirow[c]{4}{*}{LNQ+GQ 4-bit} & \multirow[c]{4}{*}{3/3} & -- & -- & 10.85 & 0.053 & \multirow[c]{4}{*}{9.3} & 136.1 & 13.44 & 0.052 & \multirow[c]{4}{*}{10.1} & 123.7\\
 &  & $\checkmark$ & -- & 10.77 & 0.046 &  & 133.6 & 13.38 & 0.047 &  & 120.8\\
 &  & -- & $\checkmark$ & 10.78 & 0.043 &  & 107.7 & 13.29 & 0.037 &  & 97.4\\
 &  & $\checkmark$ & $\checkmark$ & 10.71 & 0.037 &  & 106.8 & 13.23 & 0.032 &  & 95.5\cr
\end{supertabular*}\endgroup
\end{appendixflowtable}

\appendixfirstgroupreserve{9}
\subsection{Full Task Results}
Tables~\ref{tab:joint-extended}, \ref{tab:joint-full-grid-gb10}, and~\ref{tab:joint-full-grid-rtx5090} report RULER, LongBench, MATH500, and AIME results for the evaluated full-system configurations.
\begin{appendixflowtable}
\centering
\caption{Joint refinement at 128K on Llama-3.1-8B-Instruct and Qwen3-8B: RULER/MATH500/AIME accuracy, throughput, and GPU memory. GB includes weights, KV, and DGX Spark\textquotesingle{}s RAM caches. \methodname{} combines the 5\% weight configuration with 2/2 KV.}
\label{tab:joint-extended}
\begin{minipage}{\linewidth}\begingroup\fontsize{7}{8.5}\selectfont\setlength{\tabcolsep}{.4pt}\renewcommand{\arraystretch}{1.0}\setlength{\aboverulesep}{.2ex}\setlength{\belowrulesep}{.25ex}
\begin{tabular*}{\linewidth}{@{\extracolsep{\fill}}@{}lcccccccc@{}}
\toprule
\multicolumn{9}{@{}l}{\textbf{DGX Spark}}\\
\cmidrule(lr){1-9}
 & \multicolumn{3}{c}{\textbf{L3.1-8B-I}} & \multicolumn{5}{c}{\textbf{Qwen3-8B}}\\
\cmidrule(lr){2-4}\cmidrule(lr){5-9}
\textbf{Configuration} & \textbf{R128} & \textbf{tok/s} & \textbf{GB} & \textbf{R128} & \textbf{M500} & \textbf{AIME} & \textbf{tok/s} & \textbf{GB}\\\midrule
BF16 & 77.9 & 7.3 & 33.2 & 74.1 & 97.2 & 71.0 & 6.7 & 35.7\\
\midrule
\methodname{} (LNQ+GQ 4-bit) & 71.1 & 22.1 & 8.5 & 70.6 & 96.0 & 65.6 & 20.8 & 9.2\\
\methodname{} (YAQA 4-bit) & 73.5 & 21.4 & 8.4 & 70.3 & 96.2 & 69.0 & 20.3 & 9.1\\
\methodname{} (YAQA 3-bit) & 73.2 & 23.3 & 7.6 & 68.4 & 95.2 & 68.8 & 21.9 & 8.3\\
\methodname{} (YAQA 2-bit) & 60.5 & 25.6 & 6.7 & 65.2 & 89.6 & 40.6 & 23.9 & 7.4\cr
\bottomrule\end{tabular*}\endgroup\end{minipage}\par\vspace{3pt}
\begin{minipage}{\linewidth}\begingroup\fontsize{7}{8.5}\selectfont\setlength{\tabcolsep}{.4pt}\renewcommand{\arraystretch}{1.0}\setlength{\aboverulesep}{.2ex}\setlength{\belowrulesep}{.25ex}
\begin{tabular*}{\linewidth}{@{\extracolsep{\fill}}@{}lcccccccc@{}}
\toprule
\multicolumn{9}{@{}l}{\textbf{RTX 5090}}\\
\cmidrule(lr){1-9}
 & \multicolumn{3}{c}{\textbf{L3.1-8B-I}} & \multicolumn{5}{c}{\textbf{Qwen3-8B}}\\
\cmidrule(lr){2-4}\cmidrule(lr){5-9}
\textbf{Configuration} & \textbf{R128} & \textbf{tok/s} & \textbf{GB} & \textbf{R128} & \textbf{M500} & \textbf{AIME} & \textbf{tok/s} & \textbf{GB}\\\midrule
BF16 & 77.9 & 47.6 & 33.2 & 74.1 & 97.2 & 71.0 & OOM & 35.7\\
\midrule
LNQ+GQ 4-bit / 2/2 & 64.3 & 86.1 & 8.2 & 64.0 & 95.2 & 58.3 & 77.8 & 8.9\\
\methodname{} (LNQ+GQ 4-bit) & 75.2 & 110.4 & 8.2 & 73.8 & 94.4 & 61.5 & 99.5 & 8.9\\
YAQA 4-bit / 2/2 & 67.5 & 79.0 & 8.2 & 65.4 & 95.0 & 62.5 & 71.8 & 8.9\\
\methodname{} (YAQA 4-bit) & 77.2 & 97.5 & 8.2 & 73.5 & 96.4 & 69.2 & 90.1 & 8.9\\
YAQA 3-bit / 2/2 & 67.2 & 81.7 & 7.3 & 63.8 & 94.6 & 59.6 & 74.4 & 8.0\\
\methodname{} (YAQA 3-bit) & 77.1 & 102.0 & 7.3 & 71.2 & 95.4 & 65.6 & 91.7 & 8.0\\
YAQA 2-bit / 2/2 & 50.4 & 84.7 & 6.4 & 58.8 & 68.0 & 11.0 & 76.6 & 7.1\\
\methodname{} (YAQA 2-bit) & 63.8 & 106.6 & 6.4 & 68.3 & 91.2 & 47.1 & 97.6 & 7.1\\
llama.cpp Q4\_K\_M + q8\_0 & 75.3 & 63.1 & 14.0 & 72.0 & 95.6 & 71.9 & 57.2 & 15.3\\
EXL3 4 bpw + 4/4 & 75.5 & 101.7 & 9.8 & 71.6 & 94.4 & 66.9 & 92.9 & 10.6\\
vLLM AWQ4 + FP8 KV & 74.7 & 106.8 & 14.3 & 70.0 & 95.4 & 68.3 & 96.6 & 15.8\cr
\bottomrule\end{tabular*}\endgroup\end{minipage}\par\vspace{3pt}
\end{appendixflowtable}

\begin{appendixflowtable}
\centering
\caption{Joint refinement: DGX Spark task grid. R4--R128 give RULER accuracy at 4K--128K; LB/M500 report LongBench/MATH500 accuracy. LongBench prompts are truncated to 130,432 tokens here, versus 32,128 in the KV-only comparisons. Throughput uses 128K context. Each pair uses the same weight base and resident 2/2 KV; Base disables both refinements. Configurations match the main full-system table.}
\label{tab:joint-full-grid-gb10}
\begingroup\fontsize{7}{8.5}\selectfont\setlength{\tabcolsep}{.4pt}\renewcommand{\arraystretch}{1.0}\setlength{\aboverulesep}{.2ex}\setlength{\belowrulesep}{.25ex}
\tablehead{\toprule
\multicolumn{10}{@{}l}{\textbf{DGX Spark / Llama-3.1-8B-Instruct}}\\
\cmidrule(lr){1-10}
\textbf{Weights} & \textbf{Method} & \textbf{R4} & \textbf{R8} & \textbf{R16} & \textbf{R32} & \textbf{R64} & \textbf{R128} & \textbf{LB} & \textbf{tok/s}\\\midrule}
\tabletail{\bottomrule}\tablelasttail{\bottomrule}
\appendixfirstgroupreserve{2}
\begin{supertabular*}{\linewidth}{@{\extracolsep{\fill}}@{}llcccccccc@{}}
BF16 &  & 95.7 & 95.1 & 92.9 & 87.1 & 85.8 & 77.9 & 49.6 & 7.3\\
\midrule
\multicolumn{2}{@{}l}{llama.cpp Q4\_K\_M + q8\_0} & 95.5 & 95.0 & 93.2 & 87.8 & 85.3 & 75.3 & 49.1 & 12.7\\
\multicolumn{2}{@{}l}{EXL3 4 bpw + 4/4} & 95.3 & 94.7 & 92.6 & 88.7 & 84.9 & 75.5 & 48.1 & 23.6\\
\multicolumn{2}{@{}l}{vLLM AWQ4 + FP8 KV} & 95.1 & 94.1 & 92.1 & 87.6 & 83.2 & 74.7 & 48.4 & 16.8\\
\midrule
\appendixgroupreserve{2}
\multirow[c]{2}{*}{LNQ+GQ 4-bit} & Base & 93.7 & 92.2 & 89.9 & 84.0 & 79.8 & 64.3 & 47.9 & 21.3\\
 & \methodname{} & 95.4 & 94.2 & 92.2 & 86.1 & 84.0 & 71.1 & 48.3 & 22.1\\
\midrule
\appendixgroupreserve{2}
\multirow[c]{2}{*}{YAQA 4-bit} & Base & 94.2 & 93.6 & 91.2 & 86.0 & 80.3 & 67.5 & 48.7 & 21.2\\
 & \methodname{} & 95.9 & 94.9 & 93.1 & 87.3 & 84.5 & 73.5 & 49.0 & 21.4\\
\midrule
\appendixgroupreserve{2}
\multirow[c]{2}{*}{YAQA 3-bit} & Base & 93.8 & 92.8 & 90.4 & 85.7 & 80.6 & 67.2 & 46.8 & 23.0\\
 & \methodname{} & 95.0 & 94.7 & 92.6 & 86.9 & 84.0 & 73.2 & 47.0 & 23.3\\
\midrule
\appendixgroupreserve{2}
\multirow[c]{2}{*}{YAQA 2-bit} & Base & 91.6 & 90.5 & 87.1 & 80.4 & 73.0 & 50.4 & 46.2 & 24.8\\
 & \methodname{} & 93.9 & 93.2 & 90.9 & 85.5 & 80.0 & 60.5 & 46.5 & 25.6\cr
\end{supertabular*}\endgroup\par\vspace{3pt}
\begingroup\fontsize{6.5}{8.1}\selectfont\setlength{\tabcolsep}{.4pt}\renewcommand{\arraystretch}{1.0}\setlength{\aboverulesep}{.2ex}\setlength{\belowrulesep}{.25ex}
\tablehead{\toprule
\multicolumn{12}{@{}l}{\textbf{DGX Spark / Qwen3-8B}}\\
\cmidrule(lr){1-12}
\textbf{Weights} & \textbf{Method} & \textbf{R4} & \textbf{R8} & \textbf{R16} & \textbf{R32} & \textbf{R64} & \textbf{R128} & \textbf{LB} & \textbf{M500} & \textbf{AIME} & \textbf{tok/s}\\\midrule}
\tabletail{\bottomrule}\tablelasttail{\bottomrule}
\appendixfirstgroupreserve{2}
\begin{supertabular*}{\linewidth}{@{\extracolsep{\fill}}@{}llcccccccccc@{}}
BF16 &  & 95.2 & 93.8 & 93.8 & 92.1 & 80.6 & 74.1 & 49.4 & 97.2 & 71.0 & 6.7\\
\midrule
\multicolumn{2}{@{}l}{llama.cpp Q4\_K\_M + q8\_0} & 95.0 & 93.9 & 93.4 & 91.9 & 79.5 & 72.0 & 49.2 & 95.6 & 71.9 & 11.5\\
\multicolumn{2}{@{}l}{EXL3 4 bpw + 4/4} & 94.8 & 94.2 & 93.7 & 90.8 & 78.5 & 71.6 & 48.1 & 94.4 & 66.9 & 21.9\\
\multicolumn{2}{@{}l}{vLLM AWQ4 + FP8 KV} & 94.8 & 93.9 & 93.0 & 91.0 & 78.8 & 70.0 & 48.0 & 95.4 & 68.3 & 15.3\\
\midrule
\appendixgroupreserve{2}
\multirow[c]{2}{*}{LNQ+GQ 4-bit} & Base & 94.0 & 92.9 & 91.3 & 87.3 & 72.4 & 64.0 & 48.1 & 95.2 & 58.3 & 19.5\\
 & \methodname{} & 95.5 & 94.2 & 93.8 & 90.8 & 78.8 & 70.6 & 48.4 & 96.0 & 65.6 & 20.8\\
\midrule
\appendixgroupreserve{2}
\multirow[c]{2}{*}{YAQA 4-bit} & Base & 94.5 & 93.7 & 92.1 & 87.9 & 71.1 & 65.4 & 48.4 & 95.0 & 62.5 & 19.5\\
 & \methodname{} & 95.5 & 94.5 & 93.2 & 91.3 & 76.4 & 70.3 & 49.0 & 96.2 & 69.0 & 20.3\\
\midrule
\appendixgroupreserve{2}
\multirow[c]{2}{*}{YAQA 3-bit} & Base & 94.4 & 93.4 & 92.2 & 86.6 & 71.1 & 63.8 & 48.1 & 94.6 & 59.6 & 20.9\\
 & \methodname{} & 95.3 & 94.6 & 93.7 & 91.4 & 76.0 & 68.4 & 48.9 & 95.2 & 68.8 & 21.9\\
\midrule
\appendixgroupreserve{2}
\multirow[c]{2}{*}{YAQA 2-bit} & Base & 92.9 & 90.9 & 87.2 & 77.8 & 63.8 & 58.8 & 43.3 & 68.0 & 11.0 & 22.5\\
 & \methodname{} & 94.4 & 93.0 & 90.8 & 83.6 & 70.2 & 65.2 & 44.1 & 89.6 & 40.6 & 23.9\cr
\end{supertabular*}\endgroup\par\vspace{3pt}
\end{appendixflowtable}

\begin{appendixflowtable}
\centering
\caption{Joint refinement: RTX 5090 task grid. R4--R128 give RULER accuracy at 4K--128K; LB/M500 report LongBench/MATH500 accuracy. LongBench prompts are truncated to 130,432 tokens here, versus 32,128 in the KV-only comparisons. Throughput uses 128K context. Each pair uses the same weight base and resident 2/2 KV; Base disables both refinements. Configurations match the main full-system table.}
\label{tab:joint-full-grid-rtx5090}
\begingroup\fontsize{7}{8.5}\selectfont\setlength{\tabcolsep}{.4pt}\renewcommand{\arraystretch}{1.0}\setlength{\aboverulesep}{.2ex}\setlength{\belowrulesep}{.25ex}
\tablehead{\toprule
\multicolumn{10}{@{}l}{\textbf{RTX 5090 / Llama-3.1-8B-Instruct}}\\
\cmidrule(lr){1-10}
\textbf{Weights} & \textbf{Method} & \textbf{R4} & \textbf{R8} & \textbf{R16} & \textbf{R32} & \textbf{R64} & \textbf{R128} & \textbf{LB} & \textbf{tok/s}\\\midrule}
\tabletail{\bottomrule}\tablelasttail{\bottomrule}
\appendixfirstgroupreserve{2}
\begin{supertabular*}{\linewidth}{@{\extracolsep{\fill}}@{}llcccccccc@{}}
BF16 &  & 95.7 & 95.1 & 92.9 & 87.1 & 85.8 & 77.9 & 49.6 & 47.6\\
\midrule
\multicolumn{2}{@{}l}{llama.cpp Q4\_K\_M + q8\_0} & 95.5 & 95.0 & 93.2 & 87.8 & 85.3 & 75.3 & 49.1 & 63.1\\
\multicolumn{2}{@{}l}{EXL3 4 bpw + 4/4} & 95.3 & 94.7 & 92.6 & 88.7 & 84.9 & 75.5 & 48.1 & 101.7\\
\multicolumn{2}{@{}l}{vLLM AWQ4 + FP8 KV} & 95.1 & 94.1 & 92.1 & 87.6 & 83.2 & 74.7 & 48.4 & 106.8\\
\midrule
\appendixgroupreserve{2}
\multirow[c]{2}{*}{LNQ+GQ 4-bit} & Base & 93.7 & 92.2 & 89.9 & 84.0 & 79.8 & 64.3 & 47.9 & 86.1\\
 & \methodname{} & 95.4 & 94.3 & 91.7 & 86.2 & 84.7 & 75.2 & 48.4 & 110.4\\
\midrule
\appendixgroupreserve{2}
\multirow[c]{2}{*}{YAQA 4-bit} & Base & 94.2 & 93.6 & 91.2 & 86.0 & 80.3 & 67.5 & 48.7 & 79.0\\
 & \methodname{} & 95.9 & 94.7 & 93.5 & 87.4 & 85.3 & 77.2 & 49.1 & 97.5\\
\midrule
\appendixgroupreserve{2}
\multirow[c]{2}{*}{YAQA 3-bit} & Base & 93.8 & 92.8 & 90.4 & 85.7 & 80.6 & 67.2 & 46.8 & 81.7\\
 & \methodname{} & 94.9 & 94.7 & 93.0 & 86.4 & 83.8 & 77.1 & 47.0 & 102.0\\
\midrule
\appendixgroupreserve{2}
\multirow[c]{2}{*}{YAQA 2-bit} & Base & 91.6 & 90.5 & 87.1 & 80.4 & 73.0 & 50.4 & 46.2 & 84.7\\
 & \methodname{} & 94.1 & 92.9 & 91.4 & 84.2 & 81.4 & 63.8 & 46.5 & 106.6\cr
\end{supertabular*}\endgroup\par\vspace{3pt}
\begingroup\fontsize{6.5}{8.1}\selectfont\setlength{\tabcolsep}{.4pt}\renewcommand{\arraystretch}{1.0}\setlength{\aboverulesep}{.2ex}\setlength{\belowrulesep}{.25ex}
\tablehead{\toprule
\multicolumn{12}{@{}l}{\textbf{RTX 5090 / Qwen3-8B}}\\
\cmidrule(lr){1-12}
\textbf{Weights} & \textbf{Method} & \textbf{R4} & \textbf{R8} & \textbf{R16} & \textbf{R32} & \textbf{R64} & \textbf{R128} & \textbf{LB} & \textbf{M500} & \textbf{AIME} & \textbf{tok/s}\\\midrule}
\tabletail{\bottomrule}\tablelasttail{\bottomrule}
\appendixfirstgroupreserve{2}
\begin{supertabular*}{\linewidth}{@{\extracolsep{\fill}}@{}llcccccccccc@{}}
\multicolumn{2}{@{}l}{llama.cpp Q4\_K\_M + q8\_0} & 95.0 & 93.9 & 93.4 & 91.9 & 79.5 & 72.0 & 49.2 & 95.6 & 71.9 & 57.2\\
\multicolumn{2}{@{}l}{EXL3 4 bpw + 4/4} & 94.8 & 94.2 & 93.7 & 90.8 & 78.5 & 71.6 & 48.1 & 94.4 & 66.9 & 92.9\\
\multicolumn{2}{@{}l}{vLLM AWQ4 + FP8 KV} & 94.8 & 93.9 & 93.0 & 91.0 & 78.8 & 70.0 & 48.0 & 95.4 & 68.3 & 96.6\\
\midrule
\appendixgroupreserve{2}
\multirow[c]{2}{*}{LNQ+GQ 4-bit} & Base & 94.0 & 92.9 & 91.3 & 87.3 & 72.4 & 64.0 & 48.1 & 95.2 & 58.3 & 77.8\\
 & \methodname{} & 95.4 & 94.0 & 93.8 & 91.5 & 81.0 & 73.8 & 48.7 & 94.4 & 61.5 & 99.5\\
\midrule
\appendixgroupreserve{2}
\multirow[c]{2}{*}{YAQA 4-bit} & Base & 94.5 & 93.7 & 92.1 & 87.9 & 71.1 & 65.4 & 48.4 & 95.0 & 62.5 & 71.8\\
 & \methodname{} & 95.5 & 94.5 & 93.4 & 91.1 & 79.5 & 73.5 & 49.0 & 96.4 & 69.2 & 90.1\\
\midrule
\appendixgroupreserve{2}
\multirow[c]{2}{*}{YAQA 3-bit} & Base & 94.4 & 93.4 & 92.2 & 86.6 & 71.1 & 63.8 & 48.1 & 94.6 & 59.6 & 74.4\\
 & \methodname{} & 95.4 & 94.4 & 94.0 & 91.3 & 78.9 & 71.2 & 48.8 & 95.4 & 65.6 & 91.7\\
\midrule
\appendixgroupreserve{2}
\multirow[c]{2}{*}{YAQA 2-bit} & Base & 92.9 & 90.9 & 87.2 & 77.8 & 63.8 & 58.8 & 43.3 & 68.0 & 11.0 & 76.6\\
 & \methodname{} & 94.3 & 92.7 & 91.2 & 85.6 & 72.3 & 68.3 & 44.5 & 91.2 & 47.1 & 97.6\cr
\end{supertabular*}\endgroup\par\vspace{3pt}
\end{appendixflowtable}

\section{Applicability to Other KV Bases}
\label{sec:appendix-kv-bases}

The main KV implementation adopts KVarN's quantization format. Other bases require a refinement encoding and decoder that combine their resident codes with separately stored refinement bits. The selection of token groups and the full-sequence attention update remain the same, but the refinement levels must match the base's quantization bins---the value ranges mapped to each resident code---and its coordinate transformations and metadata.

\textbf{Nonuniform quantization bins.} Our KVQuant-style adaptation starts from sensitivity-weighted nonuniform codebooks~\citep{hooper2024kvquant} and fits finer levels within each bin using sensitivity-weighted calibration samples. Our TurboQuant adaptation uses its scalar quantizer optimized for mean squared error in the rotated, normalized coordinate system~\citep{zandieh2026turboquant}; each bin is refined using levels derived from the base's coordinate distribution restricted to that bin. Both keep the resident code as the identifier of the coarse bin and store the finer-level index separately.

\begin{minipage}{\linewidth}
\setlength{\parskip}{6pt}
\textbf{Evaluation scope.} Table~\ref{tab:kv_other_bases-1} compares Base and \methodname{}-KV reconstruction accuracy for the KVQuant and TurboQuant adaptations.

\begin{appendixcolumntable}
\centering
\caption{KV-only refinement: nested refinements for other resident formats, PG-19 at 128K. Base and \methodname{}-KV rows report PPL/KL side by side. Refinement uses +4/+4 bits, two-token groups, and a 1/128 fetch fraction.}
\label{tab:kv_other_bases-1}
\fontsize{7}{8.3}\selectfont
\setlength{\tabcolsep}{1.3pt}
\renewcommand{\arraystretch}{1.0}\setlength{\aboverulesep}{.2ex}\setlength{\belowrulesep}{.25ex}
\begin{tabular*}{\linewidth}{@{\extracolsep{\fill}}@{}l lcccc@{}}
\toprule
 & \multicolumn{1}{c}{\textbf{}} & \multicolumn{2}{c}{\textbf{L3.1-8B-I}} & \multicolumn{2}{c}{\textbf{Qwen3-8B}} \\
\cmidrule(lr){3-4}\cmidrule(lr){5-6}
\textbf{Resident base} & \textbf{Method} & \textbf{PPL} & \textbf{KL} & \textbf{PPL} & \textbf{KL} \\
\midrule
\multirow[c]{2}{*}{KVQuant 2-bit + 1\% outliers} & Base & 11.71 & 0.125 & 14.40 & 0.167 \\
 & \methodname{}-KV & 10.46 & 0.012 & 13.07 & 0.021 \\
\midrule
\multirow[c]{2}{*}{KVQuant 3-bit + 1\% outliers} & Base & 10.56 & 0.024 & 13.21 & 0.041 \\
 & \methodname{}-KV & 10.35 & 0.003 & 12.92 & 0.005 \\
\midrule
\multirow[c]{2}{*}{KVQuant 4-bit + 1\% outliers} & Base & 10.38 & 0.007 & 12.96 & 0.011 \\
 & \methodname{}-KV & 10.33 & 0.001 & 12.89 & 0.002 \\
\midrule
\multirow[c]{2}{*}{TurboQuant 2.5-bit} & Base & 10.98 & 0.055 & 36.02 & 1.115 \\
 & \methodname{}-KV & 10.47 & 0.005 & 14.03 & 0.091 \\
\midrule
\multirow[c]{2}{*}{TurboQuant 3.5-bit} & Base & 10.43 & 0.010 & 13.99 & 0.102 \\
 & \methodname{}-KV & 10.34 & 0.001 & 12.89 & 0.004 \\
\bottomrule
\end{tabular*}
\end{appendixcolumntable}

\end{minipage}
\section{Mac Evaluation}
\label{sec:appendix-mac}
Table~\ref{tab:mac-results} compares weight-only, KV-only, and joint refinement on a MacBook Air M3 (16 GB, Metal), using Llama-3.1-8B-Instruct with llama.cpp\textquotesingle{}s Q3\_K\_M weights and KVarN 2/2 KV.
\begin{appendixcolumntable}
\centering
\caption{MacBook Air M3 weight/KV refinement: WikiText-2 PPL/KL on one sequentially scored 16K-token window. Throughput averages 30 paired 32-token blocks at the listed initial contexts.}
\label{tab:mac-results}
\fontsize{7}{8.3}\selectfont
\setlength{\tabcolsep}{1.3pt}
\renewcommand{\arraystretch}{1.0}\setlength{\aboverulesep}{.2ex}\setlength{\belowrulesep}{.25ex}
\begin{tabular*}{\linewidth}{@{\extracolsep{\fill}}@{}lcccccc@{}}
\toprule
 & \multicolumn{2}{c}{\textbf{\methodname{}}} & & & \multicolumn{2}{c}{\textbf{Throughput (tok/s)}} \\
\cmidrule(lr){2-3}\cmidrule(lr){6-7}
\textbf{Configuration} & \textbf{Weight} & \textbf{KV} & \textbf{PPL} & \textbf{KL} & \textbf{Empty context} & \textbf{16K context} \\
\midrule
Base & -- & -- & 6.71 & 0.126 & 14.1 & 10.6 \\
\methodname{}-W & $\checkmark$ & -- & 6.66 & 0.118 & 15.1 & 11.3 \\
\methodname{}-KV & -- & $\checkmark$ & 6.44 & 0.081 & 13.8 & 9.9 \\
\methodname{} & $\checkmark$ & $\checkmark$ & 6.38 & 0.073 & 14.4 & 10.5 \\
\bottomrule
\end{tabular*}
\end{appendixcolumntable}

Weight refinement selects $k=2$ channels per 1,024 in all four projection sets. Both refinement components use separate 128 MiB SSD-backed RAM caches; KV refinement uses the main unified-memory settings.

\begingroup
\section{Parallel Weight Refinement with CUDA Streams}
\label{sec:concurrent-weight-refinement}

Separate CUDA streams do not guarantee parallel execution. If the base projection starts first, its large grid can occupy all streaming multiprocessors (SMs), leaving refinement queued until an SM becomes available. Merely assigning refinement to another stream does not prevent this serialization.

\textbf{DecDEC's projection-local overlap.} DecDEC~\citep{park2025decdec} keeps refinement on the current stream and places the base projection on another stream behind an event wait. The intent is for faster same-stream scheduling to start refinement before the cross-stream base projection, including during CUDA Graph replay. Refinement's small grid leaves SMs idle, allowing the scheduler to place the projection's thread blocks on those SMs while refinement runs.

The event is recorded before refinement, so the cross-stream dependency does not wait for refinement to finish. A second event joins the projection back into the current stream before the output is consumed; that stream's own ordering also ensures refinement is complete. Both kernels add to the output atomically. The snippets show the stream/event structure captured in the CUDA Graph, with kernel arguments and event/stream allocation omitted.

\begin{lstlisting}[language=C++,basicstyle=\fontsize{6.5}{7.5}\selectfont\ttfamily\color{black},columns=fullflexible,keepspaces=true,breaklines=true,breakatwhitespace=false,frame=single,framerule=0.3pt,rulecolor=\color{codeframe},backgroundcolor=\color{codebackground},framesep=2pt,xleftmargin=5pt,xrightmargin=5pt,aboveskip=10pt,belowskip=10pt]
cudaEventRecord(fork, current);
refine<<<small_grid,threads,0,current>>>(...);
cudaStreamWaitEvent(other, fork, 0);
base_gemv(..., other);
cudaEventRecord(done, other);
cudaStreamWaitEvent(current, done, 0);
\end{lstlisting}

\begin{minipage}{\linewidth}
\setlength{\parskip}{6pt}
\textbf{\methodname{}'s fork-and-continue scheduling.} We continue the graph on another stream while the fetch-lead or fetch-lag kernel remains in flight on the previous execution stream. Each outstanding operation $j$ has its own events and continuation stream. The graph waits for its completion only where its result is needed, allowing projection sets with different shift distances $s$ to overlap independently. For fetch lead, when the projection's input activations become available, we fork again: the refinement branch waits for the earlier fetch and applies refinement using those activations, while the base projection proceeds on another stream. The branches join before the projection output is consumed. Fetch lag instead joins before the deferred residual update is consumed. Contributions use atomic additions to initialized accumulators.

\begin{lstlisting}[language=C++,basicstyle=\fontsize{6.5}{7.5}\selectfont\ttfamily\color{black},columns=fullflexible,keepspaces=true,breaklines=true,breakatwhitespace=false,frame=single,framerule=0.3pt,rulecolor=\color{codeframe},backgroundcolor=\color{codebackground},framesep=2pt,xleftmargin=5pt,xrightmargin=5pt,aboveskip=10pt,belowskip=10pt]
cudaEventRecord(fork[j], current);
fetch_or_refine<<<small_grid,threads,0,current>>>(...);
cudaEventRecord(done[j], current);
cudaStreamWaitEvent(next[j], fork[j], 0);
current = next[j];
// Where j's result is consumed:
cudaStreamWaitEvent(current, done[j], 0);
\end{lstlisting}
\end{minipage}

\par\endgroup

\end{document}